%% file: main.tex
\PassOptionsToPackage{table,dvipsnames}{xcolor}

\documentclass[]{style} 

\usepackage{amsmath}
\usepackage{amssymb}
\usepackage{amsthm}
\usepackage{bm}

\usepackage{array}
\usepackage{longtable}

\usepackage{url}

\usepackage{soul}
\sethlcolor{yellow}
\newif\ifshowdiff\showdifftrue

\ifshowdiff  \else  \fi

\input{macros}

\input{math_commands.tex}

\definecolor{tabhead}{HTML}{EAF2F8}   
\definecolor{tabrow}{HTML}{EEF1F5}    
\newcommand{\wamheadrow}{%
  \rowcolor{tabhead}%
  \textbf{WAM} &
  \textbf{Date} &
  \textbf{Substrate} &
  \textbf{Backbone} &
  \textbf{Action coupling} &
  \textbf{Deployment} \\}
\newcommand{\wamband}[1]{%
  \rowcolor{tabhead}%
  \multicolumn{6}{c}{\textbf{#1}}\\}

\newif\ifshowplans
\showplanstrue   

\definecolor{PlanInk}{HTML}{244B6B}

\renewcommand\authorformat[2][]{{\normalsize\sffamily #2$^{#1}$}}
\renewcommand\affiliationformat[2][]{{\normalsize\sffamily #2}}
\newsavebox{\titlesubtitlebox}
\newcommand{\titlesubtitle}[1]{%
  \sbox{\titlesubtitlebox}{\large\sffamily\bfseries #1}%
  \makebox[\linewidth][c]{%
    \hbox{%
      \pdfsave\pdfsetmatrix{1 0 0.22 1}%
      \rlap{\usebox{\titlesubtitlebox}}%
      \pdfrestore
      \hskip\wd\titlesubtitlebox
    }%
  }%
}

\title{World Action Models: A Survey\\[-0.15em]\titlesubtitle{Dream Less, Act More}}

\author{Qiuhong Shen, Shihua Zhang, Yue Liao, Qi Li, Zhenxiong Tan,  Shizun Wang, \\ Shuicheng Yan, Xinchao Wang$^\dagger$}
\affiliation{National University of Singapore}

\abstract{
World Action Models (WAMs) are embodied predictive-action models that make a
forecast of the future available to action. Recent WAMs repurpose large video
generation models, and a parallel line relies on language or vision-language
backbones without a video-generation core. This rapid expansion has blurred the
boundary among broad world models, video generation models, action-grounded video
world models, Vision-Language-Action policies, and WAMs. This survey gives the field a common account. It first clarifies
these boundaries, then organizes existing works through two complementary views. The
first view asks what each method is required to generate, spanning rendered futures, latent futures, and video-generation-free action reasoning. The second view decomposes each
method by predictive substrate, backbone, action coupling, and deployment regime.
This anatomy supports a unified discussion of interactability, causality,
persistence, physical plausibility, and generalization, followed by data, evaluation,
and open challenges. Across these axes, a consistent design pattern
emerges: WAMs are not simply video generators with action heads, but
predictive-action methods whose design choices trade representational richness
against compute, memory, latency, and action-label cost. The field is moving
toward methods that generate less of the future while preserving what control
requires. The survey homepage is available at \url{https://world-action-models.github.io/}.
}

\begin{document}

\maketitle
\begingroup
\makeatletter
\def\@thefnmark{}
\@footnotetext{Corresponding author: Xinchao Wang (\email{xinchao@nus.edu.sg}).}
\makeatother
\endgroup

\tableofcontents
\vspace{1em}

\input{sections/01-introduction/introduction}
\input{sections/02-vwm-to-wam-and-map/vwm-to-wam-and-map}
\input{sections/03-design-philosophies/design-philosophies}
\input{sections/04-what-makes-a-wam/what-makes-a-wam}
\input{sections/05-core-properties/core-properties}
\input{sections/06-data-and-evaluation/data-and-evaluation}
\input{sections/07-open-challenges/open-challenges}
\input{sections/08-conclusion/conclusion}

\input{sections/_refs_nocite}

\bibliographystyle{assets/plainnat}
\bibliography{main}

\end{document}

%% file: macros.tex
\usepackage{todonotes}

%% file: math_commands.tex
\usepackage{amsmath,amsfonts,bm}

\def\eqref#1{equation~\ref{#1}}

\def\1{\bm{1}}

\DeclareMathAlphabet{\mathsfit}{\encodingdefault}{\sfdefault}{m}{sl}
\SetMathAlphabet{\mathsfit}{bold}{\encodingdefault}{\sfdefault}{bx}{n}



%% file: sections/01-introduction/introduction.tex
\section{Introduction}
\label{sec:intro}

The long-standing goal of embodied AI is to build agents that perceive, reason, and act in unstructured physical environments. Over the past two years the field has moved toward policies that do more than react to the current observation: they also anticipate how the world may change before choosing what to do.
The trend began with Vision-Language-Action (VLA) models, which repurpose pretrained vision-language models to map an observation and an instruction directly to an action. Works such as RT-2~\citep{RT-2}, OpenVLA~\citep{OpenVLA}, and $\pi_0$~\citep{pi0} showed that the semantic knowledge accumulated during internet-scale pretraining can be grounded into robotic behavior, generalizing across instructions, objects, and embodiments with modest fine-tuning.
A standard VLA, however, never models how the environment changes under its own intervention. It learns an observation-to-action mapping with no explicit account of physics, contact, or perspective change. This omission limits the policy's ability to reason about consequences before acting.

A rapidly growing body of work now equips such a policy with an explicit predictive component drawn from world models. Large-scale pretrained video generation models serve as a major source of such components for predicting future dynamics, but they are not the only choice among existing works.
Despite their variety, these works fall into three recognizable streams, distinguished by how far the model carries its prediction before it decodes an action.
The first stream adopts a video generation backbone all the way to pixels and then decodes action from the rendered future, as in UniPi~\citep{unipi} and the video-plan policies that followed~\citep{du2023vlp,ko2024avdc}.
The second keeps the same backbone but halts before pixel decoding, recovering action from intermediate latents, flow fields, or masks, as in VPP~\citep{hu2024vpp} and later latent-space foresight~\citep{pai2025mimicvideo,yan2026svam}.
The third forgoes the video generator entirely and predicts in a language-token, audio, or joint-embedding space, as in DUST~\citep{DUST}, Audio-WM~\citep{learningrobotmanipulation2025}, and latent-dynamics policies over frozen vision features~\citep{LDA1B}.
Despite these differences, the three streams share a contract that a plain VLA does not. Each makes a future substrate action-facing, either by training future and action prediction together or by using a separate action module that consumes the predicted future. We name this family \textbf{World Action Models (WAMs)}, following recent usage of this terminology, and make its definition precise in Section~\ref{sec:what-makes-wam}.

This label has spread faster than the understanding behind it. Works that call themselves World Action Models arrive from the video-generation, robot-learning, and language-model communities, and two of them can agree on the name while sharing almost no implementation details.
Two others can avoid the name entirely and yet build the same thing.
Underneath the disagreement lies a common interface question: what predicted future is retained for action, and where along the path is the action decoded.
Each answer buys predictive richness at a price in compute, memory, and latency inside a control loop, which is why the strongest WAMs tend to dream less of the future while still acting on what they need.

This survey organizes the resulting works so that any new method can be placed and compared, and it does so from two complementary viewpoints.
The first is a design-philosophy-level taxonomy, presented in Section~\ref{sec:philosophies}, that categorizes works by where action is decoded along the inference path and sorts every WAM into exactly one of three families: Render-and-Decode, which carries generation to pixels, Latent-Only, which stops at intermediate representations, and Video-Generation-Free, which is built without the video generator.
The second viewpoint, developed in Section~\ref{sec:what-makes-wam}, is a component-level anatomy. It places each WAM on four axes, the predictive substrate that represents the future available to action, the architectural backbone that produces the prediction, the action coupling that joins prediction to control, and the deployment regime the model is built for. A unified notation, made precise in that section, expresses every WAM as a 4-tuple over these axes. Across both viewpoints we cover the works in our census, listed in Tables~\ref{tab:wams-video} and~\ref{tab:wams-genfree}, the first holding the video-generation-based families of Render-and-Decode and Latent-Only and the second the Video-Generation-Free family that drops the video backbone. The two viewpoints are complementary, since the philosophy fixes what a model is required to generate while the 4-tuple fixes how it is built.

Organizing the design space is only half of what this survey provides. The other half is a critical account of what these models must do once they are deployed in embodied settings. We examine five properties that embodiment demands, treated in Section~\ref{sec:core-properties}. A WAM must be interactable, accepting control signals during generation rather than only at the start. It must be causal, never letting the future leak into the action executed now. It must be persistent, holding long-horizon predictions together as the robot acts and replans. It must be physically plausible, predicting futures that the embodiment can actually realize. And it must generalize, keeping the same predictive-action contract useful when tasks, objects, scenes, cameras, and embodiments change. For each property we examine how existing methods meet or miss it, and we track what their solutions cost in compute, memory, data, and the effort to evaluate them.

The remainder of this survey is organized as follows.

\begin{itemize}
    \item \textbf{From World Models to World Action Models (Section~\ref{sec:vwm-to-wam}).} The discussion separates VLAs, broad world models, video generation models, video world models, and WAMs, then defines the action-facing future contract that turns a world model into a WAM.
    \item \textbf{Three Design Philosophies of World Action Models (Section~\ref{sec:philosophies}).} The philosophy-level taxonomy places every WAM considered here into Render-and-Decode, Latent-Only, or Video-Generation-Free according to what each method is required to generate.
    \item \textbf{What Makes a World Action Model (Section~\ref{sec:what-makes-wam}).} The formal anatomy introduces the unified notation and the four axes: predictive substrate, architectural backbone, action coupling, and deployment regime. It then records each WAM as a 4-tuple in Tables~\ref{tab:wams-video} and~\ref{tab:wams-genfree}.
    \item \textbf{Core Properties of World Action Models (Section~\ref{sec:core-properties}).} The property discussion examines interactability, causality, persistence, physical plausibility, and generalization, together with how existing methods address each requirement.
    \item \textbf{Data and Evaluation (Section~\ref{sec:data-eval}).} The section organizes the data sources that fuel WAM training, from teleoperation and portable human demonstrations to simulation and internet-scale egocentric video, alongside the evaluation practices the field is settling on.
    \item \textbf{Open Challenges (Section~\ref{sec:open}).} The final discussion identifies the open problems that the four-axis decomposition brings into view, grouped by where they arise in the WAM stack.
\end{itemize}

Throughout, one observation persists through the survey. Every design choice in a WAM carries a practical consequence for compute, memory, and latency inside a control loop. The interesting question is rarely how much a WAM costs on its own.
It is how that cost trades against the properties embodiment imposes. Viewed over time, the field has moved toward generating less of the future while holding on to what control requires.

%% file: sections/02-vwm-to-wam-and-map/vwm-to-wam-and-map.tex
\section{The Emergence of World Action Models}
\label{sec:vwm-to-wam}


\subsection{Vision-Language-Action Models: Policies from Vision-Language Pretraining}
\label{sec:vla-models}

Consider an embodied agent that observes $o$, receives an optional language instruction $l$, and must choose an action $a$. Vision-Language-Action (VLA) models became influential because they showed that pretrained vision-language and language backbones could be grounded into this action channel. In its simplest form, a VLA learns
\begin{equation}
\mathcal{L}_{\text{VLA}}(\theta)=
\mathbb{E}_{(o,l,a)}
\!\left[-\log p_\theta(a\mid o,l)\right].
\label{eq:vla}
\end{equation}
RT-2~\citep{RT-2} established the web-to-robot transfer pattern by co-fine-tuning a VLM to output action tokens. OpenVLA~\citep{OpenVLA} made the recipe open and scalable across robot data mixtures, while the $\pi$ series~\citep{pi0,pi05,pi0.7} moved the action side toward continuous flow matching and broader embodiment coverage. These methods are the closest neighbors of WAMs because they already bind vision, language, and action in one action-producing policy.

The VLA objective also states the boundary. Equation~\ref{eq:vla} does not require the model to predict what will be observed after the action is executed. A direct VLA can inherit semantic and spatial priors from a pretrained VLM, yet still act from the present observation alone. It becomes a WAM only when a predicted future observation helps produce, choose, or check the action.

\subsection{World Models: Predictive Dynamics Beyond Video}
\label{sec:world-models}

A world model asks the complementary question. Rather than predicting the action directly, it predicts what future observation $o'$ should follow from the current observation and an intervention. A recent robot-learning survey frames world models as predictive representations of how environments evolve under actions, with uses in policy learning, planning, simulation, evaluation, and data generation~\citep{hou2026worldmodelrobotlearning}. For the present survey, the minimal form is
\begin{equation}
\mathcal{L}_{\text{WM}}(\theta)=
\mathbb{E}_{(o,l,a,o')}
\!\left[-\log p_\theta(o'\mid o,a,l)\right],
\label{eq:wm}
\end{equation}
where $o'$ is written as an observation for clarity. The symbol is deliberately broad in this section. Depending on the method, it may be rendered pixels, a hidden feature, a geometric state, an affordance map, an audio cue, a symbolic state, or a token-level description. Section~\ref{sec:what-makes-wam} later separates these cases. In latent model-based reinforcement learning, PlaNet~\citep{PlaNet}, DreamerV3~\citep{DreamerV3}, TransDreamer~\citep{TransDreamer}, and Dreamer~4~\citep{Dreamer4} instantiate this idea with compact dynamics states used for imagination and planning. V-JEPA~\citep{V-JEPA} and V-JEPA~2~\citep{V-JEPA-2} show the same predictive idea in feature space, while iVideoGPT~\citep{iVideoGPT}, RoboDreamer~\citep{RoboDreamer}, EnerVerse~\citep{EnerVerse}, and InteractiveWorldSimulator~\citep{InteractiveWorldSimulator} express it through token or video prediction.

Video generation is one visible instance of this broader idea, not its definition. A video generation model learns
\begin{equation}
\mathcal{L}_{\text{VGM}}(\theta)=
\mathbb{E}_{(r,o')}\!\left[-\log p_\theta(o'\mid r)\right],
\label{eq:vgm}
\end{equation}
where $r$ is a prompt, image, observation, or other conditioning signal. When the conditioning includes the agent's action and the output is a future visual observation, the same family becomes a video world model:
\begin{equation}
\mathcal{L}_{\text{VWM}}(\theta)=
\mathbb{E}_{(o,l,a,o')}
\!\left[-\log p_\theta(o'\mid o,a,l)\right].
\label{eq:vwm}
\end{equation}
Internet-scale video generators such as Wan, CogVideoX, Cosmos, NOVA, Sora, Latte, AnimateDiff, and VideoPoet~\citep{Wan,CogVideoX,cosmos-predict2,NOVA,Sora,Sora_2,Latte,AnimateDiff,VideoPoet} made this branch newly influential because they supplied transferable visual dynamics priors. The rise of WAMs follows from this broader world-model success. Video world models gave the field a high-capacity source of predictive structure, while feature, geometric, audio, symbolic, and token predictors showed that the useful future need not always be rendered video.

\subsection{World Action Models: Making the Future Action-Facing}
\label{sec:vwm-shift}

\begin{figure}[t]
  \centering
  \includegraphics[width=\linewidth]{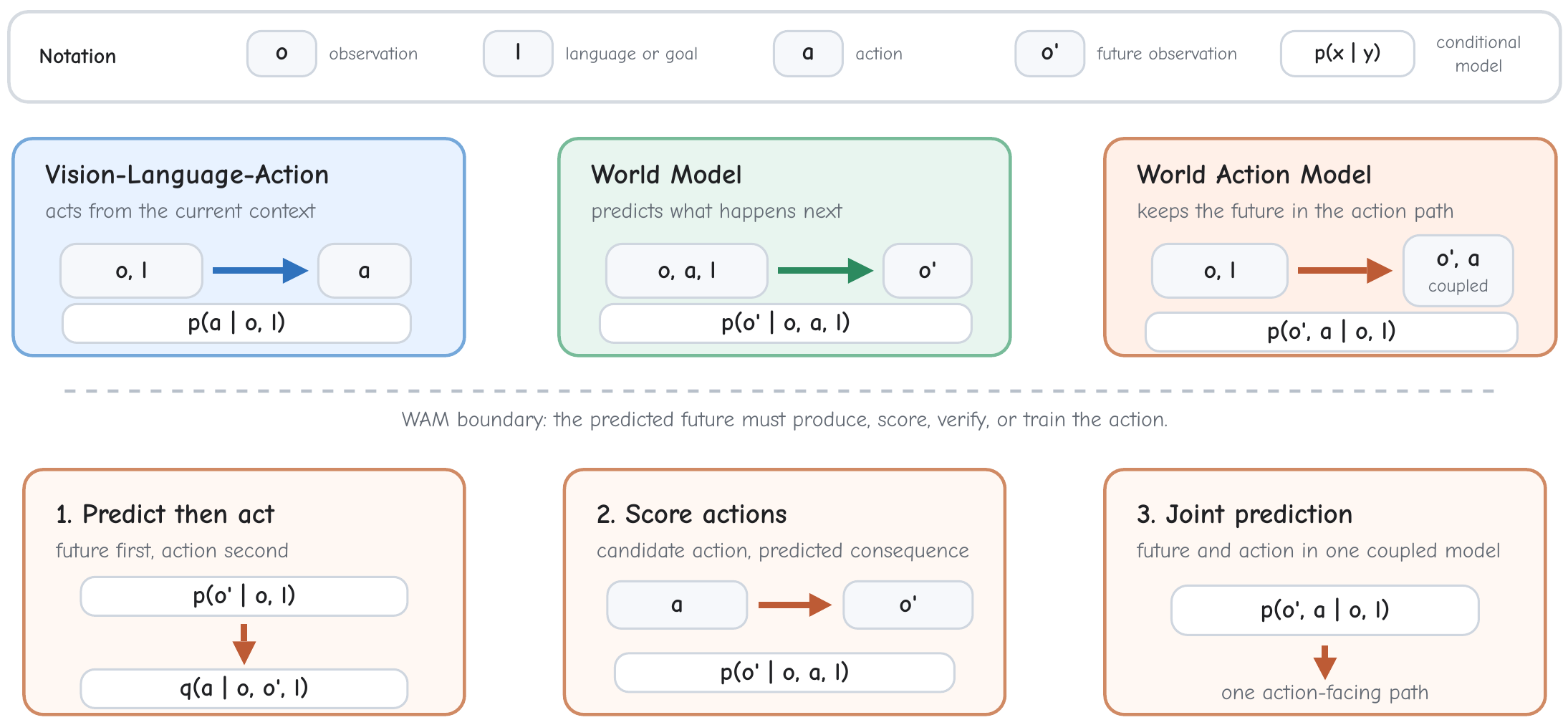}
  \caption{Definition of a World Action Model. A direct VLA predicts action from the present context, and a world model predicts a future observation. A WAM requires that future to stay in the action path, either through predict-then-act cascades, action-scoring rollouts, or joint future-action prediction.}
  \label{fig:wam-definition}
\end{figure}

A World Action Model begins when the predicted future observation becomes part of how the action is obtained. The historically early form is a cascade:
\begin{equation}
p_\Theta(o',a\mid o,l)
=p_\theta(o'\mid o,l)\,
q_\psi(a\mid o,o',l).
\label{eq:wam-cascade}
\end{equation}
The action module $q_\psi$ may be an inverse-dynamics model, a pose tracker, a trajectory optimizer, a planner, or a separately trained policy. Presenting this cascaded form first is historically defensible. Early WAMs such as UniPi~\citep{unipi}, VLP~\citep{du2023vlp}, and AVDC~\citep{ko2024avdc} generated or scored future visual trajectories before recovering executable actions with a separate module. Later cascaded WAMs keep the same logic while changing what $o'$ looks like, as in video-to-trajectory, flow-to-policy, trace-to-control, point-track-to-policy, or future-audio-to-policy pipelines~\citep{DreamGen,liu2025fourdgen,patel2025rigvid,li2025novaflow,dharmarajan2025dream2flow,tracegenworldmodeling2025,3pointr3dpoint2026,roboflow4dlightweightflow2026,learningrobotmanipulation2025}. These works remain WAMs because the predicted future is used to obtain the action, even when the predictor and actor are trained separately.

A second form proposes an action first and then predicts its consequence:
\begin{equation}
p_\Theta(o',a\mid o,l)
=q_\psi(a\mid o,l)\,
p_\theta(o'\mid o,a,l).
\label{eq:wam-rollout}
\end{equation}
This factorization covers candidate-action scoring and planning, where a proposed action is evaluated through its predicted consequence. It is still a WAM when the predicted consequence decides which action is executed~\citep{3danchoredlookahead2026,dreamavoidcriticalphase2026,egoexowmunlocking2026,pointworldscaling3d2026,feedbackworldmodel2026}.

The same idea can also live inside one trained model:
\begin{equation}
\mathcal{L}_{\text{WAM}}(\theta)=
\mathbb{E}_{(o,l,o',a)}
\!\left[-\log p_\theta(o',a\mid o,l)\right].
\label{eq:wam}
\end{equation}
Here one backbone or tightly coupled set of experts predicts the future observation and the action together. GR-1 and GR-2~\citep{GR-1,GR2} helped establish this shared video-action direction with autoregressive image and action tokens, while PAD~\citep{PAD}, UWM~\citep{UWM}, WorldVLA~\citep{WorldVLA}, DreamZero~\citep{DreamZero}, AIM~\citep{AIM}, and newer joint designs~\citep{drivewamvideogenerative2026,wallwmcarving2026,tau0wm2026} instantiate the same idea through diffusion, autoregression, or hybrid backbones. Feature-tap methods such as Fast-WAM~\citep{FastWAM} show that the future branch need not be fully rendered at inference as long as the action path still uses a representation shaped by future prediction.

The WAM definition is therefore direct and is summarized in Figure~\ref{fig:wam-definition}. A VLA models $p(a\mid o,l)$ without needing an explicit future. A world model models $p(o'\mid o,a,l)$ or $p(o'\mid o,l)$ without necessarily choosing the action. A WAM links the two: its predicted future observation helps produce the action, score the action, or train the action path inside one model. This definition uses $o'$ as a Section~2 notational simplification. Section~\ref{sec:what-makes-wam} instantiates that simplification as different predictive substrates and expands the equations into horizon-level notation. A direct VLA with an auxiliary future loss, a simulator used only as an RL environment, or a future head discarded before action use does not satisfy the definition.

\FloatBarrier

%% file: sections/03-design-philosophies/design-philosophies.tex
\section{Three Design Philosophies of World Action Models}
\label{sec:philosophies}


Existing world action models share the WAM definition introduced in Section~\ref{sec:vwm-to-wam}. They differ, however, in where the future predictor meets the action module. Some methods train a shared video-action backbone, while others keep a world model and a separate policy, tracker, inverse-dynamics module, or optimizer in a cascade. The training pipeline alone therefore does not expose the main design split. The decisive distinction is visible either in the inference forward pass or in how action supervision is wired during training: action prediction may pass through a rendered pixel future, stop inside a video backbone before rendering, or avoid the video-generation path altogether.

The resulting taxonomy has three mutually exclusive philosophies. \textbf{Render-and-Decode} runs the video-generation backbone all the way to pixel output before action is obtained. The action module may be trained jointly with the generator, trained separately, or supplied by tracking and trajectory optimization. \textbf{Latent-Only} keeps the video-world-model lineage but stops the inference path before pixel decoding. In many cases this means a video-generation trunk whose VAE decoder is bypassed, while in others it means a video-trained or video-derived latent predictor whose action-facing future has no decoded-pixel stage. Its action signal comes from intermediate latents, partially denoised features, flow fields, semantic masks, or value maps. \textbf{Video-Generation-Free} removes the video-generation backbone from the predictive path. The predictive component instead produces a future representation in the embedding or token space of a large language model, a vision-language model, a joint-embedding-predictive encoder, a deterministic regressor over a frozen vision foundation model, or a non-video diffusion or hybrid backbone over a compact non-pixel substrate. The taxonomy therefore tracks the requirement imposed during inference or action-supervision, rather than the training data, backbone family, or action interface alone.

\begin{figure}[t]
  \centering
  \includegraphics[width=\linewidth]{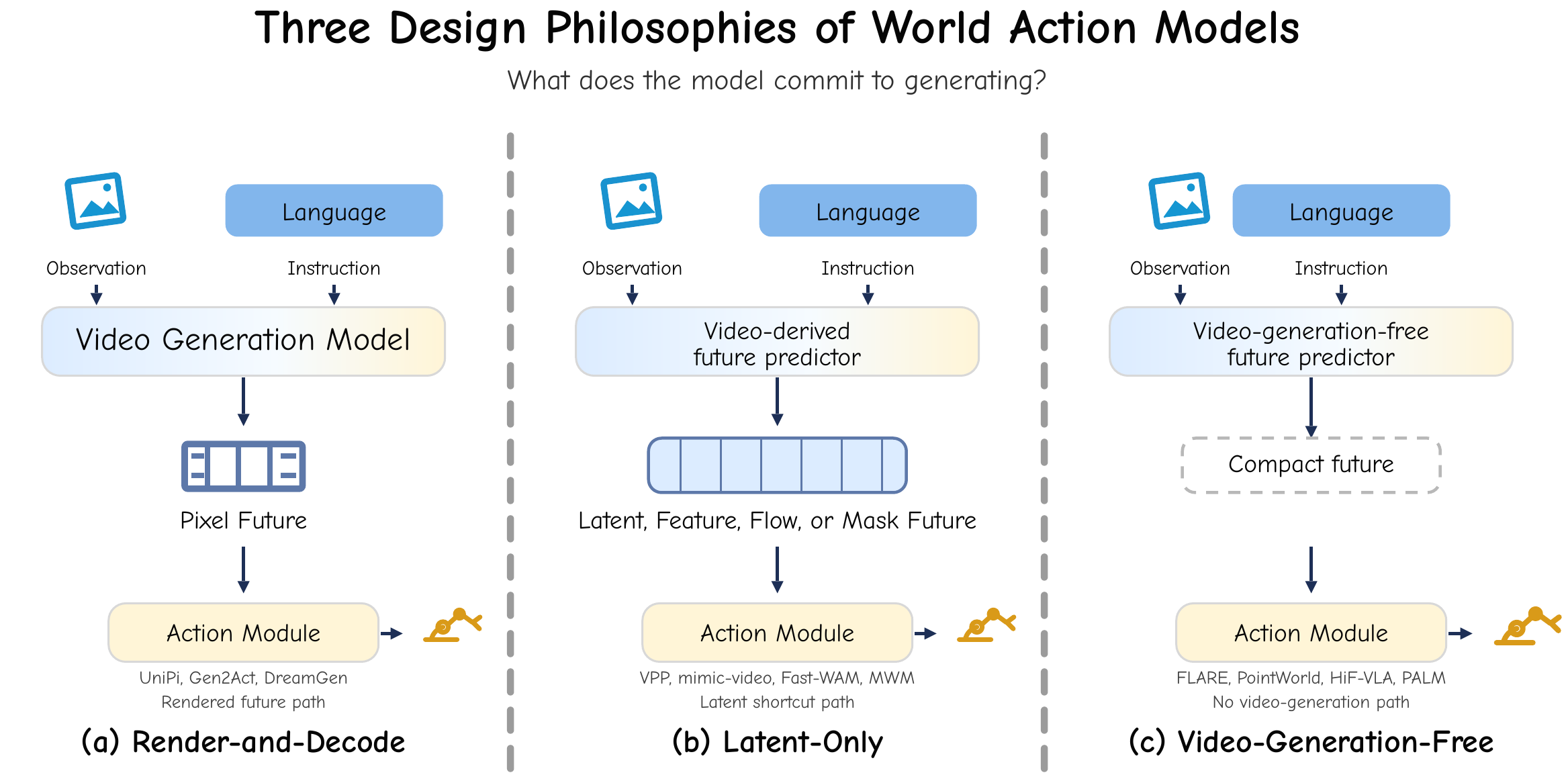}
  \caption{Three design philosophies of World Action Models. The columns separate the last future representation required before action is decoded: a rendered pixel future, an intermediate video-derived latent or feature, or a non-video-generation representation. The categorization is exhaustive over the WAM census and separable from the action-coupling and backbone choices treated in Section~\ref{sec:what-makes-wam}.}
  \label{fig:three-philosophies}
\end{figure}

This split is separable from the cascaded-versus-joint distinction introduced in Section~\ref{sec:vwm-shift}. Cascaded-versus-joint asks how prediction and action are arranged inside the model. The philosophy-level taxonomy asks where the action prediction is grounded along the inference path: at the rendered pixel output for Render-and-Decode, at an intermediate latent or feature for Latent-Only, or outside the video-generation path for Video-Generation-Free. A Latent-Only WAM can be either cascaded or joint, and the same holds for a Render-and-Decode WAM. Figure~\ref{fig:three-philosophies} sketches the inference data flow under each philosophy.

\begin{figure}[t]
  \centering
  \includegraphics[width=\linewidth]{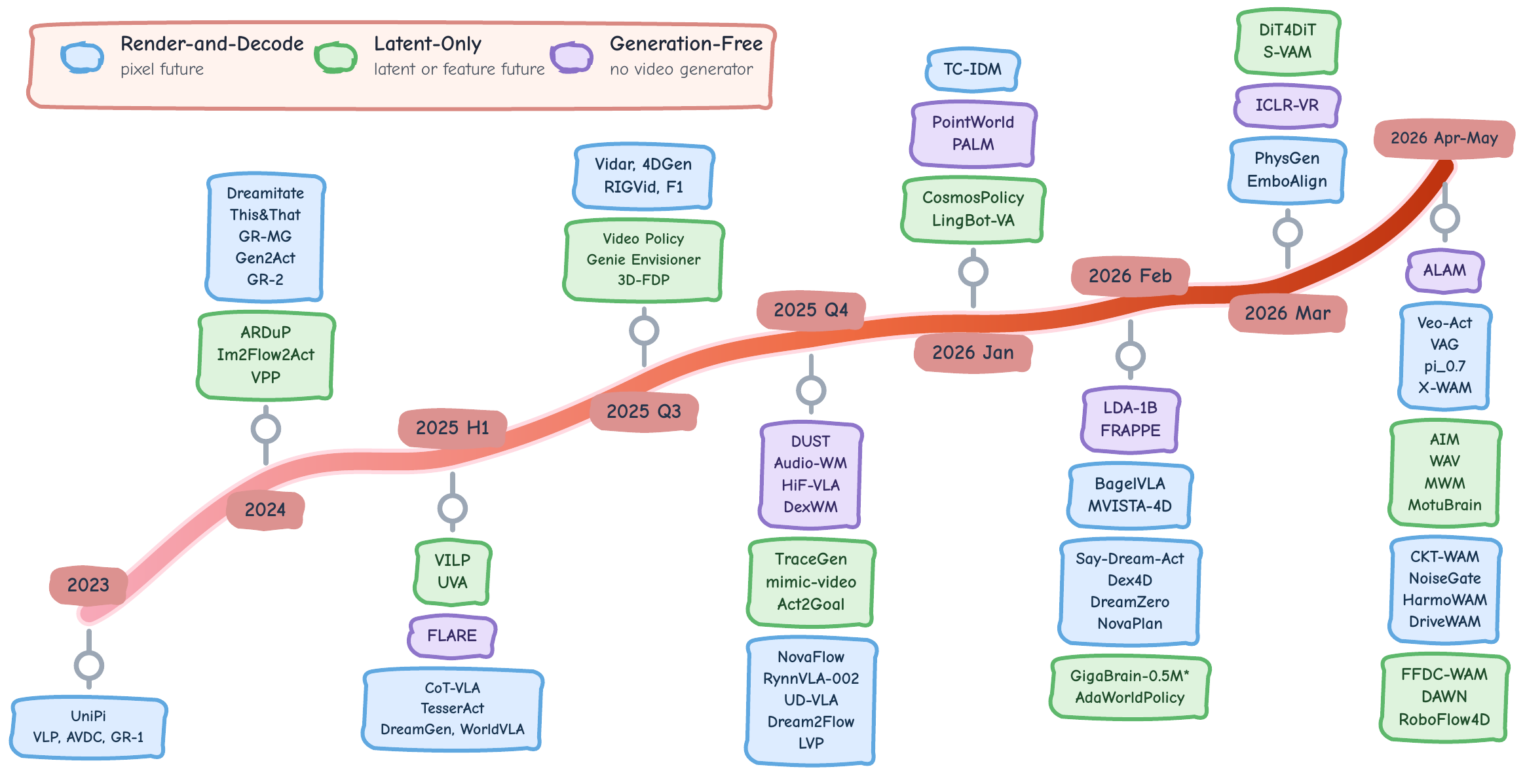}
  \caption{Chronological stream of representative WAMs grouped by design philosophy. The timeline uses coarse bins for 2023, 2024, and 2025 H1, then finer bins for the denser late-2025 and early-2026 period through May. Each date marker attaches a compact vertical stack to the curve, alternating above and below the stream. Dense periods retain early or central verified WAMs so labels remain legible. Stacks above the curve place the widest block nearest the stream, while stacks below the curve begin with the narrowest block. Render-and-Decode appears earliest, Latent-Only emerges as the field starts to drop pixel decoding from the control path, and Video-Generation-Free entries appear as LLM, VLM, JEPA, and non-video diffusion or hybrid backbones begin to carry the predictive component without a video-generation core.}
  \label{fig:wam-timeline}
\end{figure}

Figure~\ref{fig:wam-timeline} adds the chronological dimension that the static taxonomy omits. Early WAMs largely adopt the direct path from video generation to control: generate a visual future, then decode action from it. Later Latent-Only methods preserve a video-derived prior but move the action signal earlier in the inference path, before pixel decoding. Video-Generation-Free methods appear as a more recent alternative, replacing the video generator with predictive supervision in the embedding or token space of LLM, VLM, JEPA, or non-video diffusion or hybrid backbones. Over the same period, the application domain has also broadened. Recent entries now cover tabletop manipulation, dexterous hand control, contact-rich tactile interaction, autonomous driving, and aerial manipulation. They instantiate the same three philosophies under substantially different latency requirements and observation regimes. With this chronological view in place, the remainder of this section explains how each philosophy instantiates this design choice across the WAM census. Section~\ref{sec:what-makes-wam} then gives the component-level anatomy behind these choices through substrate, backbone family, action coupling, and deployment regime.

\subsection{Render-and-Decode World Action Models}
\label{sec:phil-explicit}

Render-and-Decode is the most direct path from video generation to action. Its premise is that a rendered future is worth producing inside the inference-time control loop because it preserves the full visual prior learned by the video backbone. By running that backbone all the way to pixel output, these methods expose appearance, motion, contact, and scene dynamics to the actor in an inspectable form. The inference pattern is simple: condition a video generator on observation history and instruction, produce a future video, and decode the next action from that rendered future. The output may be raw RGB, a multi-view or 4D RGB-D stream, or a sequence of visual tokens that decode back to images, while the action decoder may be inverse dynamics, 6-DoF tracking, dense correspondences, or a dedicated action head. This premise gives Render-and-Decode its conceptual clarity, but also makes visual synthesis part of the policy's latency budget.

The founding branch treats rendered video itself as the plan. UniPi~\citep{unipi} establishes the template with text-conditioned video followed by inverse dynamics, VLP~\citep{du2023vlp} adds search over language-level plans, and AVDC~\citep{ko2024avdc} shows that dense correspondences can replace action labels. The same interface then moves from simulated or task video to demonstrations with different sources of embodiment evidence. Dreamitate~\citep{liang2024dreamitate} tracks generated stereo video, Gen2Act~\citep{bharadhwaj2024gen2act} conditions policy learning on generated human motion, This\&That~\citep{wang2024thisthat} adds gesture coordinates, and GR-MG~\citep{grmg} turns a progress-aware goal image into a manipulation condition. Later variants refine how the rendered plan becomes executable. CreFlow~\citep{creflowcorrectivereflow2026} post-trains the video-as-policy generator with violation masks from a temporal-logic monitor, while MoLA~\citep{imaginedfuturesexecutable2026} maps a frozen SVD rollout into semantic, depth, and flow latent-action codebooks before action decoding. The shared lesson is that the rendered future can be a common planning currency even when the recovery module changes.

Modular Render-and-Decode WAMs make that separation explicit. DreamGen~\citep{DreamGen} uses generated robot futures to obtain pseudo-actions before downstream policy training, so the coupling is offline rather than inference-time imagination. 4DGen~\citep{liu2025fourdgen}, RIGVid~\citep{patel2025rigvid}, TC-IDM~\citep{tcidmgrounding2026}, and GraspDreamer~\citep{graspyoudream2026} keep the world model and action recovery separate through pose tracking, VLM filtering, tool-centric point-cloud conversion, or human-grasp retargeting. VERA~\citep{turningvideomodels2026} makes the split even sharper by leaving the video planner action-free and training a Jacobian inverse-dynamics translator. These works are useful because the world model can be reused across embodiments, but the downstream module must absorb the gap between visual plausibility and robot executability.

A related modular pattern inserts geometric or search structure between the generated future and the action. NovaFlow~\citep{li2025novaflow} and Dream2Flow~\citep{dharmarajan2025dream2flow} distill generated videos into 3D object flow, while EmboAlign~\citep{emboalignaligningvideo2026} filters rollouts with compositional constraints before trajectory optimization. NovaPlan~\citep{novaplanzeroshot2026} places generated video inside a closed-loop manipulation planner, 3D-ALP~\citep{3danchoredlookahead2026} uses a 3D world model as a rollout oracle during test-time search, and AeroPlace-Flow~\citep{aeroplaceflowlanguage2026} grounds a task-complete image into object flow for aerial placement. The pixel future remains the first prediction, but control increasingly depends on the structured object recovered from it.

Joint Render-and-Decode WAMs move action prediction inside the video backbone rather than attaching a separate decoder afterwards. GR-1~\citep{GR-1} is the anchor for this line, because future image tokens and action tokens are learned in one autoregressive stream. GR-2~\citep{GR2} scales the same video-language-action recipe to a larger pretraining and robot-trajectory mixture. WorldVLA~\citep{WorldVLA} adds masking to reduce action leakage while still predicting future VQ images, RynnVLA-002~\citep{rynnvla2} refines actions through future image-state prediction, PhysGen~\citep{PhysGen} adapts a continuous autoregressive video backbone, and DriveWAM~\citep{drivewamvideogenerative2026} interleaves driving video latents with ego-action tokens on a shared DiT. This group changes the coupling pattern, but the action token is still grounded by a pixel-decodable future.

Diffusion and hybrid variants make the same joint target less sequential. PAD~\citep{PAD} jointly denoises future images and actions in one DiT. UD-VLA~\citep{udvla1}, CoVAR~\citep{CoVAR}, and VideoVLA~\citep{VideoVLA} vary the bridge through discrete diffusion, cross-stream attention, or video-generator co-training. Motus~\citep{Motus} is placed here by its joint-prediction path, where sparse future video and actions are denoised together. Its scheduler can switch to action-only or inverse-dynamics use cases, but those shortcuts do not change the category assignment. These works keep the action and the future in the same generative state, but make the scheduling less rigid than autoregression.

Other joint variants expand what the generated state contains. ActionImages~\citep{actionimagesend2026} encodes actions as multi-view RGB heatmaps so the observation and action videos share a generated channel grid. DriveVA~\citep{drivevavideoaction2026} applies flow matching to driving video latents and action tokens, HarmoWAM~\citep{harmowamharmonizinggeneralizable2026} sends rendered frames and latent features to different action experts, and NoiseGate~\citep{noisegatelearningper2026} learns per-latent denoising schedules around a frozen joint backbone. The family therefore grows through stronger coupling, not through removing the rendering obligation.

Other Render-and-Decode methods change what is rendered so that the future carries more task structure than RGB alone. TesserAct~\citep{TesserAct1} generates RGB-D-Normal 4D video, MVISTA-4D~\citep{wang2026mvista4d} infers arbitrary-view RGB-D through test-time latent optimization, and X-WAM~\citep{guo2026xwam} combines multi-view RGB-D prediction with explicit 3D reconstruction. Dex4D~\citep{dex4dtaskagnostic2026} lifts generated frames into 3D object point tracks for dexterous control. DriveDreamer-Policy~\citep{drivedreamerpolicygeometry2026} threads depth, video, and action through modular experts, while MV-VDP~\citep{multiviewvideo2026} diffuses future multi-view RGB together with end-effector heatmap videos. These methods keep the visual future, but they make geometry or affordance structure easier for the action side to extract.

A parallel sparse-future line keeps the output pixel-grounded while reducing how much video must be produced. RoboEnvision~\citep{yang2025roboenvision} uses instruction-aligned keyframes as long-horizon plans. CoT-VLA~\citep{CoTVLA} is the influential VLA-side example, because it generates a visual chain-of-thought before action prediction rather than relying on hidden language reasoning alone. pi0.7~\citep{pi0.7} follows the same pixel-grounded logic: a BAGEL-class generator refreshes multi-view subgoal images asynchronously, and a PaliGemma action expert consumes the latest available goals. BagelVLA~\citep{bagelvlaenhancinglong2026} interleaves subtask language, future keyframes, and action chunks, while SWEET~\citep{sweetsparseworld2026} feeds sparse edited keyframes to a goal-conditioned diffusion policy. Here the design aim is not to avoid pixels, but to render only the visual milestones that the actor can use.

Foundation-scale Render-and-Decode WAMs focus on whether a large video prior can be made responsive enough for closed-loop control. DreamZero~\citep{DreamZero} is the anchor: it turns a Wan-class autoregressive video diffusion backbone into an online video-action policy, uses chunked prediction with observation replacement in the KV cache, and shows that a large rendered-future WAM can stay inside a real control loop. Say-Dream-Act~\citep{gu2026saydreamact} follows the same practicality question from the distillation side, compressing Cosmos-style imagination into a few-step generator before action fusion.

Other foundation video planners keep the rendered future as a planning object and differ mainly in how they ground it. Vidar~\citep{vidar}, LVP~\citep{chen2025lvp}, and Veo-Act~\citep{zhang2026veoact} use masked inverse dynamics, Wan-style image-to-video generation, or a precision VLA executor. VAG~\citep{lang2026vag} and F1~\citep{f1vla} attach action experts to generated or pooled foresight, while EVA~\citep{evaaligningvideo2026} aligns the video planner with an inverse-dynamics reward so the rendered future stays executable.

The newest foundation-scale variants spend the video prior more selectively. CKT-WAM~\citep{cktwamparameter2026} transfers context between heterogeneous WAMs, DreamAvoid~\citep{dreamavoidcriticalphase2026} invokes video imagination only on critical phases, and Pelican-Unify 1.0~\citep{pelicanunify12026} links VLM reasoning to a Wan-style future generator. This group keeps the rendered future available when it carries task information, while shifting routine action selection toward cheaper context transfer, critical-phase triggering, or grounded VLM guidance.

The defining limitation of Render-and-Decode is the price of producing pixels. Each prediction step pays for the full denoising or autoregression schedule, even though the actor rarely consumes the rendered output in its entirety. Moreover, visual-quality metrics inherited from video generation only weakly predict downstream task success. This practical tension links the category closely to the open-loop and chunked closed-loop regimes unpacked in Section~\ref{sec:what-deploy}.

\subsection{Latent-Only World Action Models}
\label{sec:phil-latent}

The Latent-Only philosophy keeps the video-world-model prior of Render-and-Decode but removes pixel decoding from the inference-time control path. Its motivation is to retain temporal and physical structure learned from video while reducing inference cost. The action signal is decoded directly from a latent, an intermediate denoising feature, a flow field, a semantic mask, a value map, or another video-derived feature future, each of which is cheaper to produce than a fully rendered video. The trade-off is that these methods give up the pixel-space supervision and visual-quality metrics that make Render-and-Decode easy to inspect. The designs differ mainly in where they intercept the predictive path: some consume the model's own video latent, some tap an intermediate denoising state, some predict a structured substrate such as object flow or masks, and some inherit a joint video-action backbone but disable pixel decoding at inference. In all cases, the key move is to keep a video-shaped dynamic prior while refusing to pay for the final renderer during action selection.

The earliest Latent-Only entries make the interception point explicit. ARDuP~\citep{huang2024ardup} focuses latent video diffusion on active regions, VPP~\citep{hu2024vpp} conditions inverse dynamics on Stable Video Diffusion representations, and VILP~\citep{xu2025vilp} recovers actions from time-aligned multi-view latent video. UVA~\citep{UVA} and Video Policy~\citep{liang2025videopolicy} show the two main training patterns: either share a video-action latent across heads, or freeze the video generator and train the policy on its intermediate features. Genie Envisioner~\citep{liao2025genieenvisioner} is the clearest bridge between these patterns. GE-Base remains an instruction-conditioned multi-view video DiT, while GE-Act decodes action chunks from a one-step denoised latent cache through flow matching. Act2Goal~\citep{Act2Goal} reuses the same foundation for goal-conditioned transition features, and WAV~\citep{WAV} performs value-guided latent trajectory inference before action decoding. These methods show that the first shortcut is to stop at the representation that action already needs.

As video diffusion backbones scale, Latent-Only designs increasingly intercept the denoising trajectory rather than a finished latent. mimic-video~\citep{pai2025mimicvideo} attaches an action decoder to an intermediate ODE checkpoint of Cosmos-Predict2, and DiT4DiT~\citep{DiT4DiT} conditions its action expert on intermediate Cosmos-Predict2.5 features in a deterministic step. S-VAM~\citep{yan2026svam} self-distills multi-step denoising into one pass over geometric and semantic features. Fast-WAM~\citep{FastWAM} and GigaWorld-Policy~\citep{gigaworld-policy} keep future-video co-training but make full generation optional or unnecessary at inference. LaMP~\citep{lamplearningvision2026} exposes a one-step 3D motion hidden state to a VLA through gated cross-attention, and VAMPO~\citep{vampopolicyoptimization2026} tunes early-step VPM latents against expert visual dynamics. The practical shift is from pixel synthesis to feature extraction.

Another branch predicts structured motion before action decoding. Im2Flow2Act~\citep{xu2024im2flow2act} and 3DFlowAction~\citep{zhi2025threedflowaction} replace appearance with object flow, while 3D-FDP~\citep{3dflowdiffusion2025} predicts dense 3D flow over query points before a second action diffusion model runs. TraceGen~\citep{tracegenworldmodeling2025} predicts 3D traces that inverse kinematics converts into joint commands. This flow-and-trace group compresses the future toward motion rather than texture.

The structured-substrate branch then broadens beyond flow. 3PoinTr~\citep{3pointr3dpoint2026} and RoboFlow4D~\citep{roboflow4dlightweightflow2026} freeze future point-track or 4D-flow predictors and train downstream policies to consume the predicted motion. MWM~\citep{lou2026mwm} keeps a video-diffusion architecture but predicts semantic-mask evolution, OmniVTA~\citep{zheng2026omnivta} adds tactile latents for contact-rich control, and EgoExo-WM~\citep{egoexowmunlocking2026} uses a DINOv3 latent regressor under model-predictive control after exo-to-ego data conversion. This branch makes the future closer to contact and task state than a rendered frame would be.

The clearest Latent-Only inference pattern appears when a joint video-action backbone is trained with visual prediction but used through a latent action path. UWM~\citep{UWM} is the reference case: independent diffusion timesteps let the method collapse the visual branch at inference while retaining the video-trained representation. CosmosPolicy~\citep{CosmosPolicy}, AdaWorldPolicy~\citep{AdaWorldPolicy}, and MotuBrain~\citep{MotuBrain} follow the pixel-latent version of the same idea, where future RGB or video latents remain useful during training but the action call at inference does not require a fully rendered RGB forecast. LingBot-VA~\citep{LingBotVA} exposes semantic structures before pixel reconstruction, and AIM~\citep{AIM} decodes action through a spatial value map rather than directly through future RGB. The design consequence is clear: visual prediction shapes the representation, but action selection does not have to render the visual prediction.

Recent Latent-Only WAMs extend the same idea across modalities and memory designs. GigaBrain-0.5M*~\citep{gigabrain05m2026} feeds the policy with future-state latents and value embeddings from a Wan2.2 world model. VTAM~\citep{vtamvideotactile2026} treats camera views and GelSight tactile input as decoder-bound latent views, while CLWM~\citep{dexworldmodelcausallatent2026} forecasts future DINOv3 feature maps and replaces a growing KV cache with Dual-State Test-Time Training memory.

The same inference logic also travels across domains and joint-state designs. DAWN~\citep{dawnworldaction2026} iteratively refines V-JEPA latent world tokens against DiT action hypotheses, JOPAT~\citep{pointtrackingimproves2026} adds point tracks and visibility to the joint denoising state, and tau0-WM~\citep{tau0wm2026} keeps Wan-style video latents in the joint state but decodes action without routine pixel rendering. WALL-WM~\citep{wallwmcarving2026} pushes the same idea into depth-wise cross-attention between matched video blocks and action queries. These examples make Latent-Only a design philosophy rather than a single substrate choice.

A newer branch uses latent imagination as an adaptive control signal. FFDC-WAM~\citep{whentrustimagination2026} caches Motus latent video tokens with the proposed action chunk and uses a causal attention verifier to decide when a full WAM replan is needed. tau0-WM~\citep{tau0wm2026} can also reactivate its action-conditioned video simulator only when candidate scoring is useful. This turns imagination into a learned controller decision: the policy invokes the heavier future model only when confidence drops.

Across this family, the central tension is clear. Video pretraining supplies a valuable dynamic prior, but full pixel rendering imposes latency and memory costs that are difficult to justify for every control step. Latent-Only methods preserve the prior while bypassing the renderer, enabling several members of the family to reach real-time inference at the cost of reduced direct visual interpretability.

\subsection{Video-Generation-Free World Action Models}
\label{sec:phil-genfree}

The Video-Generation-Free philosophy removes the video-generation backbone altogether. Here \emph{video-generation-free} refers strictly to the absence of a pixel-level video backbone in the predictive path. The LLM, VLM, JEPA, and flow-matching components below remain generative models in the broader sense. Its motivation is to avoid the training and inference cost of video generation while retaining predictive supervision in a more compact representational space. These methods instead leverage large language models, vision-language models, joint-embedding-predictive encoders, deterministic regressors over frozen vision foundation models, or non-video diffusion and hybrid backbones over a compact non-pixel substrate. The dominant pattern attaches a lightweight action expert to a VLM, learns a latent-action vocabulary that bridges actionless video and robot control, or supervises a policy with an auxiliary feature-prediction loss in a vision-foundation embedding space. The family is listed in Table~\ref{tab:wams-genfree}, where it complements the Render-and-Decode and Latent-Only entries in Table~\ref{tab:wams-video}.

The feature-prediction line replaces pixel prediction with teacher-target or foundation-feature prediction. FLARE~\citep{FLARE} is the anchor: additional policy tokens are aligned to future-observation embeddings from a frozen teacher, and the action expert consumes those predicted future tokens at inference. FRAPPE~\citep{FRAPPE} extends the idea by aligning a generalist policy to future representations from multiple vision foundation models, while DexWM~\citep{DexWM} applies deterministic future-feature regression to dexterous hand-object interaction. LDA-1B~\citep{LDA1B} moves the same principle into a Qwen3-VL stack by jointly learning dynamics, policy, and forecasting in a structured DINO latent. These works keep the predictive training signal, but place it in an embedding space whose quality is judged through action rather than photorealism.

Another Video-Generation-Free line treats the future as a learned token or latent transition. DUST~\citep{DUST} augments an Eagle-2 VLM with a dual-stream diffusion head that predicts future observation embedding tokens and an action together without a video-generation foundation. ALAM~\citep{alamalgebraicallyconsistent2026} learns latent transitions from action-free video and regularizes them with composition and reversal consistency before a flow-matching action expert consumes the codes. The shared move is to keep a future object in the inference path while choosing a representation that is cheaper than video.

Structured non-pixel futures give the family its second anchor. PointWorld~\citep{pointworldscaling3d2026} predicts action-conditioned 3D point flow through a deterministic Point Transformer and plans with MPPI over the compact future. PALM~\citep{palmprogressaware2026} structures foresight as affordance maps plus progress, while HiF-VLA~\citep{hifvlahindsight2025} uses compact motion-vector foresight tokens and ICLR-VR~\citep{iclrcontextimitation2026} predicts a gripper-keypoint polyline before each action chunk. These works replace video with geometry, affordance, or trace-like variables that are closer to the control interface.

The remaining generation-free variants broaden the same principle across modalities and feedback. Audio-WM~\citep{learningrobotmanipulation2025} predicts future Mel latents that condition a robot policy when contact sound carries task evidence. DDP~\citep{dreamingunseenworld2026} switches to latent imagination when a real-imagination gap indicates an out-of-distribution event, and Feedback-WM~\citep{feedbackworldmodel2026} treats a learned latent transition model as an observer that corrects diffusion-policy denoising online. Together, these methods show what is gained by dropping the video generator entirely, and what must be replaced by tokens, features, audio latents, affordances, or geometry.

\subsection{Where the three philosophies meet the formal anatomy}
\label{sec:phil-handoff}

The three philosophies answer the question of \emph{where} action prediction sits along the video-generation inference path: at the rendered pixel output for Render-and-Decode, at an intermediate latent or feature for Latent-Only, or outside the video-generation path for Video-Generation-Free. They do not yet answer \emph{how} the model produces that prediction, where action is injected, or when the model is invoked relative to the control loop. Tables~\ref{tab:wams-video} and~\ref{tab:wams-genfree} place every WAM in the same census under a four-axis design space that captures these component-level choices. Section~\ref{sec:what-makes-wam} introduces the formal notation behind that four-axis view and uses it to compare WAMs across substrate, backbone family, action coupling, and deployment regime. Together, the philosophy-level taxonomy of this section and the component-level anatomy of the next provide two complementary viewpoints for navigating the field.

%% file: sections/04-what-makes-a-wam/what-makes-a-wam.tex
\section{What Makes a World Action Model}
\label{sec:what-makes-wam}

\begin{figure}[t]
  \centering
  \includegraphics[width=\linewidth]{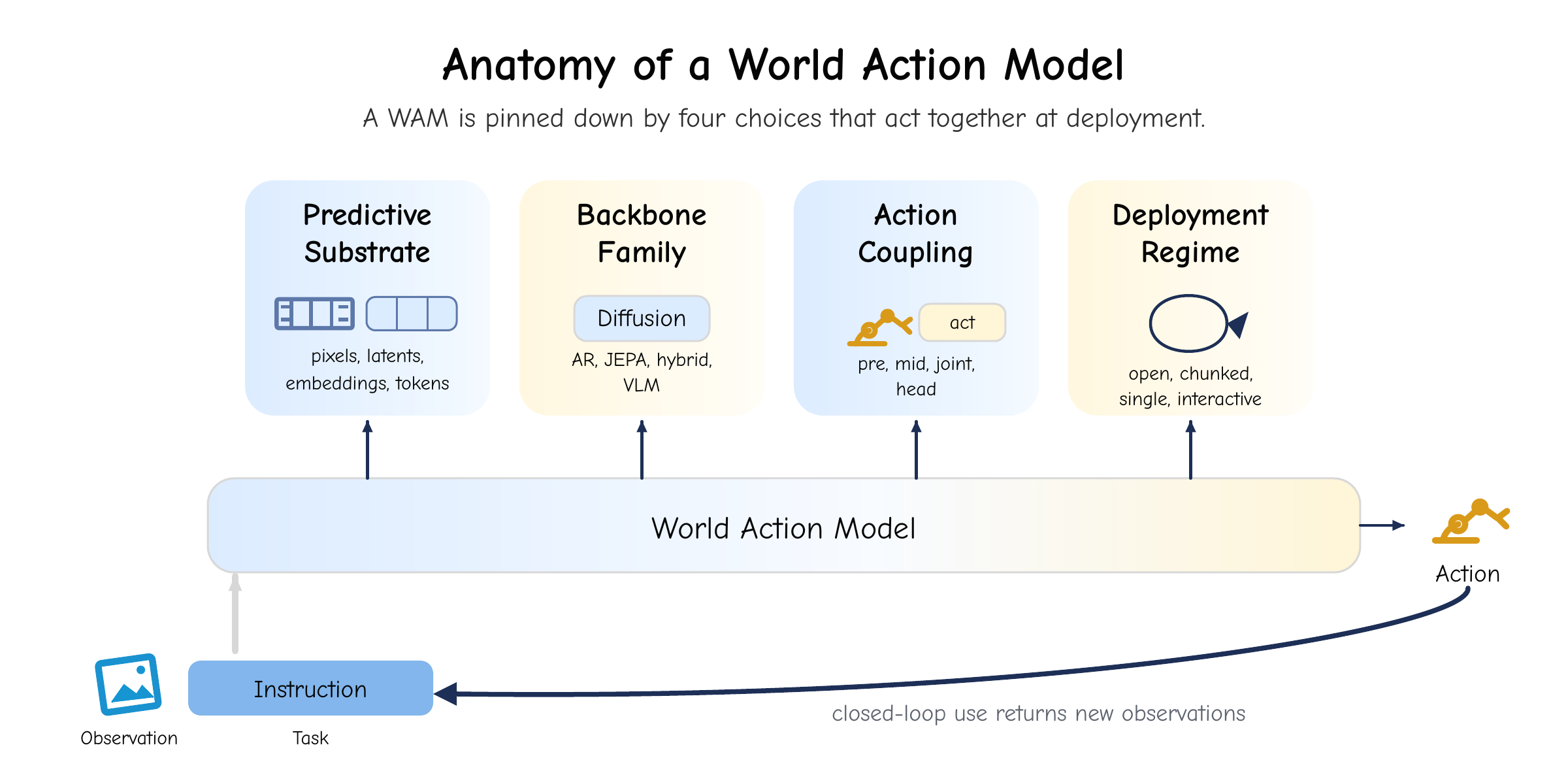}
  \caption{The anatomy of a world action model. Any existing WAM can be specified by four separable but interacting choices: what it predicts (Section~\ref{sec:what-substrate}), how action is coupled to that prediction (Section~\ref{sec:what-coupling}), the function family that produces the prediction (Section~\ref{sec:what-backbone}), and the deployment regime it is intended for (Section~\ref{sec:what-deploy}). Two methods with the same labels but different choices on any one axis behave very differently in practice.}
  \label{fig:wam-anatomy}
\end{figure}

The term \emph{world action model} now spans works built in different communities, with different vocabularies, and on top of different backbones. A paper-by-paper enumeration is insufficient for comparing them.
Methods that share the WAM label can differ in nearly every implementation detail, whereas closely related methods may use entirely different terminology.
This section takes a different route.
It treats WAMs as instances of a common mathematical object: a conditional joint distribution over future predictions and future actions. Under this view, the main design choices specify what future is represented, how actions are coupled to that future, what model family produces the prediction, and how the resulting model is invoked during control.
We refer to these choices as the \emph{predictive substrate}, \emph{action coupling}, \emph{architectural backbone}, and \emph{deployment regime}, respectively.
Figure~\ref{fig:wam-anatomy} previews how the four ingredients combine. Section~\ref{sec:what-notation} introduces the shared notation. Sections~\ref{sec:what-substrate} through~\ref{sec:what-deploy} then walk through the four ingredients one by one. They sort existing methods along the choices each ingredient offers and note the practical trade-off that each choice carries.

\subsection{A Unified Notation for World Action Models}
\label{sec:what-notation}

At the right level of generality, WAMs can be described by a common abstract object. The object is a parameterized conditional joint distribution over a window of future predictions and a window of future actions:
\begin{equation}
p_\Theta\!\left(\,s_{t+1:t+H},\ a_{t:t+H-1}\ \middle|\ o_{\le t},\ a_{<t},\ l\,\right).
\label{eq:wam-joint}
\end{equation}
Here $o_{\le t} = (o_0,\dots,o_t)$ is the observation history, with each $o_i \in \mathcal{O}$ an observation, such as a frame, video chunk, or multimodal sensory packet, $a_{<t} = (a_0,\dots,a_{t-1})$ is the past action history with $a_i \in \mathcal{A}$, and $l$ is a task instruction in language, a goal image, or another auxiliary modality. The model predicts a future trajectory $s_{t+1:t+H}$ of length $H$ in a chosen \emph{substrate space} $\mathcal{S}$ together with a future action chunk $a_{t:t+H-1}$ of the same length. The parameter collection $\Theta$ may be one shared backbone or a retained world-model module paired with an action module. To keep later equations compact, we abbreviate the full conditioning context as
\begin{equation}
c \;\equiv\; (\,o_{\le t},\ a_{<t},\ l\,),
\label{eq:context}
\end{equation}
and write Equation~\ref{eq:wam-joint} as $p_\Theta(s_{t+1:t+H},\,a_{t:t+H-1}\mid c)$ whenever no ambiguity arises.

The substrate space $\mathcal{S}$ is the space of the future representation that the WAM exposes to action production or action evaluation. We write it as the image of an interface encoder $\phi:\mathcal{O}\to\mathcal{S}$, but $\phi$ should not be taken as the generator's internal implementation. A pixel-grounded WAM exposes decoded future frames, or the VAE / VQ latent grid that the generator decodes to a clip, even if the action side never invokes the pixel decode at inference. A feature WAM exposes a learned hidden state or token block with no decodable visual interpretation, whether it is a frozen self-supervised teacher's target, a feature tap of a generative trunk, or a VLM future-token block. A geometric-primitive WAM exposes a structured object whose dimensions are physical coordinates, such as flow, point clouds, depth, pose, or motion vectors. An affordance-map WAM exposes a task-relevant label or score map, such as a value, affordance, or end-effector heatmap. The symbol $p_\Theta$ in Equation~\ref{eq:wam-joint} is a placeholder for an abstract joint, not necessarily one monolithic network. Section~\ref{sec:what-backbone} lists the five concrete families that realize its future-prediction factor in practice, each with its own parameterization. Until that point, $p_\Theta$ should be treated as the action-facing future contract that every WAM must satisfy.

The four ingredients of this section are choices about how Equation~\ref{eq:wam-joint} is realized in practice.

\paragraph{(i) Predictive substrate.} A choice of action-facing future representation, and therefore of the space $\mathcal{S}$ in which $s_{t+1:t+H}$ lives. Section~\ref{sec:what-substrate}.

\paragraph{(ii) Action coupling.} A choice of factorization of the joint in Equation~\ref{eq:wam-joint}. Three top-level factorization families appear across existing WAMs and return in Section~\ref{sec:what-coupling}:
\begin{align}
\text{action-conditioned rollout}\ &:
\begin{cases}
q_\psi\!\left(a_{t:t+H-1}\mid c\right)\,p_\theta\!\left(s_{t+1:t+H}\,\middle|\,c,\,a_{t:t+H-1}\right), & \text{chunk-level},\\
\prod_{k=0}^{H-1} q_\psi\!\left(a_{t+k}\mid h_k,c\right)\,p_\theta\!\left(s_{t+k+1}\,\middle|\,h_k,\,a_{t+k},\,c\right), & \text{step-wise},
\end{cases}
\label{eq:fact-rollout}\\
\text{joint generation}\ &:\ p_\theta\!\left(s_{t+1:t+H},\,a_{t:t+H-1}\,\middle|\,c\right), \label{eq:fact-joint}\\
\text{post-prediction head}\ &:\ p_\theta\!\left(s_{t+1:t+H}\,\middle|\,c\right)\;q_\psi\!\left(a_{t:t+H-1}\,\middle|\,s_{t+1:t+H},\,c\right). \label{eq:fact-head}
\end{align}
where $h_k=(s_{t+1:t+k},a_{t:t+k-1})$ in the step-wise submode. Action-conditioned rollout composes an action source with a world predictor, either by fixing an action chunk before prediction or by updating the action after each predicted step. Joint generation produces the future substrate and the future action chunk from one coupled generative model $p_\theta(s_{t+1:t+H}, a_{t:t+H-1}\mid c)$, where $\theta$ may include a shared trunk as well as modality-specific heads or experts
The post-prediction head factorization predicts the substrate first and then decodes action from it with a smaller actor $q_\psi$, which is often a tracker, inverse-dynamics model, optimizer, planner, or policy that can be trained separately from $\theta$. All three families carry the past action history through the abbreviation $c$, so the conditioning sides remain consistent with Equation~\ref{eq:wam-joint}.

\paragraph{(iii) Architectural backbone.} A choice of function family that realizes the future-prediction factor of $p_\Theta$, namely an iterative denoising network, an autoregressive decoder, a joint-embedding predictor, a hybrid that combines two of the above, or a large language or vision-language model with an attached action decoder. Each family carries its own parameterization, which we introduce only when the family is treated. Section~\ref{sec:what-backbone}.

\paragraph{(iv) Deployment regime.} A choice of when the WAM is invoked and over what window. Open-loop rollout invokes the WAM once with $H\!\approx\!T$ and specifies the whole window before execution. Chunked closed-loop invokes the WAM every $K$ control steps with $H\!=\!K$ and replans after each chunk is executed. Single-step closed-loop invokes the WAM at every control step with $H\!=\!1$. Interactive operation typically reuses a trained model $\theta$ and extends $H$ indefinitely while carrying state across calls through a key-value cache or a persistent latent. Section~\ref{sec:what-deploy}.

This decomposition is useful not because the four axes are separable, but because they separate different questions about a WAM. The substrate asks what kind of future variable $s$ is represented, the backbone asks how that variable is predicted, the coupling asks how actions enter the prediction or are recovered from it, and the deployment regime asks how the resulting model is invoked inside a control loop. In practice, the substrate and backbone are usually fixed by training, while deployment and some wrapper-level coupling choices can be adjusted at inference. Changing the coupling family itself, however, often requires an additional action head, planner, or fine-tuning stage. We use these four questions as the organizing lens for the subsections below.

Tables~\ref{tab:wams-video} and~\ref{tab:wams-genfree} place every WAM in our census as a 4-tuple over these choices, and the subsections that follow walk through the choices one by one and anchor each one in the same equation.

\subsection{Predictive Substrate: Where WAMs Dream}
\label{sec:what-substrate}

\noindent
The first choice in Equation~\ref{eq:wam-joint} is where the future variable $s_{t+1:t+H}$ lives. We call this space the predictive substrate. The substrate is not determined by the backbone alone, nor by the last tensor consumed by the action head. For example, a video-diffusion model may denoise a VAE grid and skip pixel decoding during control. As long as a fixed decoder maps that grid back to video, the predicted future remains pixel-grounded. Conversely, a policy may consume an intermediate state from a video-trained trunk, but that state is a feature substrate if it has no fixed observation decoder. In this section, we therefore classify a WAM by the representation in which it forms the future used for action prediction or evaluation.

We use four substrate categories. Pixel-grounded substrates represent future observations, either as decoded RGB, RGB-D, multi-view video, or as VAE / VQ latents with a fixed video decoder. Acoustic observation latents are rare in the current literature, so we treat them as multimodal observation-latent edge cases rather than as a separate category. Feature substrates represent learned states, teacher embeddings, or VLM token blocks without a fixed observation decoder. Geometric substrates represent physical structure or motion, such as optical flow, point tracks, depth, pose, motion vectors, or polylines. Affordance substrates represent task-specific maps or score fields, such as value maps, contact maps, semantic masks, or heatmaps. Some WAMs predict more than one kind of future in the same forward process. The tables mark these joint cases with $\wedge$.

\subsubsection{Pixel-grounded substrates}
\label{sec:sub-pixel}

The pixel-grounded category is the direct descendant of video generation. Its decoded form is an observation tensor,
\begin{equation}
s_{t+1:t+H} \in \mathbb{R}^{H \times C_\text{o} \times H_\text{px} \times W_\text{px}},
\label{eq:sub-pixel}
\end{equation}
where $C_\text{o}$ is the observation channel count. A decoder-bound latent variant uses the same category when $s$ is a VAE, VQ, or 3D-VAE code whose fixed decoder maps it back to video. The model may act before that decoder is called, but the future still lives in an observation coordinate frame. This criterion separates decoder-bound observation latents from feature latents below.

\textbf{Decoded observation futures.} Early WAMs take the direct route: generate a future image or video, then decode action from it. UniPi~\citep{unipi} establishes this pattern with text-conditioned video plus inverse dynamics, while AVDC~\citep{ko2024avdc} shows that dense correspondences from the decoded video can replace action labels. The same substrate category covers cascaded designs when the action module is separate. DreamGen~\citep{DreamGen} uses decoded robot futures to obtain pseudo-actions before policy training, RIGVid~\citep{patel2025rigvid} retargets tracked generated demonstrations, and VERA~\citep{turningvideomodels2026} lets a Jacobian inverse-dynamics model act from a generated visual lookahead. Later decoded-observation methods keep the same substrate but broaden the content of the forecast. Vidar~\citep{vidar}, LVP~\citep{chen2025lvp}, and EVA~\citep{evaaligningvideo2026} use stronger Wan-class planners, while Veo-Act~\citep{zhang2026veoact} uses a precision VLA executor on top of a generated visual plan. These wrappers still belong to the pixel-grounded category when geometry is recovered after the visual future has been produced.

\textbf{Pixel-decodable latent futures.} The second route keeps the future in the pixel-decodable latent space of a video generator. GR-1~\citep{GR-1} and GR-2~\citep{GR2} predict VQ image tokens and action tokens in one autoregressive stream. PAD~\citep{PAD}, UWM~\citep{UWM}, WorldVLA~\citep{WorldVLA}, VideoVLA~\citep{VideoVLA}, DriveVA~\citep{drivevavideoaction2026}, and $\tau_0$-WM~\citep{tau0wm2026} instead couple action to a diffusion or flow-matching trajectory over VAE video latents. Hybrid entries such as F1~\citep{f1vla}, Motus~\citep{Motus}, Pelican-Unify~1.0~\citep{pelicanunify12026}, NoiseGate~\citep{noisegatelearningper2026}, and WALL-WM~\citep{wallwmcarving2026} keep the same substrate while changing the routing between video and action streams. CKT-WAM~\citep{cktwamparameter2026} and FFDC-WAM~\citep{whentrustimagination2026} add a teacher context or verifier around that substrate, but the action-facing future remains the student or Motus pixel-decodable visual forecast. These works are not feature-substrate WAMs under the present criterion because the future is still a compressed observation under a fixed observation decoder.

\textbf{Sparse, multimodal, and joint observation futures.} The pixel-grounded category also covers reduced, multi-channel, and rare non-visual observation displays. SWEET~\citep{sweetsparseworld2026} shrinks the forecast to task-critical keyframes. Audio-WM~\citep{learningrobotmanipulation2025} predicts future Mel latents that condition a robot policy, and it is grouped here as an acoustic observation-latent edge case because it follows the same decoder-bound criterion without changing the category name. DriveDreamer-Policy~\citep{drivedreamerpolicygeometry2026} predicts decoded video together with depth, while MV-VDP~\citep{multiviewvideo2026} and ActionImages~\citep{actionimagesend2026} predict RGB together with Gaussian end-effector heatmaps. OmniVTA~\citep{zheng2026omnivta} is also an observation-future case rather than a geometric one, because its future is a pair of visual and tactile decoder-bound latents with fixed decoders, not a coordinate tensor. HarmoWAM~\citep{harmowamharmonizinggeneralizable2026} is a dual observation case because its Wan backbone produces both rendered frames and pixel-decodable latent features that feed different action experts. The common trade-off is clear: observation futures preserve appearance, contact traces, and broad video priors, but they pay for detail that many action decisions do not need.

\subsubsection{Feature substrates}
\label{sec:sub-latent}

The feature category covers futures that are learned states rather than decoder-bound observation states. A single sample can be written as
\begin{equation}
s_{t+1:t+H} \in \mathbb{R}^{H \times N_\text{tok} \times d_\text{emb}},
\label{eq:sub-latent}
\end{equation}
with $N_\text{tok}$ tokens per step and embedding dimension $d_\text{emb}$. The key property is absence of a fixed observation decoder. The state may be trained through future prediction, but it is interpreted only by the action pathway or by an embedding loss.

\textbf{Encoder-only and feature-tap futures.} This variant keeps the future-trained backbone but routes action through an internal state. Fast-WAM~\citep{FastWAM} is the clean encoder-only case: it keeps video co-training but masks the test-time future branch, so action is predicted from a hidden state shaped by future prediction. VPP~\citep{hu2024vpp}, Genie Envisioner~\citep{liao2025genieenvisioner}, VidMan~\citep{VidMan}, mimic-video~\citep{pai2025mimicvideo}, VAG~\citep{lang2026vag}, and WAV~\citep{WAV} keep more of the generative trajectory but decode action from latent or intermediate denoising states. LaMP~\citep{lamplearningvision2026} is the VLM-stack version of the same shortcut, exposing a motion-expert hidden state through gated cross-attention before the action expert fires. The boundary is that an auxiliary hidden state is not enough to make the substrate a feature future when the inference-time action-facing future is still a pixel-decodable visual forecast.

\textbf{Teacher-target futures.} A second feature route predicts the embedding of a frozen vision encoder, a multimodal encoder, or a policy/expert latent state. FLARE~\citep{FLARE} aligns policy tokens to future-observation embeddings, and LDA-1B~\citep{LDA1B} learns dynamics, policy, and forecasting in a structured DINO feature space on a Qwen3-VL backbone. DAWN~\citep{dawnworldaction2026}, CLWM~\citep{dexworldmodelcausallatent2026}, EgoExo-WM~\citep{egoexowmunlocking2026}, DexWM~\citep{DexWM}, DDP~\citep{dreamingunseenworld2026}, Feedback-WM~\citep{feedbackworldmodel2026}, and FRAPPE~\citep{FRAPPE} all keep the future in a learned feature space rather than in a decoder-bound observation latent. The benefit is compact prediction and strong semantic invariance. The cost is that the substrate has no direct visual-fidelity metric.

\textbf{VLM-token futures.} The third feature route represents the future as tokens in a VLM vocabulary. CoT-VLA~\citep{CoTVLA} predicts visual chain-of-thought tokens before action generation. DUST~\citep{DUST} jointly diffuses a next-state observation token and an action token, and ALAM~\citep{alamalgebraicallyconsistent2026} interleaves view-specific latent-transition tokens with action tokens under a structured attention mask. These methods use tokens as future carriers, not merely as action discretisation. The feature category is therefore broad, but its practical boundary is whether the predicted future lacks a fixed observation decoder and is not a physical coordinate tensor or task map.

\subsubsection{Geometric primitive substrates}
\label{sec:sub-geom}

The third category exposes a structured object whose dimensions are physical coordinates. A common instance is dense optical flow, for which $\phi$ is an off-the-shelf flow operator and a single sample is
\begin{equation}
s_{t+1:t+H} \in \mathbb{R}^{H \times 2 \times H_\text{px} \times W_\text{px}},
\label{eq:sub-geom}
\end{equation}
namely two channels carrying the $(u,v)$ displacement field. Variants replace the flow field with $N_\text{kp}$ tracked keypoints in $\mathbb{R}^{H \times N_\text{kp} \times 3}$, a depth tensor, a 6-DoF pose stream, a sparse waypoint polyline, or a macroblock motion-vector grid. A method belongs here when geometry is the predicted future representation, not when a tracker merely measures geometry after a pixel-grounded future has already been chosen.

\textbf{Flow and point-motion futures.} Im2Flow2Act~\citep{xu2024im2flow2act} and 3DFlowAction~\citep{zhi2025threedflowaction} motivate the category by replacing full appearance with object motion. NovaFlow~\citep{li2025novaflow} and Dream2Flow~\citep{dharmarajan2025dream2flow} take the cascaded route by extracting 3D object flow from generated videos before planning. TraceGen~\citep{tracegenworldmodeling2025} predicts 3D traces, while 3PoinTr~\citep{3pointr3dpoint2026} and RoboFlow4D~\citep{roboflow4dlightweightflow2026} feed separately predicted point tracks or flow into downstream policies. 3D-FDP~\citep{3dflowdiffusion2025}, PointWorld~\citep{pointworldscaling3d2026}, and Dex4D~\citep{dex4dtaskagnostic2026} extend this idea from dense image flow to sparse or scene-level 3D point trajectories. The common payoff is a future that is closer to control than RGB.

\textbf{Structured physical futures.} TesserAct~\citep{TesserAct1} reconstructs a 4D point cloud as the control-facing representation. HiF-VLA~\citep{hifvlahindsight2025} uses MPEG-4 motion vectors as a compact future-motion field, and ICLR-VR~\citep{iclrcontextimitation2026} reduces the future to a five-point gripper polyline. JOPAT~\citep{pointtrackingimproves2026} is one prominent joint case because it denoises a pixel-decodable visual future together with point tracks and visibility. Geometric futures are cheaper and often easier to supervise than pixel-grounded futures, but they lose appearance cues that are irrelevant to motion yet useful for semantic grounding.

The structured tensor is far smaller than the pixel tensor at the same spatial resolution, and it can often be trained stably on less data. The trade-off is that these predictions are most reliable when motion, contact, or geometry carries the task-relevant information. They are less suitable when appearance or semantics is essential.

\subsubsection{Affordance map substrates}
\label{sec:sub-aff}

The fourth category exposes a task-relevant label or score map. A single sample is
\begin{equation}
s_{t+1:t+H} \in \mathbb{R}^{H \times C \times H_\text{px} \times W_\text{px}},
\label{eq:sub-aff}
\end{equation}
with $C$ channels carrying value, affordance, contact likelihood, segmentation, or progress information. The substrate is not an observation, a feature, or a physical coordinate field. It is a spatial answer to a task-specific question.

\textbf{Value, mask, and affordance futures.} AIM~\citep{AIM} inserts a spatial value map between future visual dynamics and action, with intent-causal attention forcing action to pass through that value representation. PALM~\citep{palmprogressaware2026} predicts a bundle of future affordance maps, including object masks, contact likelihood, placement candidates, and motion regions. MWM~\citep{lou2026mwm} predicts semantic-mask dynamics rather than RGB, so its future is a label map rather than a generic hidden state. ActionImages~\citep{actionimagesend2026} and MV-VDP~\citep{multiviewvideo2026} are joint pixel-grounded-and-affordance cases because they predict RGB together with Gaussian end-effector heatmaps.
Affordance futures are therefore more task-specific than observation or feature futures. They are compact and often directly useful for manipulation, but their labels must be defined for the task at hand.

\subsubsection{Substrate and coupling are distinct axes}
\label{sec:sub-axis-distinction}

\begin{figure}[t]
  \centering
  \makebox[\linewidth][c]{\includegraphics[width=1.08\linewidth]{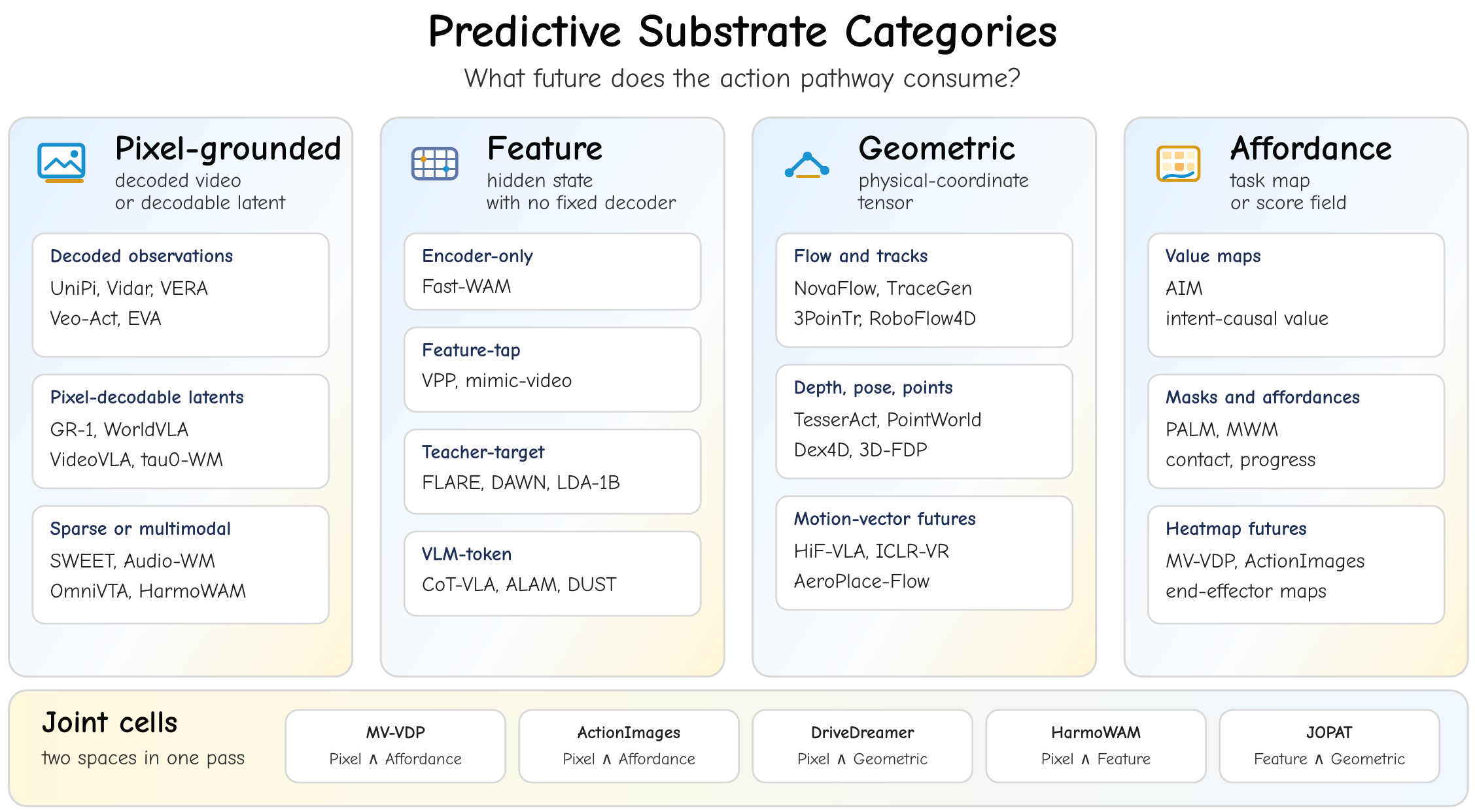}}
  \caption{The four predictive-substrate categories of Section~\ref{sec:what-substrate}, classified by the representational space in which the WAM forms its future. Pixel-grounded substrates include decoded observations and pixel-decodable latents, with Audio-WM treated as a rare acoustic observation-latent edge case (\citealp{unipi,GR-1,WorldVLA,tau0wm2026,learningrobotmanipulation2025}). Feature substrates include encoder-only, feature-tap, teacher-target, and VLM-token futures with no fixed observation decoder (\citealp{FastWAM,hu2024vpp,FLARE,dawnworldaction2026,alamalgebraicallyconsistent2026}). Geometric substrates expose flow, point clouds, depth, polylines, or motion vectors (\citealp{TesserAct1,pointworldscaling3d2026,hifvlahindsight2025}). Affordance substrates expose value maps, masks, affordances, or heatmaps (\citealp{AIM,lou2026mwm,palmprogressaware2026}). Joint cells use the $\wedge$ separator when one forward pass predicts futures in two categories.}
  \label{fig:substrate-ladder}
\end{figure}

The four categories classify the space of the future variable $s$, while Section~\ref{sec:what-coupling} classifies how $s$ and $a$ are tied together. The distinction matters. Fast-WAM~\citep{FastWAM} is feature-substrate and encoder-only because its future branch is masked at inference and the action expert uses a hidden state. $\tau_0$-WM~\citep{tau0wm2026} is pixel-grounded-substrate and joint-denoising because it predicts a Wan VAE video latent together with action velocity. mimic-video~\citep{pai2025mimicvideo} is feature-substrate and post-prediction-head because the action head taps an intermediate ODE state of a frozen video trunk. The substrate names where the WAM forms its future. The coupling axis names how that future reaches action. Figure~\ref{fig:substrate-ladder} arranges the categories and representative members on the same panel.

\subsection{Action Coupling: How Action Enters and Leaves}
\label{sec:what-coupling}

\begin{figure}[t]
  \centering
  \includegraphics[width=\linewidth]{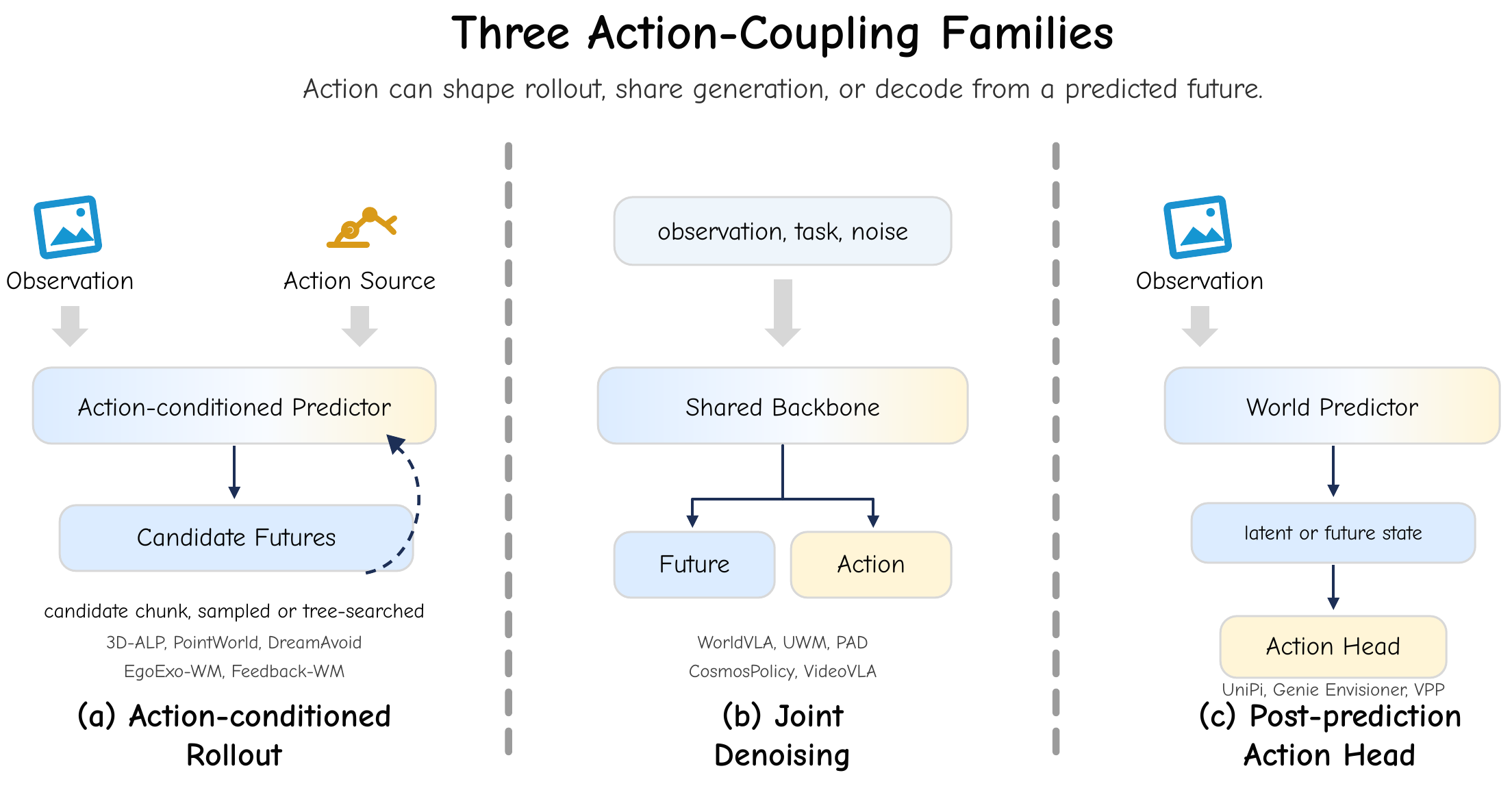}
  \caption{Three common action-coupling families in existing WAMs. The arrangements differ in where the action representation is bound to the substrate prediction, and that binding is what determines latency, controllability, and per-step inference cost.}
  \label{fig:coupling-regimes}
\end{figure}

Action coupling is the structural decision that turns a predictive world model into a WAM. The factorization families in Equations~\ref{eq:fact-rollout}, \ref{eq:fact-joint}, and~\ref{eq:fact-head} pin down the main arrangements. Action-conditioned rollout is a composite family: an outer planner, policy, candidate sampler, or teleoperation stream supplies actions through $q_\psi$, and an action-conditioned world model predicts their consequences. Joint generation and post-prediction heads are action-producing families, either because the future action chunk is produced by the same joint model or because it is decoded from the produced future by a separate head. Within each arrangement, existing methods further differ on how an action is represented and on how many actions are produced per forward pass. Figure~\ref{fig:coupling-regimes} sketches the three top-level arrangements and shows representative candidate forms inside the first. Notely, in this section, when we refer to a method as non-census, we mean that it is cited for background but is not listed as a WAM in Tables~\ref{tab:wams-video} and~\ref{tab:wams-genfree}.

\subsubsection{Action-conditioned rollout}
\label{sec:coup-rollout}

Action-conditioned rollout composes an action source with an action-conditioned world predictor. Repeating Equation~\ref{eq:fact-rollout},
\begin{equation}
\begin{aligned}
p_{\theta,\psi}\!\left(s_{t+1:t+H},a_{t:t+H-1}\mid c\right)
&=
\begin{cases}
q_\psi\!\left(a_{t:t+H-1}\mid c\right)\,
p_\theta\!\left(s_{t+1:t+H}\,\middle|\,c,\,a_{t:t+H-1}\right), & \text{chunk-level},\\
\prod_{k=0}^{H-1}
q_\psi\!\left(a_{t+k}\mid h_k,c\right)
p_\theta\!\left(s_{t+k+1}\,\middle|\,h_k,\,a_{t+k},\,c\right), & \text{step-wise},\\
\end{cases}\\
h_k&=\left(s_{t+1:t+k},a_{t:t+k-1}\right).
\end{aligned}
\label{eq:coup-rollout}
\end{equation}
The first line is chunk-level rollout. The future action chunk is output by $q_\psi$ and enters the substrate predictor as conditioning. The second line is step-wise rollout. It alternates action selection with action-conditioned prediction, so $q_\psi$ can update the next action after each predicted state. Training usually applies the standard substrate-prediction loss of Section~\ref{sec:what-backbone} to $p_\theta$, with the proposed action or action chunk added to the conditioning side. The action source may be a planner, a policy, a human command stream, a fixed candidate sampler, or, in the step-wise background lineage, a diffusion policy queried after each predicted state.

In the current census, the WAMs listed in the tables in this family all sit on the chunk-level submode. The step-wise submode is retained to connect the WAM notation to the model-based RL and interactive-prediction lineage discussed below.

The chunk-level submode is the natural form when the method scores candidate futures before execution. Cosmos-Transfer1~\citep{Cosmos-Transfer1} illustrates a nearby non-census case: segmentation, depth, edge, and other spatial control maps condition world generation for data generation, but the method does not use generated futures as an inference-time robot-action selector. PointWorld~\citep{pointworldscaling3d2026} conditions a point-flow dynamics model on a proposed sequence of robot point flows from MPPI and scores the predicted scene flow, so action search happens outside the predictor rather than inside the trunk. DreamAvoid~\citep{dreamavoidcriticalphase2026} samples flow-policy action chunks only at detected critical phases, renders each chunk with DreamDojo, and executes the candidate with the highest predicted progress value. EgoExo-WM~\citep{egoexowmunlocking2026} conditions a DINOv3-L latent predictor on horizon-length UniEgoMotion candidate body-motion sequences, rolls each candidate forward under MPC, and selects the sequence whose predicted latent ends closest to a visual goal.

Tree-search and feedback-guided variants keep the same family but change the cadence of candidate evaluation. 3D-ALP~\citep{3danchoredlookahead2026} uses MCTS to propose joint actions, converts each proposal into a 3D camera pose through kinematics, and uses a 3D-consistent renderer as a rollout oracle before executing the selected action. Feedback-WM~\citep{feedbackworldmodel2026} uses a lightweight latent transition model to guide a diffusion policy while that policy forms an action chunk, passes the corrected chunk to the retained world-prediction path, and then updates an auxiliary feedback state from the observation after execution. These variants explain why the family should not be split by whether actions arrive as one candidate chunk, a tree node, or a corrected action sequence. In all cases, actions are supplied to a predictor that evaluates their consequences.

The step-wise submode remains useful for background and for future WAMs whose action source reacts inside the imagined trajectory. AdaWorld~\citep{AdaWorld} conditions an SVD-style latent-video diffusion predictor at each rollout step, although it is not part of the census. FlowDreamer~\citep{FlowDreamer} cascades a geometric-primitive flow prediction into a diffusion next-frame generator, a two-stage variant of the same arrangement. The latent-imagination Dreamer line of PlaNet~\citep{PlaNet}, DreamerV3~\citep{DreamerV3}, Dreamer~4~\citep{Dreamer4}, and TransDreamer~\citep{TransDreamer} supplies the model-based RL ancestor of this design, where a planner, actor, critic, or compact dynamics model advances imagined steps depending on the member. The non-WAM antecedent iVideoGPT~\citep{iVideoGPT} interleaves observation, action, and reward tokens in an autoregressive stream, while InteractiveWorldSimulator~\citep{InteractiveWorldSimulator} and RoboScape~\citep{RoboScape} extend interactive prediction with streaming inputs or physics-aware branches.

Action-conditioned rollout is therefore a composite WAM factorization rather than an action-producing world model by itself. Its advantage is counterfactual control: candidate actions shape the future before selection. Its limitation depends on the submode. Chunk-level rollout can evaluate many candidates in parallel, but it cannot react to a control signal that changes inside the predicted window. Step-wise rollout can react after each predicted state, but it loses the parallelism of a single generated window.

\subsubsection{Joint generation}
\label{sec:coup-joint}

The joint factorization produces substrate and action from a single generative process on a shared backbone. Repeating Equation~\ref{eq:fact-joint},
\begin{equation}
p_\theta\!\left(s_{t+1:t+H},\,a_{t:t+H-1}\,\middle|\,c\right).
\label{eq:coup-joint}
\end{equation}
Training is often expressible as, or approximated by, a weighted sum of a substrate-side generative loss and an action-side regression or classification loss,
\begin{equation}
\mathcal{L}_\text{joint}(\theta) \;=\; \mathcal{L}_\text{gen}(s) \;+\; \lambda\,\mathcal{L}_\text{act}(a),
\label{eq:coup-joint-loss}
\end{equation}
where $\mathcal{L}_\text{gen}$ is a diffusion or autoregression loss on $s$ and $\mathcal{L}_\text{act}$ is a flow-matching, regression, or cross-entropy loss on $a$. Section~\ref{sec:what-backbone} reuses Equation~\ref{eq:coup-joint-loss} as the canonical training objective of hybrid backbones.

\textbf{Joint diffusion of video and action.} The joint-diffusion group differs less in its factorization than in the interface through which the two streams meet. UWM~\citep{UWM}, CosmosPolicy~\citep{CosmosPolicy}, VideoVLA~\citep{VideoVLA}, DriveVA~\citep{drivevavideoaction2026}, and DriveWAM~\citep{drivewamvideogenerative2026} place video-side latents and action-side variables in the same denoising or flow-matching state, so one sampled trajectory carries both the imagined future and the action chunk. CoVAR~\citep{CoVAR}, UD-VLA~\citep{udvla1}, GigaWorld-Policy~\citep{gigaworld-policy}, and CKT-WAM~\citep{cktwamparameter2026} keep the same joint target but change the information routing through bridge attention, discrete diffusion, action-centered masking, or teacher-context prefixes. Multisubstrate variants add another future channel without changing the coupling: AIM~\citep{AIM} inserts a value map, DriveDreamer-Policy~\citep{drivedreamerpolicygeometry2026} orders depth, video, and action inside one training pass, MV-VDP~\citep{multiviewvideo2026} and ActionImages~\citep{actionimagesend2026} attach end-effector heatmaps to the visual stream, VTAM~\citep{vtamvideotactile2026} adds a tactile force proxy, HiF-VLA~\citep{hifvlahindsight2025} adds motion-vector foresight tokens, and DUST~\citep{DUST} applies the same idea to a VLM-substrate setting. The useful distinction is therefore whether the added channel changes the substrate, not whether the action is still jointly produced.

\textbf{Joint autoregression of frame and action tokens.} GR-1~\citep{GR-1} and GR-2~\citep{GR2} jointly predict next-frame and next-action tokens autoregressively over a discrete codebook. WorldVLA~\citep{WorldVLA} carries the same recipe to a Chameleon backbone with an action-chunk attention mask between action and frame tokens. PhysGen~\citep{PhysGen} adapts the autoregressive NOVA backbone with continuous physical tokens and a per-token diffusion detokenizer, so that autoregressive image and action streams share a joint training loss in the sense of Equation~\ref{eq:coup-joint-loss}. ICLR-VR~\citep{iclrcontextimitation2026} serializes state, reasoning, and action tokens into a single Llama2-style causal stream under a combined next-token loss, with the gripper-keypoint polyline decoded immediately before each action chunk.

\textbf{Mixture-of-experts hybrids that joint-denoise.} The hybrid branch uses separate experts or heads so that video and action can share a trunk without forcing every layer to treat both variables identically. UVA~\citep{UVA} and PAD~\citep{PAD} are the clean baselines: one shared pass supports both visual prediction and action, and the visual branch can be bypassed when only control is needed. F1~\citep{f1vla}, Motus~\citep{Motus}, MotuBrain~\citep{MotuBrain}, RynnVLA-002~\citep{rynnvla2}, BagelVLA~\citep{bagelvlaenhancinglong2026}, Pelican-Unify~1.0~\citep{pelicanunify12026}, and ALAM~\citep{alamalgebraicallyconsistent2026} instantiate the mixture pattern with different expert allocations for understanding, video, latent transition, and action streams. A second set keeps the same joint target but changes the schedule: DreamZero~\citep{DreamZero} uses one shared timestep across a Wan-style video-action state, X-WAM~\citep{guo2026xwam} lets action finish before multi-view video, NoiseGate~\citep{noisegatelearningper2026} learns per-latent denoising increments, $\tau_0$-WM~\citep{tau0wm2026} samples video and action flow times independently, and WALL-WM~\citep{wallwmcarving2026} couples action queries into matched video blocks at every depth. CLWM~\citep{dexworldmodelcausallatent2026}, DAWN~\citep{dawnworldaction2026}, JOPAT~\citep{pointtrackingimproves2026}, and DDP~\citep{dreamingunseenworld2026} show the same coupling on feature, track, or compact latent substrates, while FFDC-WAM~\citep{whentrustimagination2026} keeps the Motus joint rollout and adds a verifier that decides when the cached future is still trustworthy. The family is diverse, but the shared claim is narrow: the action and the future substrate are trained as mutually constrained variables rather than as a planner output followed by a passive decoder.

The benefit of the family is one coupled sample or rollout that produces both predictions and a strong mutual consistency between them.
The trade-off is potential training instability because the generation and action losses can pull the shared representation in different directions. Practitioners therefore often pretrain the substrate head on video and introduce the action head later, sometimes with a careful $\lambda$ schedule in Equation~\ref{eq:coup-joint-loss}.

\subsubsection{Post-prediction action head}
\label{sec:coup-head}

The post-prediction head factorizes into a substrate generator and a smaller action expert $q_\psi$ that decodes action from the substrate. Repeating Equation~\ref{eq:fact-head},
\begin{equation}
p_\theta\!\left(s_{t+1:t+H},\,a_{t:t+H-1}\,\middle|\,c\right) = p_\theta\!\left(s_{t+1:t+H}\,\middle|\,c\right)\;q_\psi\!\left(a_{t:t+H-1}\,\middle|\,s_{t+1:t+H},\,c\right),
\label{eq:coup-head}
\end{equation}
with $\theta$ frozen during $q_\psi$ training in many implementations. This arrangement is common because it lets a pretrained predictor stay fixed while the action head changes across embodiments.

\textbf{Video-plan then action-recovery WAMs.} UniPi~\citep{unipi} a canonical early example: a text-conditioned diffusion planner first synthesizes a future image trajectory, and a task-specific inverse-dynamics module then recovers the low-level actions between generated frames. VLP~\citep{du2023vlp} adds a VLM policy and value-guided search over abstract text actions, but the executable controls still come from goal-conditioned policies that consume the synthesized video plan. EVA~\citep{evaaligningvideo2026} pairs a Wan2.1-14B planner with a frozen spatial-softmax IDM and uses the IDM-derived smoothness-and-limits reward to align the planner via GRPO without folding the IDM into the joint denoiser. CreFlow~\citep{creflowcorrectivereflow2026} post-trains the Vidar text-image-to-video trunk against an LTL-grounded corrective reflow loss and keeps the Vidar inverse-dynamics expert frozen. These methods are often described as video planning or video world modeling, but under the notation here their action coupling is the post-prediction factorization of Equation~\ref{eq:coup-head}.

\textbf{Action expert on a latent, token, or multimodal substrate.} This group keeps the action expert separate, so the key question is what future object $q_\psi$ consumes. Some members expose a task-structured future before action decoding: $\pi_{0.7}$~\citep{pi0.7} uses BAGEL-style subgoal images as visual targets, CoT-VLA~\citep{CoTVLA} uses visual chain-of-thought tokens, PALM~\citep{palmprogressaware2026} uses affordance maps, and OmniVTA~\citep{zheng2026omnivta} uses visual and tactile latents whose fixed decoders tie them back to observation space for slow-fast contact-rich control. Feature-substrate members differ mainly in where the future feature comes from. FLARE~\citep{FLARE}, LDA-1B~\citep{LDA1B}, FRAPPE~\citep{FRAPPE}, and GigaBrain-0.5M*~\citep{gigabrain05m2026} rely on future embeddings or value-bearing latent tokens, while VPP~\citep{hu2024vpp}, VideoPolicy~\citep{liang2025videopolicy}, Genie Envisioner~\citep{liao2025genieenvisioner}, LaMP~\citep{lamplearningvision2026}, and VAMPO~\citep{vampopolicyoptimization2026} intercept intermediate states of a video-trained trunk. The action expert can therefore be small, but its reliability depends on whether the chosen substrate preserves the action-relevant part of the forecast.

\textbf{Action expert decodes structured geometry from a generated future.} The geometric post-head line uses motion or structure as the intermediate action-facing representation. Im2Flow2Act~\citep{xu2024im2flow2act}, 3DFlowAction~\citep{zhi2025threedflowaction}, 3D-FDP~\citep{3dflowdiffusion2025}, NovaFlow~\citep{li2025novaflow}, Dream2Flow~\citep{dharmarajan2025dream2flow}, and AeroPlace-Flow~\citep{aeroplaceflowlanguage2026} make flow the intermediate future, while TesserAct~\citep{TesserAct1}, 4DGen~\citep{liu2025fourdgen}, MVISTA-4D~\citep{wang2026mvista4d}, TC-IDM~\citep{tcidmgrounding2026}, and Dex4D~\citep{dex4dtaskagnostic2026} recover pose or point structure before control. Earlier decoded-video cascades such as AVDC~\citep{ko2024avdc}, Dreamitate~\citep{liang2024dreamitate}, This\&That~\citep{wang2024thisthat}, ARDuP~\citep{huang2024ardup}, Gen2Act~\citep{bharadhwaj2024gen2act}, GR-MG~\citep{grmg}, and GraspDreamer~\citep{graspyoudream2026} fit the same factorization even when the extracted geometry is implicit in a tracker or learned head. RoboEnvision~\citep{yang2025roboenvision}, Vidar~\citep{vidar}, LVP~\citep{chen2025lvp}, Veo-Act~\citep{zhang2026veoact}, Say-Dream-Act~\citep{gu2026saydreamact}, NovaPlan~\citep{novaplanzeroshot2026}, and MoLA~\citep{imaginedfuturesexecutable2026} keep the same post-head order but add stronger video planners, VLM scoring, or modality-specific inverse dynamics. The shared lesson is that a generated future can be action-facing even when the final action module is trained separately.

\textbf{Generated future is partly or fully optional at inference.} The optional-future pattern trains with a substrate generator, then shortens or bypasses that generator when the inference-time action path can use an internal state. VidMan~\citep{VidMan} is the clearest representative: it mounts a policy head on frozen video-diffusion features rather than waiting for a rendered future. mimic-video~\citep{pai2025mimicvideo}, S-VAM~\citep{yan2026svam}, DiT4DiT~\citep{DiT4DiT}, Fast-WAM~\citep{FastWAM}, and VAG~\citep{lang2026vag} make the same move through ODE checkpoints, distilled foresight states, fixed denoising timesteps, masked future-video tokens, or pooled clean latents. WAV~\citep{WAV}, MWM~\citep{lou2026mwm}, DexWM~\citep{DexWM}, and SWEET~\citep{sweetsparseworld2026} instead keep a compact action-facing substrate, such as valued latent rollouts, semantic-mask dynamics, DINO latents, or sparse keyframes. LingBot-VA~\citep{LingBotVA} and HarmoWAM~\citep{harmowamharmonizinggeneralizable2026} show the mixture-backbone version, where visual dynamics remain available but a lighter action expert or gate handles the frequent control path.
This design keeps $q_\psi$ cheap and embodiment-specific. But the risk is equally clear, since the skipped future can help only if the retained state still preserves the part of the forecast that action needs.

\subsubsection{Action representation and chunk size}
\label{sec:coup-rep}

Across the three coupling families, the action variable $a_t\in\mathcal{A}$ is parameterized differently. Three representations are common. The first is per-step continuous controls, with $\mathcal{A}=\mathbb{R}^{d_a}$ for joint velocities, end-effector deltas, or gripper commands. The second is discrete tokens from a fixed codebook, as in autoregressive joint denoisers such as WorldVLA~\citep{WorldVLA}, where $\mathcal{A}$ is a finite vocabulary obtained from per-dimension binning. The third is learned latent actions that act as currency between the world model and the policy. In that case $\mathcal{A}=\mathbb{R}^{d_z}$ for some learned $d_z\ll d_a$, as in Motus~\citep{Motus}, ALAM~\citep{alamalgebraicallyconsistent2026}, and LDA-1B~\citep{LDA1B}. The chunk size is the second nested choice. A model that produces one action per forward pass pays a low per-action latency but invokes the full backbone at the control frequency. A model that produces a long chunk per pass spreads the backbone cost over the chunk but cannot react during it. Closed-loop control on real hardware pushes methods toward smaller chunks or toward cheaper backbones, and many of the architectural choices in Section~\ref{sec:what-backbone} can be reinterpreted as a response to that pressure.

\subsection{Architectural Backbone: How the Prediction Is Produced}
\label{sec:what-backbone}

Once a substrate space $\mathcal{S}$ is fixed and a coupling factorization is chosen, the next ingredient is the function family that realizes $p_\theta(\cdot\mid c)$. We group current backbones into five families, and for each we list the defining parameterization, the canonical training objective, and the WAMs that use it.

\subsubsection{Iterative-denoising backbones (video diffusion)}
\label{sec:bb-diffusion}

The diffusion family writes $p_\theta$ as the marginal of a learned reverse process over a sequence of progressively denoised latents $s^{(N_\text{diff})},s^{(N_\text{diff}-1)},\dots,s^{(0)}=s_{t+1:t+H}$:
\begin{equation}
p_\theta(s_{t+1:t+H}\mid c) = \int p(s^{(N_\text{diff})})\prod_{n=1}^{N_\text{diff}} p_\theta\!\left(s^{(n-1)}\mid s^{(n)},c\right)\,\mathrm{d}s^{(N_\text{diff}:1)},
\label{eq:bb-diff}
\end{equation}
where $p(s^{(N_\text{diff})})$ is a Gaussian noise prior and each reverse step is parameterized by a noise-prediction network $\epsilon_\theta(s^{(n)},n,c)$. Training minimizes the standard score-matching loss
\begin{equation}
\mathcal{L}_\text{diff}(\theta) = \mathbb{E}_{n,s^{(0)},\epsilon}\left[\big\|\epsilon - \epsilon_\theta\!\left(s^{(n)},\,n,\,c\right)\big\|^2\right],\quad s^{(n)}=\alpha_n s^{(0)}+\sigma_n\epsilon.
\label{eq:bb-diff-loss}
\end{equation}
At inference, sampling draws $s^{(N_\text{diff})}\!\sim\!p(s^{(N_\text{diff})})$ and runs $N_\text{diff}$ reverse steps.

\textbf{Founding video-diffusion backbones.} The dominant topology is a video diffusion transformer that attends jointly over space and time, as in Latte~\citep{Latte}, CogVideoX~\citep{CogVideoX}, Wan~\citep{Wan}, and Cosmos-Predict2~\citep{cosmos-predict2}. AnimateDiff~\citep{AnimateDiff} represents an earlier generation that adapts a frozen image-diffusion backbone with a learned temporal motion module.

Because standard diffusion denoises a whole window at once rather than advancing one element at a time, the coupling-type landscape is narrower than the autoregressive one. The census mainly uses diffusion in three ways:
\begin{equation}
\left(\mathcal{Y},c_\mathrm{diff}\right)=
\begin{cases}
\left(s_{t+1:t+H},\, (c,a_{t:t+H-1})\right), & \text{action-conditioned rollout},\\
\left((s_{t+1:t+H},\,a_{t:t+H-1}),\,c\right), & \text{joint generation},\\
\left(s_{t+1:t+H},\,c\right)\ \text{followed by }q_\psi(a_{t:t+H-1}\mid s_{t+1:t+H},c), & \text{post-prediction head}.
\end{cases}
\label{eq:bb-diff-coupling-map}
\end{equation}
Pure diffusion supports action-conditioned rollout most naturally when the action source proposes a whole candidate chunk before the reverse chain begins, as in DreamAvoid~\citep{dreamavoidcriticalphase2026}. Step-wise rollout is less natural for a windowed reverse chain, so it appears mainly in non-census foundation or simulator works such as AdaWorld~\citep{AdaWorld} and FlowDreamer~\citep{FlowDreamer}, or in hybrid wrappers that alternate an action source and a predictor.

\noindent\textbf{Post-prediction head (Eq.~\ref{eq:fact-head}).} Many diffusion WAMs first denoise the substrate $s^{(0)}$ and then decode action from it with a separate module $q_\psi$. The important variation is which substrate $q_\psi$ consumes, not the mere fact that the action head is separate.

Decoded-observation members use the generated visual future as the common interface. UniPi~\citep{unipi} is the representative case, since a text-conditioned diffusion rollout is followed by inverse dynamics. VLP~\citep{du2023vlp} changes the conditioning context before sampling, while AVDC~\citep{ko2024avdc}, Dreamitate~\citep{liang2024dreamitate}, This\&That~\citep{wang2024thisthat}, Gen2Act~\citep{bharadhwaj2024gen2act}, and GR-MG~\citep{grmg} recover actions, tool motion, or goals from the produced frames. Later visual planners mostly strengthen the generator or the search wrapper: RoboEnvision~\citep{yang2025roboenvision}, Vidar~\citep{vidar}, MVISTA-4D~\citep{wang2026mvista4d}, LVP~\citep{chen2025lvp}, Veo-Act~\citep{zhang2026veoact}, and Say-Dream-Act~\citep{gu2026saydreamact} keep the same factorization while changing horizon, view structure, executor, or sampling schedule.

Pixel-latent and geometric variants keep the post-head order but make the substrate cheaper or closer to control. ARDuP~\citep{huang2024ardup}, VILP~\citep{xu2025vilp}, DiT4DiT~\citep{DiT4DiT}, and OmniVTA~\citep{zheng2026omnivta} act from latent video, intermediate denoising states, or visual-tactile decoder-bound latents. Im2Flow2Act~\citep{xu2024im2flow2act} is the representative geometric instance because it denoises object flow as the action-facing future. 3DFlowAction~\citep{zhi2025threedflowaction} and TesserAct~\citep{TesserAct1} extend the same idea to 3D flow or reconstructed point clouds. The trade-off is direct: these substrates reduce appearance burden, but their usefulness depends on whether the chosen geometry captures the task-relevant contact and motion.

Feature-level post heads push the shortcut further by consuming latent states rather than decoded futures. VPP~\citep{hu2024vpp}, Video Policy~\citep{liang2025videopolicy}, Genie Envisioner~\citep{liao2025genieenvisioner}, Act2Goal~\citep{Act2Goal}, WAV~\citep{WAV}, VAG~\citep{lang2026vag}, EnerVerse~\citep{EnerVerse}, VidMan~\citep{VidMan}, mimic-video~\citep{pai2025mimicvideo}, S-VAM~\citep{yan2026svam}, and Fast-WAM~\citep{FastWAM} use variations of this shortcut, such as tapping intermediate denoising states, distilling foresight features, fixing a small number of denoising steps, or bypassing explicit video generation at inference. Related compact-substrate variants, such as MWM~\citep{lou2026mwm}, pursue a similar latency goal through task maps rather than through a pure feature tap.

\noindent\textbf{Joint generation (Eq.~\ref{eq:fact-joint}).} The second group of diffusion WAMs models the future substrate $s$ and the future action $a$ within a coupled denoising or flow-matching process, so Equation~\ref{eq:bb-diff} operates on the joint variable $(s,a)$. PAD~\citep{PAD} is a compact image-action DiT instance, while VideoVLA~\citep{VideoVLA}, CosmosPolicy~\citep{CosmosPolicy}, CoVAR~\citep{CoVAR}, UD-VLA~\citep{udvla1}, GigaWorld-Policy~\citep{gigaworld-policy}, and CKT-WAM~\citep{cktwamparameter2026} vary the shared denoising state through a pretrained video trunk, bridge attention, discrete tokens, action-centered masking, or transfer prefixes. GigaWorld-Policy is the action-centered case in this group: its causal mask prevents future-video tokens from influencing action tokens, so deployment can sample the action chunk without instantiating future-video tokens. AIM~\citep{AIM}, X-WAM~\citep{guo2026xwam}, and AdaWorldPolicy~\citep{AdaWorldPolicy} show the same diffusion coupling with value maps, multi-view 4D video, or force-prediction modules attached to the joint state. The benefit is a coupled generative process that encourages consistency between action and future prediction. The trade-off is potential training instability because the generation and action losses can impose competing demands on the shared representation. Practitioners therefore often pretrain the substrate head on video and introduce the action head later, sometimes with a careful $\lambda$ schedule.

\noindent\textbf{Flow matching as a variant of Equation~\ref{eq:bb-diff}.} A large subset of the diffusion family instantiates $\epsilon_\theta$ as a velocity-field predictor under flow matching rather than as a noise predictor under score matching, but both still satisfy the iterative-reverse-chain parameterization of Equation~\ref{eq:bb-diff}. Conditional flow matching learns $v_\theta(x_t,t,c)$ to regress the linear interpolation velocity $x_1-x_0$ under $\mathcal{L}_\text{FM}=\mathbb{E}\,\|(x_1-x_0)-v_\theta(x_t,t,c)\|_2^2$, and at inference the same Euler reverse integration takes the place of the noise-prediction reverse chain. The flow-matching variant is therefore not a new backbone family but a different choice of the score parameterization inside Equation~\ref{eq:bb-diff}. Among the post-prediction-head WAMs above, 3D-FDP~\citep{3dflowdiffusion2025}, EVA~\citep{evaaligningvideo2026}, VAMPO~\citep{vampopolicyoptimization2026}, and MoLA~\citep{imaginedfuturesexecutable2026} run flow matching as the score parameterization. Among the joint-denoising WAMs, DriveVA~\citep{drivevavideoaction2026}, ActionImages~\citep{actionimagesend2026}, MV-VDP~\citep{multiviewvideo2026}, and CreFlow~\citep{creflowcorrectivereflow2026} take the same route on a Wan-class trunk. The choice between score matching and flow matching remains separable from the coupling family within Equation~\ref{eq:bb-diff}.

\subsubsection{Autoregressive backbones (next-frame, next-token)}
\label{sec:bb-ar}

The autoregressive family is a backbone parameterization, not an action-coupling regime by itself. It realizes whichever generative factor from Section~\ref{sec:what-coupling} is assigned to it by serializing that factor's output into a causal stream. Let
\begin{equation}
y_{1:M}=\operatorname{Serialize}(\mathcal{Y}),
\label{eq:bb-ar-stream}
\end{equation}
where $\mathcal{Y}$ is the set of variables generated by the autoregressive backbone, and let $c_\mathrm{ar}$ denote the local context supplied to that backbone. The generic autoregressive parameterization is
\begin{equation}
p_\theta(y_{1:M}\mid c_\mathrm{ar})
=
\prod_{j=1}^{M}
p_\theta\!\left(y_j\mid y_{<j},c_\mathrm{ar}\right),
\label{eq:bb-ar}
\end{equation}
where $y_j$ may be a frame, a latent-grid block, a discrete visual token, a continuous latent token, an action token, or a compact latent state. Training minimizes the corresponding next-element loss
\begin{equation}
\mathcal{L}_\text{ar}(\theta)
=
-\sum_{j=1}^{M}
\log p_\theta\!\left(y_j\mid y_{<j},c_\mathrm{ar}\right),
\label{eq:bb-ar-loss}
\end{equation}
with cross-entropy for discrete tokens and a continuous likelihood, regression, diffusion, or flow-matching loss for continuous elements. At inference, a key-value cache reuses prefix computation, so each newly appended element requires one forward pass over the new tokens rather than recomputation over the whole history.

The pair $(\mathcal{Y},c_\mathrm{ar})$ is inherited from the coupling choice rather than from autoregression itself:
\begin{equation}
\left(\mathcal{Y},c_\mathrm{ar}\right)=
\begin{cases}
\left(s_{t+1:t+H},\, (c,a_{t:t+H-1})\right)\ \text{or}\ \left(s_{t+k+1},\, (h_k,a_{t+k},c)\right), & \text{action-conditioned rollout},\\
\left((s_{t+1:t+H},a_{t:t+H-1}),\,c\right), & \text{joint autoregression},\\
\left(s_{t+1:t+H},\,c\right)\ \text{followed by }q_\psi(a_{t:t+H-1}\mid s_{t+1:t+H},c), & \text{post-prediction head}.
\end{cases}
\label{eq:bb-ar-coupling-map}
\end{equation}
Equation~\ref{eq:bb-ar-coupling-map} is only a backbone-level map. The full joint factorizations remain those in Section~\ref{sec:what-coupling}. Its purpose is to prevent the common confusion that autoregression implies joint action prediction. In many WAMs the autoregressive stream predicts only the future substrate. The future action may be supplied to the rollout, generated in the same stream, or decoded later by $q_\psi$.

\noindent\textbf{Autoregressive video priors (not WAMs).} The following examples provide the video prior but not the WAM coupling. The list is selective rather than exhaustive, because these works cover the substrate forms that later WAMs reuse. VideoGPT~\citep{VideoGPT} gives the VQ-VAE plus GPT recipe over discrete video latents. VideoPoet~\citep{VideoPoet} is a decoder-only multimodal video language model over tokenized visual and audio streams. NOVA~\citep{NOVA} removes the discrete codebook and predicts continuous video latents. These works define reusable autoregressive video priors, but they are not WAMs by themselves. Earlier GAN-based video generators such as MoCoGAN~\citep{MoCoGAN} belong to the historical substrate discussion of Section~\ref{sec:what-substrate}. They are not instances of the next-element parameterization in Equation~\ref{eq:bb-ar}.

\noindent\textbf{Action-conditioned rollout (Eq.~\ref{eq:fact-rollout}).} The following methods use autoregression for action-conditioned sequential dynamics. iVideoGPT~\citep{iVideoGPT} brings the VideoGPT-style token stream into interactive world modeling by interleaving observations, actions, and rewards. AdaWorld~\citep{AdaWorld} is the boundary case in this group. The paper calls it an autoregressive world model because inference repeats action-conditioned next-frame prediction, but each local transition is an SVD-based denoising problem. It is therefore best understood as an autoregressive rollout with a diffusion transition, not as a homogeneous next-token decoder. PlaNet~\citep{PlaNet}, DreamerV3~\citep{DreamerV3}, Dreamer~4~\citep{Dreamer4}, and TransDreamer~\citep{TransDreamer} are latent-imagination methods based on compact dynamics models. These methods explain the sequential-dynamics side of the family, but they usually pair the dynamics model with a separate actor, planner, or critic and therefore serve here as background rather than census WAMs.

\noindent\textbf{Joint autoregression (Eq.~\ref{eq:fact-joint}).} The following methods use autoregression as the trunk for streams that include both future substrate elements and action-facing elements. WorldVLA~\citep{WorldVLA} is the clearest discrete instance: it serializes language, image, and action tokens in one autoregressive token framework. PhysGen~\citep{PhysGen} is the continuous-latent counterpart, adapting the NOVA~\citep{NOVA} autoregressive video prior to continuous physical tokens. GR-1~\citep{GR-1} establishes the GPT-style video-generative trunk for robot manipulation, and GR-2~\citep{GR2} scales that trunk with larger video-language pretraining and robot-trajectory fine-tuning. From the backbone perspective, the shared property is causal next-element prediction. The exact action head, detokenizer, or joint-output factorization is the coupling detail handled in Section~\ref{sec:what-coupling}.

\noindent\textbf{Post-prediction head (Eq.~\ref{eq:fact-head}).} The post-head line of Equation~\ref{eq:bb-ar-coupling-map} is still possible for autoregressive backbones, even when it is less common than joint token prediction. LingBot-VA~\citep{LingBotVA} is the closest current member because it models visual dynamics before inverse dynamics, but it uses an autoregressive diffusion mixture-of-transformers rather than the next-element backbone of Equation~\ref{eq:bb-ar}. The distinction matters because the backbone can be causal while the action interface remains a separate decoder over the predicted substrate.

The common pressure across the family is serial cost. Autoregression avoids the many denoising steps of diffusion and benefits from caching, but it still advances the generated stream one element at a time. For WAMs, the relevant latency is therefore governed by what the stream contains: full frames, latent blocks, visual tokens, action tokens, or compact latent states. When an autoregressive WAM misses a real-time control budget, the first places to inspect are the serialized substrate size, the action-chunk length, and whether actions are produced inside the same causal stream or decoded by a cheaper head.

\subsubsection{Joint-embedding-predictive backbones}
\label{sec:bb-jepa}

The joint-embedding-predictive family realizes $p_\theta(s_{t+1:t+H}\mid c)$ as a Dirac at a deterministic predictor rather than as a stochastic density. It learns a context encoder $\mathbf{E}^{\text{ctx}}$ and an exponential-moving-average target encoder $\mathbf{E}^{\text{tgt}}$ together with a predictor $f_\theta^{\text{pred}}$ that maps context features to target features. Given a context view $x_\text{ctx}$ of the past and a target view $x_\text{tgt}$ of the future, the objective is
\begin{equation}
\mathcal{L}_\text{jepa}(\theta) = \mathbb{E}_{x_\text{ctx},x_\text{tgt}}\left[\big\|f_\theta^{\text{pred}}\!\left(\mathbf{E}^{\text{ctx}}(x_\text{ctx})\right) - \mathrm{sg}\!\left[\mathbf{E}^{\text{tgt}}(x_\text{tgt})\right]\big\|^2\right],
\label{eq:bb-jepa}
\end{equation}
with $\mathrm{sg}[\cdot]$ a stop-gradient. Under the unified contract of Equation~\ref{eq:wam-joint}, the substrate $s_{t+1:t+H}$ is $\mathbf{E}^{\text{tgt}}(x_\text{tgt})$, and the predictive distribution $p_\theta(s_{t+1:t+H}\mid c)$ collapses to a Dirac at $f_\theta^{\text{pred}}(\mathbf{E}^{\text{ctx}}(x_\text{ctx}))$. The JEPA backbone never inverts $\mathbf{E}^{\text{tgt}}$, which makes the inference loop cheap and is the source of the family's appeal. The backbone predicts the substrate only. Actions enter through one of two coupling patterns:
\begin{equation}
\left(\mathcal{Y},c_\mathrm{jepa}\right)=
\begin{cases}
\left(s_{t+k+1},\,(h_k,a_{t+k},c)\right)\ \text{at rollout step }k, & \text{action-conditioned rollout},\\
\left(s_{t+1:t+H},\,c\right)\ \text{followed by }q_\psi(a_{t:t+H-1}\mid s_{t+1:t+H},c), & \text{post-prediction head}.
\end{cases}
\label{eq:bb-jepa-coupling-map}
\end{equation}
Action-conditioned rollout treats the JEPA predictor as a world model and recovers actions through planning over imagined consequences. A post-prediction head predicts the full substrate window first and then decodes action from it with a separate expert $q_\psi$.

I-JEPA~\citep{I-JEPA} introduces the architecture on images, predicting the latent representations of large masked target blocks from a spatially distributed context block. A-JEPA~\citep{A-JEPA} carries the same recipe to audio spectrograms under a curriculum time-frequency masking strategy, and MC-JEPA~\citep{MC-JEPA} couples the prediction objective with optical-flow learning so that one shared encoder serves both motion and content. V-JEPA~\citep{V-JEPA} and V-JEPA~2~\citep{V-JEPA-2} extend the idea to video through pure feature prediction, and the V-JEPA~2 action-conditioned variant shows how such features can support MPC through action-conditioned latent rollout. The tabulated WAMs use this lineage in two ways. FLARE~\citep{FLARE} pairs a deterministic JEPA predictor with a downstream action expert in the post-prediction-head factorization by aligning additional DiT tokens to future-observation embeddings from a frozen teacher. DAWN~\citep{dawnworldaction2026} keeps V-JEPA~2-style feature tokens inside a hybrid WAM that alternates compact latent-world prediction with action denoising. The common pressure is that the embedding space has no intrinsic visual quality measure, so downstream-task accuracy becomes the only standard test of the world model.

\subsubsection{Hybrid backbones (generative head plus action head)}
\label{sec:bb-hybrid}

The hybrid family runs a generative head and an action head on the same backbone. Schematically,
\begin{equation}
p_\theta\!\left(s_{t+1:t+H},\,a_{t:t+H-1}\mid c\right) \;\propto\; g_\theta(c)\;\longrightarrow\;\big\{h_\theta^{s},\,h_\theta^{a}\big\},
\label{eq:bb-hybrid}
\end{equation}
where $g_\theta$ is a shared encoder-decoder trunk and $h_\theta^{s},h_\theta^{a}$ are two output heads. Training minimizes the joint coupling loss of Equation~\ref{eq:coup-joint-loss} from Section~\ref{sec:what-coupling}, namely $\mathcal{L}_\text{gen}(s)+\lambda\mathcal{L}_\text{act}(a)$, with the substrate head and action head sharing the trunk $g_\theta$.

Because the shared trunk $g_\theta$ feeds both heads in one place, the hybrid family is the backbone in which every coupling family of Section~\ref{sec:what-coupling} can appear, and which one a given model realizes is fixed by how the substrate head $h_\theta^{s}$ and the action head $h_\theta^{a}$ are ordered around that trunk:
\begin{equation}
\left(h_\theta^{s},\,h_\theta^{a}\right)=
\begin{cases}
(s,a)\ \text{jointly from}\ g_\theta(c), & \text{joint generation},\\
s\ \text{from}\ h_\theta^{s},\ \text{then}\ p_\theta^{a}\!\left(a\mid s,c\right), & \text{post-prediction head},\\
q_\psi(a\mid c)\ \text{or external search},\ \text{then}\ p_\theta^{s}(s\mid c,a), & \text{action-conditioned rollout}.
\end{cases}
\label{eq:bb-hybrid-coupling-map}
\end{equation}
Here $p_\theta^{s}$ and $p_\theta^{a}$ denote the substrate-prediction and action-prediction distributions implemented by $h_\theta^{s}$ and $h_\theta^{a}$ when a shared trunk feeds multiple heads. The action-producing couplings place the action inside or after the shared pass, whereas action-conditioned rollout wraps the same trunk with a planner, sampler, tree search, or feedback-guided action proposal.

\noindent\textbf{Joint generation (Eq.~\ref{eq:fact-joint}).} Joint generation is the native arrangement of the family, because a single pass of $g_\theta(c)$ produces the future substrate $s$ and the future action $a$ under the loss of Equation~\ref{eq:coup-joint-loss}. The plain shared-trunk form is easiest to see in UVA~\citep{UVA}: the video head shapes the representation during training, and the action head can be used alone at inference. UWM~\citep{UWM}, DUST~\citep{DUST}, VTAM~\citep{vtamvideotactile2026}, DDP~\citep{dreamingunseenworld2026}, DriveDreamer-Policy~\citep{drivedreamerpolicygeometry2026}, and DriveWAM~\citep{drivewamvideogenerative2026} keep that contract while changing the substrate from pixel-decodable latents to VLM tokens, tactile force proxies, compact latent states, or driving-specific video and depth.

Mixture-of-Transformer variants keep joint generation but separate expert routes so that video and action do not have to share every layer. F1~\citep{f1vla} is the representative mixture case, with generation and action experts coupled by next-scale prediction. Motus~\citep{Motus}, MotuBrain~\citep{MotuBrain}, X-WAM~\citep{guo2026xwam}, and RynnVLA-002~\citep{rynnvla2} vary the same idea through optical-flow latent actions, Vidu latents, asynchronous denoising, or an autoregressive-image plus continuous-action split. BagelVLA~\citep{bagelvlaenhancinglong2026}, CLWM~\citep{dexworldmodelcausallatent2026}, Pelican-Unify~1.0~\citep{pelicanunify12026}, DAWN~\citep{dawnworldaction2026}, NoiseGate~\citep{noisegatelearningper2026}, WALL-WM~\citep{wallwmcarving2026}, and $\tau_0$-WM~\citep{tau0wm2026} then vary the schedule and exchange pattern through same-timestep attention, feature flow matching, iterative refinement, per-latent timestep gates, depth-wise cross-attention, or propose-rank-rectify control.

FFDC-WAM~\citep{whentrustimagination2026} sits at the boundary between cached chunked rollout and closed-loop re-planning: the Motus joint rollout remains the action-facing substrate, while a lightweight verifier decides when cached imagination should be trusted. Across the group, the shared benefit is consistency between the predicted world and the chosen action. The cost is that the action and future heads must train together without one objective overwhelming the other.

\noindent\textbf{Post-prediction head (Eq.~\ref{eq:fact-head}).} A different arrangement keeps the shared trunk but orders the heads, so $h_\theta^{s}$ completes the substrate window before $h_\theta^{a}$ decodes the action from it. FRAPPE~\citep{FRAPPE} is the teacher-feature representative, routing multiple predicted future embeddings into one action transformer. Genie Envisioner~\citep{liao2025genieenvisioner}, GigaBrain-0.5M*~\citep{gigabrain05m2026}, and LaMP~\citep{lamplearningvision2026} instead expose video-trained latent, value, or motion-expert states to the action expert. LingBot-VA~\citep{LingBotVA} and HarmoWAM~\citep{harmowamharmonizinggeneralizable2026} keep a visual-dynamics branch active but hand frequent control to inverse-dynamics or gated action experts. Dex4D~\citep{dex4dtaskagnostic2026} and DexWM~\citep{DexWM} show the same order on geometric or DINO-latent futures. The action expert here conditions only on the predicted substrate, so it stays cheap and can be retrained per embodiment without touching the trunk, provided that substrate is action-informative.

\noindent\textbf{Action-conditioned rollout (Eq.~\ref{eq:fact-rollout}).} Action-conditioned rollout enters the family when an action source is specified before or during consequence prediction. 3D-ALP~\citep{3danchoredlookahead2026} is the pixel-grounded tree-search version, where MCTS proposes joint actions, kinematics converts them into 3D camera poses, and a 3D-consistent world model renders candidate futures for scoring. PointWorld~\citep{pointworldscaling3d2026} is the geometric chunk-level version, composing an MPPI sampler over end-effector trajectories with an action-conditioned predictor over a robot point-flow substrate. EgoExo-WM~\citep{egoexowmunlocking2026} uses horizon-length body-motion candidates from UniEgoMotion to condition a DINOv3-L latent predictor under MPC. Feedback-WM~\citep{feedbackworldmodel2026} uses the retained predictor to guide or correct proposed action chunks after the action source has formed them. The retained world-prediction module therefore scores or corrects proposed actions instead of producing them as an unconstrained action head.

Across the hybrid backbone family, the main cost is maintaining both a world-prediction path and an action path in the same runtime design. The benefit depends on the coupling. Joint generation encourages consistency between predicted futures and actions within a coupled generative process. Post-prediction heads keep the action expert relatively cheap once the substrate has been produced. Action-conditioned rollout supports counterfactual scoring, but repeated candidate evaluation can dominate inference. For joint members, the main optimization challenge is that substrate and action objectives can impose competing demands on the shared trunk. Practitioners therefore often pretrain the substrate path on video before introducing the action path, sometimes with a scheduled weight on the action loss in Equation~\ref{eq:coup-joint-loss}.

\subsubsection{LLM- and VLM-backbone WAMs}
\label{sec:bb-llm}

The fifth family builds a WAM directly on a large language or vision-language backbone without using a video-generation trunk as the main policy backbone. The natural substrate is the VLM-token variant of the feature category in Equation~\ref{eq:sub-latent}, since the future is described as a token block in the VLM's vocabulary, and several members of the family reach across into the geometric or affordance categories of Section~\ref{sec:what-substrate}. The backbone is the VLM itself. The contract of Equation~\ref{eq:wam-joint} is realized by composing a VLM forward pass with an optional action expert under the post-prediction-head factorization:
\begin{equation}
s_{t+1:t+H} \sim \mathrm{VLM}_\theta\!\left(\mathrm{prompt}(c)\right),\qquad a_{t:t+H-1} \sim q_\psi\!\left(\,\cdot\,\mid s_{t+1:t+H},\,c\right).
\label{eq:bb-vlm}
\end{equation}
The factorization in Equation~\ref{eq:bb-vlm} realizes a WAM when a future substrate is part of the control pathway, either produced by the LLM/VLM itself or supplied by an attached world-model module before action decoding. A VLM policy that maps observations directly to action tokens, without predicting or consuming such a future substrate, falls outside this family under our definition. This subsection therefore covers LLM- and VLM-backbone designs in which future tokens, subgoal images, motion representations, affordance maps, or related latent transitions are present in the inference-time control context.

Among the $\pi$-series models, $\pi_{0.7}$~\citep{pi0.7} satisfies this gate by conditioning the policy on generated subgoal images. At runtime, a lightweight BAGEL-initialized world model produces multi-view near-future visual subgoals, which are encoded together with observations by the Gemma3-based VLA before its 860M-parameter flow-matching action expert predicts the action chunk.
CoT-VLA~\citep{CoTVLA} writes an explicit visual chain-of-thought into the substrate before action is decoded. LDA-1B~\citep{LDA1B} learns dynamics, policy, and forecasting jointly in a structured DINO-based latent on a Qwen3-VL backbone. HiF-VLA~\citep{hifvlahindsight2025} adopts a pretrained VLA trunk with DINOv2 and SigLIP encoders that produces both motion-vector substrate tokens and action latents in one inference pass. PALM~\citep{palmprogressaware2026} carries the family down to a GPT-2-class transformer over CLIP text, MAE vision, and MLP-encoded state, with a diffusion transformer attached as the action expert. ICLR-VR~\citep{iclrcontextimitation2026} adopts a Llama2-style causal transformer with a pretrained ViT image encoder and decodes a future-third-view gripper-keypoint polyline as the structured substrate before the action chunk fires. ALAM~\citep{alamalgebraicallyconsistent2026} pairs a PaliGemma-2B trunk with a Gemma-300M flow-matching expert, with an algebraic consistency regulariser living entirely in the frozen pretraining encoder. These examples keep the VLM family inside the WAM gate because the future substrate remains in the inference forward path.

The whole family inherits the cost of the VLM stack rather than that of video diffusion, which is why many members use action chunking and a frozen backbone at inference.

\subsubsection{The five families together}
\label{sec:bb-summary}

The five families are not mutually exclusive and the dividing lines move every few months. Equations~\ref{eq:bb-diff}, \ref{eq:bb-ar}, \ref{eq:bb-jepa}, \ref{eq:bb-hybrid}, and \ref{eq:bb-vlm} are the canonical parameterizations and they appear across substrates. A practitioner who fixes the substrate in Section~\ref{sec:what-substrate} has narrowed the backbone choices by construction, since not every substrate is compatible with every family. The remaining freedom is mostly about where the prediction lands relative to the control loop, which is the subject of Section~\ref{sec:what-deploy}.

\subsection{Deployment Regime: Interactive, Rollout, Open Loop, Closed Loop}
\label{sec:what-deploy}

The final ingredient is the regime the WAM is meant to run in. It is the choice of horizon $H$ in Equation~\ref{eq:wam-joint} and of the cadence with which the WAM is invoked relative to the control loop. Let $T$ be the task length in control steps, let $N_\text{fwd}(H)$ be the cost of one forward pass that produces a substrate trajectory of length $H$, and let $f_\text{ctrl}$ be the control frequency.
The four regimes below differ primarily in how they allocate $N_\text{fwd}$ over the task of length $T$ and how frequently new observations can revise the current plan.

\subsubsection{Open-loop rollout}
\label{sec:dep-open}

Open-loop rollout sets $H\!\approx\!T$ and invokes the WAM once. A long window of future substrate is generated once before execution and a separate actor consumes it. Total compute spent on the WAM is
\begin{equation}
\mathcal{C}_\text{open}(T) = N_\text{fwd}(T),
\label{eq:dep-open}
\end{equation}
which is paid once per scenario and does not scale with $f_\text{ctrl}$. UniPi~\citep{unipi}, AVDC~\citep{ko2024avdc}, Dreamitate~\citep{liang2024dreamitate}, This\&That~\citep{wang2024thisthat}, ARDuP~\citep{huang2024ardup}, Gen2Act~\citep{bharadhwaj2024gen2act}, RoboEnvision~\citep{yang2025roboenvision}, LVP~\citep{chen2025lvp}, and MVISTA-4D~\citep{wang2026mvista4d} run the WAM open-loop in their original setting, with the actor consuming the offline rollout through tracking or inverse dynamics. PointWorld~\citep{pointworldscaling3d2026} marks the open-loop limit of planning: MPPI evaluates chunked point-flow predictions over the horizon before execution, but the reported experiments do not replan during execution. Data-generation uses of Cosmos-Transfer1~\citep{Cosmos-Transfer1} and Sora-class video models~\citep{Sora,Sora_2} illustrate a related but non-census use of generated futures, where the future model supplies synthetic experience rather than acting as an inference-time WAM. The benefit of the regime is that the cost is paid once per scenario. Its main limitation is that the rollout cannot react to real-world deviations, so the actor that consumes it must absorb every distribution shift on its own.

\subsubsection{Chunked closed-loop control}
\label{sec:dep-chunked}

Chunked closed-loop sets $H\!=\!K$ for some replan period $K$ and invokes the WAM every $K$ control steps. The WAM produces an action chunk of length $K$, the actor executes the chunk in $K/f_\text{ctrl}$ seconds, and the next invocation begins. Total compute over a task of length $T$ is
\begin{equation}
\mathcal{C}_\text{chunk}(T,K) = \left\lceil T/K\right\rceil\,N_\text{fwd}(K),
\label{eq:dep-chunk}
\end{equation}
which interpolates between open-loop ($K\!\to\!T$) and single-step ($K\!\to\!1$). It is feasible as long as $N_\text{fwd}(K)/K < 1/f_\text{ctrl}$, so that the next chunk is prepared before the previous one is consumed. This is a common choice for real-robot deployment because it amortizes a large backbone over multiple control ticks.

\textbf{VLM-stack chunked controllers.} In VLM-stack WAMs, chunked control appears when a future-substrate token block or structured future is refreshed before an action chunk is decoded. $\pi_{0.7}$~\citep{pi0.7} produces a chunked action from its Gemma3 flow-matching expert after a separate BAGEL-class world model refreshes its multi-view subgoal images every four seconds, making it the $\pi$-series member that satisfies the inference-time gate. CoT-VLA~\citep{CoTVLA} writes a visual chain-of-thought before each chunk is decoded. HiF-VLA~\citep{hifvlahindsight2025} models motion-vector substrate tokens and action latents in parallel on a pretrained VLA trunk. PALM~\citep{palmprogressaware2026} and ICLR-VR~\citep{iclrcontextimitation2026} place a diffusion or autoregressive action expert on top of an affordance map or future-third-view gripper-keypoint polyline substrate. ALAM~\citep{alamalgebraicallyconsistent2026} uses algebraically consistent latent transitions on a PaliGemma-family trunk. FRAPPE~\citep{FRAPPE}, LDA-1B~\citep{LDA1B}, DUST~\citep{DUST}, FLARE~\citep{FLARE}, and GR-2~\citep{GR2} also use chunked control, with the substrate supplied by a future embedding, a dual-stream diffusion head, or autoregression over frame and action tokens.

\textbf{Pixel-latent and feature chunked controllers.} WAMs that avoid decoded observation output use the same amortization principle, but differ in how much of the future substrate remains active at inference. UVA~\citep{UVA} and UWM~\citep{UWM} are the representative head-skipping cases, because the video side helps training while the inference call can produce only the action chunk. F1~\citep{f1vla}, Motus~\citep{Motus}, MotuBrain~\citep{MotuBrain}, BagelVLA~\citep{bagelvlaenhancinglong2026}, NoiseGate~\citep{noisegatelearningper2026}, Pelican-Unify 1.0~\citep{pelicanunify12026}, $\tau_0$-WM~\citep{tau0wm2026}, and WALL-WM~\citep{wallwmcarving2026} keep the same chunked regime inside MoT trunks. Genie Envisioner~\citep{liao2025genieenvisioner}, mimic-video~\citep{pai2025mimicvideo}, RynnVLA-002~\citep{rynnvla2}, UD-VLA~\citep{udvla1}, and CKT-WAM~\citep{cktwamparameter2026} show adjacent ways to reduce per-call work, through a slow latent refresh, intermediate features, continuous or discrete joint autoregression, or a frozen student with updated context tokens. VideoVLA~\citep{VideoVLA}, CoVAR~\citep{CoVAR}, CosmosPolicy~\citep{CosmosPolicy}, AdaWorldPolicy~\citep{AdaWorldPolicy}, DriveVA~\citep{drivevavideoaction2026}, DriveWAM~\citep{drivewamvideogenerative2026}, and DriveDreamer-Policy~\citep{drivedreamerpolicygeometry2026} keep chunked joint generation active in manipulation or driving settings.

Other chunked designs place the chunk behind a planning or abstraction stage rather than a bare action head. VLP~\citep{du2023vlp} is an early decoded-video case, using value-guided search over abstract actions and replanning before goal-conditioned policies execute low-level controls. ActionImages~\citep{actionimagesend2026}, MV-VDP~\citep{multiviewvideo2026}, CreFlow~\citep{creflowcorrectivereflow2026}, and EVA~\citep{evaaligningvideo2026} decode chunks from pixel-substrate plans. MoLA~\citep{imaginedfuturesexecutable2026}, VAMPO~\citep{vampopolicyoptimization2026}, LaMP~\citep{lamplearningvision2026}, CLWM~\citep{dexworldmodelcausallatent2026}, DDP~\citep{dreamingunseenworld2026}, and DAWN~\citep{dawnworldaction2026} do the same from latent or JEPA-style substrate plans. Among action-conditioned rollouts, 3D-ALP~\citep{3danchoredlookahead2026}, DreamAvoid~\citep{dreamavoidcriticalphase2026}, and Feedback-WM~\citep{feedbackworldmodel2026} keep the execution window chunked, but differ in whether candidates come from tree search, critical-phase sampling, or feedback-guided denoising. EgoExo-WM~\citep{egoexowmunlocking2026} is the horizon-ranking boundary case, since it scores chunk-length body-motion candidates but reports open-loop rollout evaluation. JOPAT~\citep{pointtrackingimproves2026} and 3D-FDP~\citep{3dflowdiffusion2025} move the chunk to joint track or 3D-flow substrates, while Vidar~\citep{vidar}, GR-MG~\citep{grmg}, VILP~\citep{xu2025vilp}, VideoPolicy~\citep{liang2025videopolicy}, Act2Goal~\citep{Act2Goal}, Im2Flow2Act~\citep{xu2024im2flow2act}, TesserAct~\citep{TesserAct1}, and 3DFlowAction~\citep{zhi2025threedflowaction} show that the same chunked-cost equation can sit on pixel-grounded, feature, or geometric futures.

\textbf{Compact or optional-substrate chunked controllers.} Several chunked WAMs reduce the per-call burden by making the future sparse, latent, or optional. Veo-Act~\citep{zhang2026veoact} and Say-Dream-Act~\citep{gu2026saydreamact} still decode from generated visual plans, while VAG~\citep{lang2026vag}, S-VAM~\citep{yan2026svam}, DiT4DiT~\citep{DiT4DiT}, Fast-WAM~\citep{FastWAM}, WAV~\citep{WAV}, and DexWM~\citep{DexWM} move the action path to feature or teacher-target futures. MWM~\citep{lou2026mwm} and AIM~\citep{AIM} replace appearance with task maps, and GigaWorld-Policy~\citep{gigaworld-policy} and X-WAM~\citep{guo2026xwam} keep a joint video-action model but let the inference action path avoid the full visual schedule when possible.

\textbf{Variants of chunked closed-loop control.} Some WAMs keep Equation~\ref{eq:dep-chunk} but make the chunk schedule task-aware. FFDC-WAM~\citep{whentrustimagination2026} is the representative adaptive case, since a lightweight verifier compares predicted and observed latents and triggers an early replan when imagination drifts. NoiseGate~\citep{noisegatelearningper2026} adapts denoising increments, WALL-WM~\citep{wallwmcarving2026} ties the chunk window to the next semantic event, and PALM~\citep{palmprogressaware2026} gates subtask progress inside the chunk. HarmoWAM~\citep{harmowamharmonizinggeneralizable2026} splits reactive and predictive loops, while SWEET~\citep{sweetsparseworld2026} plus BagelVLA~\citep{bagelvlaenhancinglong2026} reduce the substrate per call to task-critical keyframes. These variants still pay the chunked cost, but they spend it when the forecast is likely to change the action.

The benefit of the chunked regime is that it spreads the cost of a large backbone over the chunk and tolerates moderate backbone latency. Its main limitation is the staleness of a long chunk, which becomes visible when the world changes during chunk execution.

\subsubsection{Single-step closed-loop control}
\label{sec:dep-single}

Single-step closed-loop sets $H\!=\!1$ and invokes the WAM at every control step. The model produces one substrate prediction and one action per call, and per-step compute is
\begin{equation}
\mathcal{C}_\text{single}(T) = T\,N_\text{fwd}(1),
\label{eq:dep-single}
\end{equation}
subject to the hard constraint $N_\text{fwd}(1) < 1/f_\text{ctrl}$. The iVideoGPT~\citep{iVideoGPT} antecedent illustrates the autoregressive version of this regime, while current census members such as WorldVLA~\citep{WorldVLA}, VidMan~\citep{VidMan}, GR-1~\citep{GR-1}, PAD~\citep{PAD}, VPP~\citep{hu2024vpp}, DreamZero~\citep{DreamZero}, PhysGen~\citep{PhysGen}, and OmniVTA~\citep{zheng2026omnivta} sit closest to the single-step WAM pattern. Dex4D~\citep{dex4dtaskagnostic2026} runs one forward pass per control tick on the student backbone to output one action and one next-joint prediction, with the goal pointer advancing only when the point-to-point distance falls below a threshold. The latent-imagination Dreamer line~\citep{DreamerV3,Dreamer4,TransDreamer} remains background for this regime when the agent acts at every imagination step. The regime gives the highest reactivity but pays the highest per-step inference cost. It is feasible mainly when the substrate has been chosen sparsely, in the feature, geometric, or affordance categories of Section~\ref{sec:what-substrate}, or in the pixel-decodable latent variant of the pixel-grounded category. In that case one forward pass can fit inside the control period.

\subsubsection{Interactive simulator operation}
\label{sec:dep-interactive}

The fourth regime is the truly interactive simulator, in which a stream of user or actor inputs continuously shapes the ongoing generation and there is no fixed endpoint. The WAM is invoked at every step but reuses the past via a key-value cache or a persistent latent of state size $\mathcal{M}(t)$. Per-step compute is
\begin{equation}
\mathcal{C}_\text{int}(t) = N_\text{fwd}^{\text{cached}}\!\left(\mathcal{M}(t)\right),
\label{eq:dep-int}
\end{equation}
where $\mathcal{M}(t)$ grows with the horizon. For attention-based backbones, the cached forward pass is linear in $\mathcal{M}(t)$ at each step and cumulative cost is quadratic in $t$. InteractiveWorldSimulator~\citep{InteractiveWorldSimulator}, EnerVerse~\citep{EnerVerse}, and LingBot-VA~\citep{LingBotVA} approach this regime among the cited interactive works, while Dreamer~4~\citep{Dreamer4} remains a background lineage point for persistent latent imagination. The latent-imagination lineage from PlaNet~\citep{PlaNet} and DreamerV3~\citep{DreamerV3} through TransDreamer~\citep{TransDreamer} also fits when the actor and the world model run inside a shared rollout loop. The defining property is that the model can be steered mid-stream and the same cache or persistent state carries the long-horizon trajectory. The cost pressure lands on memory rather than on per-step compute. Section~\ref{sec:core-properties} returns to that pressure under the heading of persistence.

\subsection{Putting the Four Ingredients Together}
\label{sec:what-together}

Sections~\ref{sec:what-notation} through~\ref{sec:what-deploy} let us pin down any existing WAM by a compact 4-tuple
\begin{equation}
\mathrm{WAM} \;\cong\; \big(\,\Phi,\ \mathcal{F},\ \mathcal{B},\ \mathcal{D}\,\big),
\label{eq:wam-tuple}
\end{equation}
whose four coordinates take values matching the table columns below. The substrate $\Phi$ from Section~\ref{sec:what-substrate} takes one of \{pixel-grounded, feature, geometric, affordance\}. The table cells record narrower variants where they matter, with decoded or latent for pixel-grounded entries, audio-latent for the acoustic observation entry, and encoder-only, tap, teacher, or VLM token for feature. Here \emph{Pixel (latent)} abbreviates the pixel-decodable latent route above. The action coupling $\mathcal{F}$ from Section~\ref{sec:what-coupling} takes one of \{action-conditioned rollout, joint generation, post-prediction head\}. The backbone $\mathcal{B}$ from Section~\ref{sec:what-backbone} takes one of \{diffusion, autoregressive, joint-embedding, hybrid, LLM/VLM\}. The deployment regime $\mathcal{D}$ from Section~\ref{sec:what-deploy} takes one of \{open, chunked, single, interactive\}.

Example placements show how the tuple is used. F1~\citep{f1vla} is (pixel-decodable latent grid, joint generation, hybrid, chunked). WorldVLA~\citep{WorldVLA} is (pixel-decodable latent grid, joint generation, autoregressive, single). FLARE~\citep{FLARE} is (latent teacher-target, post-prediction head, joint-embedding, chunked). UWM~\citep{UWM} is (pixel-decodable latent grid, joint generation, hybrid, chunked). Frame-level video-diffusion works such as Wan~\citep{Wan} and CogVideoX~\citep{CogVideoX} are \emph{not} WAMs on their own because they fix only $\Phi$ and $\mathcal{B}$. They become a WAM only once $\mathcal{F}$ and $\mathcal{D}$ are added, which is exactly the wrapper-around-a-frozen-backbone pattern that runs through CosmosPolicy~\citep{CosmosPolicy}, VidMan~\citep{VidMan}, and VideoVLA~\citep{VideoVLA} on top of related backbones.

The 4-tuple also clarifies the current direction of WAM design. The first and third coordinates, $\Phi$ and $\mathcal{B}$, receive substantial attention because they are training-time choices and because choosing a pretrained backbone is often the expensive design decision. The middle coordinate $\mathcal{F}$ and the closing coordinate $\mathcal{D}$ are more flexible at inference time, and this flexibility is where many recent mechanisms appear. The census reinforces this asymmetry. Substrate choices remain inside the four categories and backbone choices cluster around diffusion and hybrid families, while sharper variation appears in coupling mechanisms such as iterative-interactive refinement, layer-wise integration, and asynchronous-adaptive denoising, and in runtime mechanisms such as adaptive-$K$ chunking, event-triggered invocation, and dual-frequency slow-fast loops. The expensive corner of the design space combines a decoded pixel-grounded substrate, joint generation, a diffusion backbone, and single-step closed-loop deployment. Few methods can use that corner directly in production. Designs that appear to use it usually trade one coordinate for simulation, an offline rollout, or a smaller model used at inference. Section~\ref{sec:core-properties} examines the same methods through the five properties that embodiment demands of them. Section~\ref{sec:data-eval} examines the data and the evaluation practices the field has settled on to make these design choices empirically meaningful.

Tables~\ref{tab:wams-video} and~\ref{tab:wams-genfree} place every WAM in our census on the 4-tuple of Equation~\ref{eq:wam-tuple}. Table~\ref{tab:wams-video} gathers the works in the Render-and-Decode and Latent-Only philosophies of Section~\ref{sec:philosophies}, while Table~\ref{tab:wams-genfree} gathers the Video-Generation-Free works that drop the video-generation backbone from the predictive path. Within each band, rows follow arXiv first-submission date so that the family's evolution unfolds top to bottom. The substrate column reports where the WAM represents its future, following the four categories of Section~\ref{sec:what-substrate}. The backbone column refers to the families of Section~\ref{sec:what-backbone}. The action coupling column refers to the three families of Equations~\ref{eq:fact-rollout}, \ref{eq:fact-joint}, and~\ref{eq:fact-head}, with \emph{Cond. rollout} used as the compact table label for action-conditioned rollout. The deployment column refers to the regimes of Section~\ref{sec:what-deploy}.

{\scriptsize
\setlength{\tabcolsep}{6pt}
\renewcommand{\arraystretch}{1.25}
\begin{longtable}{@{}l l l l l l@{}}
\caption{Census of World Action Models in the Render-and-Decode and Latent-Only philosophies of Section~\ref{sec:philosophies}. The upper band lists Render-and-Decode methods and the lower band lists Latent-Only methods. Columns give the four-axis anatomy of Section~\ref{sec:what-makes-wam}, locating each work on the design tuple of Equation~\ref{eq:wam-tuple}, and within each band rows follow arXiv first-submission date.}
\label{tab:wams-video}\\
\toprule
\wamheadrow
\midrule
\endfirsthead
\caption[]{Census of World Action Models in the Render-and-Decode and Latent-Only philosophies (continued).}\\
\toprule
\wamheadrow
\midrule
\endhead
\midrule
\multicolumn{6}{r}{\scriptsize\emph{continued on next page}}\\
\endfoot
\bottomrule
\endlastfoot
\wamband{Render-and-Decode}
\midrule
UniPi~\citep{unipi}                                        & 2023.02 & Pixel (decoded) & Diffusion & Post-prediction head & Open-loop \\
AVDC~\citep{ko2024avdc}                                    & 2023.10 & Pixel (decoded) & Diffusion & Post-prediction head & Open-loop \\
VLP~\citep{du2023vlp}                                      & 2023.10 & Pixel (decoded) & Diffusion & Post-prediction head & Chunked \\
GR-1~\citep{GR-1}                                          & 2023.12 & Pixel (latent) & Autoregressive & Joint generation & Single-step \\
Dreamitate~\citep{liang2024dreamitate}                     & 2024.06 & Pixel (decoded) & Diffusion & Post-prediction head & Open-loop \\
This\&That~\citep{wang2024thisthat}                        & 2024.07 & Pixel (decoded) & Diffusion & Post-prediction head & Open-loop \\
GR-MG~\citep{grmg}                                         & 2024.08 & Pixel (decoded) & Diffusion & Post-prediction head & Chunked \\
Gen2Act~\citep{bharadhwaj2024gen2act}                      & 2024.09 & Pixel (decoded) & Diffusion & Post-prediction head & Open-loop \\
GR-2~\citep{GR2}                                           & 2024.10 & Pixel (latent) & Autoregressive & Joint generation & Chunked \\
PAD~\citep{PAD}                                            & 2024.11 & Pixel (latent) & Diffusion & Joint generation & Single-step \\
CoT-VLA~\citep{CoTVLA}                                     & 2025.03 & Feature (VLM token) & LLM/VLM & Post-prediction head & Chunked \\
TesserAct~\citep{TesserAct1}                               & 2025.04 & Geometric & Diffusion & Post-prediction head & Chunked \\
DreamGen~\citep{DreamGen}                                  & 2025.05 & Pixel (decoded) & Diffusion & Post-prediction head & Open-loop \\
WorldVLA~\citep{WorldVLA}                                  & 2025.06 & Pixel (latent) & Autoregressive & Joint generation & Single-step \\
RoboEnvision~\citep{yang2025roboenvision}                  & 2025.06 & Pixel (decoded) & Diffusion & Post-prediction head & Open-loop \\
4DGen~\citep{liu2025fourdgen}                              & 2025.07 & Pixel (decoded) $\wedge$ Geometric & Diffusion & Post-prediction head & Open-loop \\
RIGVid~\citep{patel2025rigvid}                             & 2025.07 & Pixel (decoded) & Diffusion & Post-prediction head & Open-loop \\
Vidar~\citep{vidar}                                        & 2025.07 & Pixel (decoded) & Diffusion & Post-prediction head & Chunked \\
F1~\citep{f1vla}                                           & 2025.09 & Pixel (latent) & Hybrid & Joint generation & Chunked \\
NovaFlow~\citep{li2025novaflow}                            & 2025.10 & Geometric & Diffusion & Post-prediction head & Open-loop \\
UD-VLA~\citep{udvla1}                                      & 2025.11 & Pixel (latent) & Diffusion & Joint generation & Chunked \\
RynnVLA-002~\citep{rynnvla2}                               & 2025.11 & Pixel (latent) & Hybrid & Joint generation & Chunked \\
VideoVLA~\citep{VideoVLA}                                  & 2025.12 & Pixel (latent) & Diffusion & Joint generation & Chunked \\
Motus~\citep{Motus}                                        & 2025.12 & Pixel (latent) & Hybrid & Joint generation & Chunked \\
LVP~\citep{chen2025lvp}                                    & 2025.12 & Pixel (decoded) & Diffusion & Post-prediction head & Open-loop \\
CoVAR~\citep{CoVAR}                                        & 2025.12 & Pixel (latent) & Diffusion & Joint generation & Chunked \\
Dream2Flow~\citep{dharmarajan2025dream2flow}               & 2025.12 & Geometric & Diffusion & Post-prediction head & Open-loop \\
TC-IDM~\citep{tcidmgrounding2026}                          & 2026.01 & Geometric & Diffusion & Post-prediction head & Open-loop \\
BagelVLA~\citep{bagelvlaenhancinglong2026}                 & 2026.02 & Pixel (latent) & Hybrid & Joint generation & Chunked \\
MVISTA-4D~\citep{wang2026mvista4d}                         & 2026.02 & Pixel (decoded) & Diffusion & Post-prediction head & Open-loop \\
Say-Dream-Act~\citep{gu2026saydreamact}                    & 2026.02 & Pixel (decoded) & Diffusion & Post-prediction head & Chunked \\
Dex4D~\citep{dex4dtaskagnostic2026}                        & 2026.02 & Geometric & Hybrid & Post-prediction head & Single-step \\
DreamZero~\citep{DreamZero}                                & 2026.02 & Pixel (latent) & AR / Diffusion & Joint generation & Single-step \\
NovaPlan~\citep{novaplanzeroshot2026}                      & 2026.02 & Geometric & Hybrid & Post-prediction head & Chunked \\
PhysGen~\citep{PhysGen}                                    & 2026.03 & Pixel (latent) & Autoregressive & Joint generation & Single-step \\
EmboAlign~\citep{emboalignaligningvideo2026}               & 2026.03 & Pixel (decoded) & Diffusion & Post-prediction head & Open-loop \\
AeroPlace-Flow~\citep{aeroplaceflowlanguage2026}           & 2026.03 & Geometric & Diffusion & Post-prediction head & Open-loop \\
EVA~\citep{evaaligningvideo2026}                           & 2026.03 & Pixel (decoded) & Diffusion & Post-prediction head & Chunked \\
DriveDreamer-Policy~\citep{drivedreamerpolicygeometry2026} & 2026.04 & Pixel (decoded) $\wedge$ Geometric & Hybrid & Joint generation & Chunked \\
MV-VDP~\citep{multiviewvideo2026}                          & 2026.04 & Pixel (decoded) $\wedge$ Affordance & Diffusion & Joint generation & Chunked \\
DriveVA~\citep{drivevavideoaction2026}                     & 2026.04 & Pixel (latent) & Diffusion & Joint generation & Chunked \\
Veo-Act~\citep{zhang2026veoact}                            & 2026.04 & Pixel (decoded) & Diffusion & Post-prediction head & Chunked \\
ActionImages~\citep{actionimagesend2026}                   & 2026.04 & Pixel (decoded) $\wedge$ Affordance & Diffusion & Joint generation & Chunked \\
GraspDreamer~\citep{graspyoudream2026}                     & 2026.04 & Pixel (decoded) $\wedge$ Geometric & Diffusion & Post-prediction head & Open-loop \\
3D-ALP~\citep{3danchoredlookahead2026}                     & 2026.04 & Pixel (decoded) & Hybrid & Cond. rollout & Chunked \\
VAG~\citep{lang2026vag}                                    & 2026.04 & Feature (tap) & Diffusion & Post-prediction head & Chunked \\
$\pi_{0.7}$~\citep{pi0.7}                                  & 2026.04 & Pixel (decoded) & Hybrid & Post-prediction head & Chunked \\
X-WAM~\citep{guo2026xwam}                                  & 2026.04 & Pixel (latent) & Hybrid & Joint generation & Chunked \\
CKT-WAM~\citep{cktwamparameter2026}                        & 2026.05 & Pixel (latent) & Diffusion & Joint generation & Chunked \\
NoiseGate~\citep{noisegatelearningper2026}                 & 2026.05 & Pixel (latent) & Hybrid & Joint generation & Chunked \\
HarmoWAM~\citep{harmowamharmonizinggeneralizable2026}      & 2026.05 & Pixel (decoded / latent) & Hybrid & Post-prediction head & Chunked \\
DreamAvoid~\citep{dreamavoidcriticalphase2026}             & 2026.05 & Pixel (decoded) & Diffusion & Cond. rollout & Chunked \\
MoLA~\citep{imaginedfuturesexecutable2026}                 & 2026.05 & Pixel (latent) & Diffusion & Post-prediction head & Chunked \\
CreFlow~\citep{creflowcorrectivereflow2026}                & 2026.05 & Pixel (decoded) & Diffusion & Post-prediction head & Chunked \\
Pelican-Unify 1.0~\citep{pelicanunify12026}                & 2026.05 & Pixel (latent) & Hybrid & Joint generation & Chunked \\
SWEET~\citep{sweetsparseworld2026}                         & 2026.05 & Pixel (decoded) & Diffusion & Post-prediction head & Chunked \\
DriveWAM~\citep{drivewamvideogenerative2026}               & 2026.05 & Pixel (latent) & Hybrid & Joint generation & Chunked \\
VERA~\citep{turningvideomodels2026}                        & 2026.05 & Pixel (decoded) & Diffusion & Post-prediction head & Chunked \\
\midrule
\wamband{Latent-Only}
\midrule
ARDuP~\citep{huang2024ardup}                       & 2024.06 & Pixel (latent) & Diffusion & Post-prediction head & Open-loop \\
Im2Flow2Act~\citep{xu2024im2flow2act}              & 2024.07 & Geometric & Diffusion & Post-prediction head & Chunked \\
VPP~\citep{hu2024vpp}                              & 2024.12 & Feature (tap) & Diffusion & Post-prediction head & Single-step \\
VILP~\citep{xu2025vilp}                            & 2025.02 & Pixel (latent) & Diffusion & Post-prediction head & Chunked \\
UVA~\citep{UVA}                                    & 2025.03 & Pixel (latent) & Hybrid & Joint generation & Chunked \\
UWM~\citep{UWM}                                    & 2025.04 & Pixel (latent) & Hybrid & Joint generation & Chunked \\
3DFlowAction~\citep{zhi2025threedflowaction}       & 2025.06 & Geometric & Diffusion & Post-prediction head & Chunked \\
Video Policy~\citep{liang2025videopolicy}          & 2025.08 & Feature (tap) & Diffusion & Post-prediction head & Chunked \\
Genie Envisioner~\citep{liao2025genieenvisioner}   & 2025.08 & Feature (tap) & Diffusion & Post-prediction head & Chunked \\
3D-FDP~\citep{3dflowdiffusion2025}                 & 2025.09 & Geometric & Diffusion & Post-prediction head & Chunked \\
TraceGen~\citep{tracegenworldmodeling2025}         & 2025.11 & Geometric & Diffusion & Post-prediction head & Chunked \\
mimic-video~\citep{pai2025mimicvideo}              & 2025.12 & Feature (tap) & Diffusion & Post-prediction head & Chunked \\
Act2Goal~\citep{Act2Goal}                          & 2025.12 & Feature (tap) & Diffusion & Post-prediction head & Chunked \\
CosmosPolicy~\citep{CosmosPolicy}                  & 2026.01 & Pixel (latent) & Diffusion & Joint generation & Chunked \\
LingBot-VA~\citep{LingBotVA}                       & 2026.01 & Pixel (latent) & Hybrid & Post-prediction head & Interactive \\
GigaBrain-0.5M*~\citep{gigabrain05m2026}           & 2026.02 & Pixel (latent) & Hybrid & Post-prediction head & Chunked \\
AdaWorldPolicy~\citep{AdaWorldPolicy}              & 2026.02 & Pixel (latent) & Hybrid & Joint generation & Chunked \\
3PoinTr~\citep{3pointr3dpoint2026}                 & 2026.03 & Geometric & Hybrid & Post-prediction head & Chunked \\
DiT4DiT~\citep{DiT4DiT}                            & 2026.03 & Pixel (latent) & Diffusion & Post-prediction head & Chunked \\
S-VAM~\citep{yan2026svam}                          & 2026.03 & Feature (tap) & Diffusion & Post-prediction head & Chunked \\
Fast-WAM~\citep{FastWAM}                           & 2026.03 & Feature (encoder-only) & Diffusion & Post-prediction head & Chunked \\
GigaWorld-Policy~\citep{gigaworld-policy}          & 2026.03 & Pixel (latent) & Diffusion & Joint generation & Chunked \\
OmniVTA~\citep{zheng2026omnivta}                   & 2026.03 & Pixel (latent) & Diffusion & Post-prediction head & Single-step \\
VAMPO~\citep{vampopolicyoptimization2026}          & 2026.03 & Feature (tap) & Diffusion & Post-prediction head & Chunked \\
VTAM~\citep{vtamvideotactile2026}                  & 2026.03 & Pixel (latent) & Hybrid & Joint generation & Chunked \\
LaMP~\citep{lamplearningvision2026}                & 2026.03 & Feature (tap) & Hybrid & Post-prediction head & Chunked \\
AIM~\citep{AIM}                                    & 2026.04 & Affordance & Diffusion & Joint generation & Chunked \\
WAV~\citep{WAV}                                    & 2026.04 & Feature (tap) & Diffusion & Post-prediction head & Chunked \\
CLWM~\citep{dexworldmodelcausallatent2026}         & 2026.04 & Feature (teacher) & Hybrid & Joint generation & Chunked \\
MWM~\citep{lou2026mwm}                             & 2026.04 & Affordance & Diffusion & Post-prediction head & Chunked \\
MotuBrain~\citep{MotuBrain}                        & 2026.04 & Pixel (latent) & Hybrid & Joint generation & Chunked \\
FFDC-WAM~\citep{whentrustimagination2026}          & 2026.05 & Pixel (latent) & Hybrid & Joint generation & Chunked \\
DAWN~\citep{dawnworldaction2026}                   & 2026.05 & Feature (teacher) & Hybrid & Joint generation & Chunked \\
EgoExo-WM~\citep{egoexowmunlocking2026}            & 2026.05 & Feature (teacher) & Hybrid & Cond. rollout & Open / Chunked \\
RoboFlow4D~\citep{roboflow4dlightweightflow2026}   & 2026.05 & Geometric & Diffusion & Post-prediction head & Chunked \\
JOPAT~\citep{pointtrackingimproves2026}            & 2026.05 & Pixel (latent) $\wedge$ Geometric & Diffusion & Joint generation & Chunked \\
$\tau_0$-WM~\citep{tau0wm2026}                     & 2026.06 & Pixel (latent) & Hybrid & Joint generation & Chunked \\
WALL-WM~\citep{wallwmcarving2026}                  & 2026.06 & Pixel (latent) & Hybrid & Joint generation & Chunked \\
\end{longtable}
}

\medskip

{\scriptsize
\setlength{\tabcolsep}{6pt}
\renewcommand{\arraystretch}{1.25}
\begin{longtable}{@{}l l l l l l@{}}
\caption{Census of Video-Generation-Free World Action Models, the third design philosophy of Section~\ref{sec:philosophies}. No video-generation backbone is in the predictive path. Columns match Table~\ref{tab:wams-video}, locating each work on the design tuple of Equation~\ref{eq:wam-tuple}, and rows follow arXiv first-submission date.}
\label{tab:wams-genfree}\\
\toprule
\wamheadrow
\midrule
\endfirsthead
\caption[]{Census of Video-Generation-Free World Action Models (continued).}\\
\toprule
\wamheadrow
\midrule
\endhead
\midrule
\multicolumn{6}{r}{\scriptsize\emph{continued on next page}}\\
\endfoot
\bottomrule
\endlastfoot
\wamband{Video-Generation-Free}
\midrule
FLARE~\citep{FLARE}                            & 2025.05 & Feature (teacher) & Joint-embedding & Post-prediction head & Chunked \\
DUST~\citep{DUST}                              & 2025.10 & Feature (VLM token) & Hybrid & Joint generation & Chunked \\
Audio-WM~\citep{learningrobotmanipulation2025} & 2025.12 & Audio (latent) & Diffusion & Post-prediction head & Chunked \\
HiF-VLA~\citep{hifvlahindsight2025}            & 2025.12 & Geometric & LLM/VLM & Joint generation & Chunked \\
DexWM~\citep{DexWM}                            & 2025.12 & Feature (teacher) & Hybrid & Post-prediction head & Chunked \\
PointWorld~\citep{pointworldscaling3d2026}     & 2026.01 & Geometric & Hybrid & Cond. rollout & Open / Chunked \\
PALM~\citep{palmprogressaware2026}             & 2026.01 & Affordance & LLM/VLM & Post-prediction head & Chunked \\
LDA-1B~\citep{LDA1B}                           & 2026.02 & Feature (teacher) & LLM/VLM & Post-prediction head & Chunked \\
FRAPPE~\citep{FRAPPE}                          & 2026.02 & Feature (teacher) & Hybrid & Post-prediction head & Chunked \\
ICLR-VR~\citep{iclrcontextimitation2026}       & 2026.03 & Geometric & LLM/VLM & Joint generation & Chunked \\
DDP~\citep{dreamingunseenworld2026}            & 2026.03 & Feature (teacher) & Hybrid & Joint generation & Chunked \\
ALAM~\citep{alamalgebraicallyconsistent2026}   & 2026.05 & Feature (VLM token) & LLM/VLM & Joint generation & Chunked \\
Feedback-WM~\citep{feedbackworldmodel2026}     & 2026.05 & Feature (teacher) & Hybrid & Cond. rollout & Chunked \\
\end{longtable}
}

%% file: sections/05-core-properties/core-properties.tex
\section{Core Properties of World Action Models}
\label{sec:core-properties}


The four-axis anatomy of Section~\ref{sec:what-makes-wam} explains how a WAM is
assembled. This section asks a different question. Once that assembly is placed inside
a control loop and made to run, what must remain true of it? We focus on five properties. A WAM must be
\textbf{interactable}, so that control signals can shape the predicted future rather
than only be decoded after generation. It must be \textbf{causal}, so that future
information cannot leak into the action being executed. It must be \textbf{persistent},
so that the predicted state remains coherent as the robot acts, observes, and
replans. It must be \textbf{physically plausible}, so that the future is realizable
by the embodiment rather than only visually convincing. It must be
\textbf{generalizable}, so that the same predictive-action contract remains useful
when tasks, objects, scenes, cameras, or embodiments change. These properties are not
independent. A design that improves one property often shifts cost, latency, memory,
or error into another, which is why WAM design is best understood as a set of
trade-offs among prediction quality, runtime feasibility, and control bandwidth.

\subsection{Interactability}
\label{sec:prop-interact}

A video world model is only weakly interactable by default. Prompted to show a cup
being lifted, it will render a plausible lift, but nothing ties that imagined motion to
the trajectory the controller actually intends to send. The action is not yet the
variable that governs the continuation. A WAM
becomes interactable by deciding where the action enters the prediction loop. The
factorization families in Equations~\ref{eq:fact-rollout}, \ref{eq:fact-joint},
and~\ref{eq:fact-head} lay out this design space. This subsection walks from
post-prediction decoding, where action is recovered only after the future has
been produced, through action-conditioned rollout, where proposed actions shape
candidate futures, to joint prediction, where action and future are produced
together.

\noindent\textbf{Post-prediction decoding.} In this form, the model synthesizes a
whole future first and only then recovers control from it. UniPi, AVDC, and VLP establish this form by generating visual plans before
extracting control~\citep{unipi,ko2024avdc,du2023vlp}.
Dreamitate, Gen2Act, and RoboEnvision extend the same post-prediction
interface to tool use, human video, and long-horizon keyframes~\citep{liang2024dreamitate,bharadhwaj2024gen2act,yang2025roboenvision}.
This end of the spectrum is clearest in cascaded WAMs: NovaFlow,
Dream2Flow, and TC-IDM keep the future predictor separate, then convert generated
motion into object flow, tool trajectories, or inverse-dynamics controls~\citep{li2025novaflow,dharmarajan2025dream2flow,tcidmgrounding2026}.
Audio-WM shows the same pattern outside vision, where generated future sound becomes
the policy condition~\citep{learningrobotmanipulation2025}.
This interface is easy to attach to a pretrained video generator, but controllability
arrives only after the full generation cost has been paid.

\noindent\textbf{In-generation control.} A more integrated interface pushes control into the
generative process itself, and the works here line up as a near-monotone progression in
how early the action binds. This\&That starts the progression by conditioning video
diffusion on spatial language and paired gesture coordinates, with ablations showing the
gesture channel is essential for precise manipulation~\citep{wang2024thisthat}. ARDuP
narrows that conditioning to automatically discovered active regions and decodes action
from latent representations rather than rendered pixels~\citep{huang2024ardup}. More integrated
still, CoVAR and VAG run a dedicated action branch alongside video denoising and couple
the two through bridge attention or step-synchronized denoising, passing only compact
pooled video context to the action side~\citep{CoVAR,lang2026vag}. UWM, UVA, F1, Motus,
and Cosmos Policy drop the separate branch and share substrate prediction and action
heads on a single backbone~\citep{UWM,UVA,f1vla,Motus,CosmosPolicy}. AdaWorld binds
earliest of all, injecting learned latent actions at every denoising
step~\citep{AdaWorld,AdaWorldPolicy}. Taken in order, the progression is a dial. The
later the action binds, the cheaper and more modular the model, and the earlier it
binds, the more the predicted future is genuinely shaped by the act it must inform.

\noindent\textbf{Bottlenecked action pathways.} The most bottlenecked designs restrict the
information that the action pathway can consume. AIM exposes the future through a
predicted spatial value map rather than raw future-pixel tokens~\citep{AIM}.
GigaWorld-Policy makes action the primary prediction target and treats video as an
optional action-conditioned by-product~\citep{gigaworld-policy}.
From the subgoal side, $\pi_{0.7}$ supplies sparse multi-view goal images to the
policy context rather than a dense rollout~\citep{pi0.7}. Narrowing the channel this way is exactly what buys cheap, controllable inference,
and it is also what caps the approach. The action can react only to whatever future
variables survive the bottleneck.

\subsection{Causality}
\label{sec:prop-causal}

Causality governs both correctness and latency. A standard video diffusion model
denoises an entire clip jointly, so future frames can influence the representation
that later drives the current action. This bidirectional pattern improves temporal
smoothness, but it prevents the model from acting before imagination has finished.
Worse, a model that quietly attends to the next frame can grasp flawlessly inside its
own rollout and then close on empty air at inference, because the cue it leaned on was
never available in real time. A WAM must instead produce predictions that depend only
on past observations and the
chosen action, and it must expose the relevant part of that prediction quickly enough
for control.

\noindent\textbf{Causal token streams.} Causal streaming makes the ordering explicit:
future and action are decoded as a token stream, and past computation is reused
through a key-value cache~\citep{GR-1,GR2,PhysGen,iVideoGPT,WorldVLA}.
PhysGen makes the mechanism explicit by combining causal masking, lookahead
multi-token prediction, and key-value caching. It predicts several future action
tokens at each step, while execution consumes only the leading token~\citep{PhysGen}.
The gain comes from avoiding repeated computation over context that has already been
fixed.

\noindent\textbf{Leakage-free denoising.} When prediction remains chunked, the key
constraint is leakage control inside the chunk. Action tokens are forbidden from
attending to future-pixel tokens or later actions, so the action head cannot benefit
from information that would be unavailable at inference
time~\citep{WorldVLA,CoTVLA,udvla1}. WorldVLA introduced this mask to control error
propagation in autoregressive action chunks, and the same mask also removes
expensive attention between action tokens and future-pixel tokens~\citep{WorldVLA}.
CoT-VLA uses causal attention to generate a discrete visual chain-of-thought, then
switches to full attention only for the short action sequence that reaches the
goal~\citep{CoTVLA}.

\noindent\textbf{Action-prioritized inference.} Once leakage is controlled, latency is
what is left to fight. Intermediate ODE checkpoints, single forward passes,
and early denoising exits work because the controller usually needs the next action
more than it needs a fully refined future video~\citep{pai2025mimicvideo,yan2026svam,udvla1,DiT4DiT,FastWAM}.
UD-VLA formalizes this point with synchronous joint discrete denoising and reports
roughly fourfold faster inference than autoregressive counterparts~\citep{udvla1}.
The saved computation is mainly the tail of the denoising trajectory, where distant
pixels are refined but the next action has already become stable.

The same prioritization also overlaps prediction with execution. LingBot-VA,
MotuBrain, and DreamZero hide part of world-model latency behind the execution of the
previous action chunk~\citep{LingBotVA,MotuBrain,DreamZero}. LingBot-VA pairs a
Mixture-of-Transformers backbone, which gives shared and modality-specific tokens
their own expert streams, with an asynchronous pipeline that runs action prediction
and motor execution at the same time~\citep{LingBotVA}. DreamZero pushes this overlap
furthest, sustaining closed-loop control from a large autoregressive video-diffusion
model through the cache-level re-grounding we examine under persistence in
Section~\ref{sec:prop-persist}~\citep{DreamZero}.
Diffusion forcing supplies a related middle point: it assigns low noise to the
near-certain past and high noise to the uncertain future under causal masking, so
diffusion quality is retained where uncertainty remains~\citep{chen2025lvp}. Causal
structure thus does two jobs at once. It blocks the future from leaking into the
current action, and it decides where computation is worth spending, on the part of
the rollout that can still move the next decision.

\subsection{Persistence}
\label{sec:prop-persist}

Persistence is coherence under repeated action, observation, and replanning. It fails
through drift, cost growth, and forgetting. Drift appears because each predicted step
conditions the next, so small errors compound until the rollout leaves the data
manifold. Cost grows when a model attends to full history, since memory and attention
increase with the episode horizon. Forgetting appears when a finite context discards
scene identity, so an occluded object can reappear in the wrong place. These failures
pull against each other. Full-history attention fights drift and forgetting but
raises cost. Sliding context bounds cost but discards old evidence.

\noindent\textbf{Observation replacement.} The cheapest persistence mechanism is
re-grounding. After each executed chunk, the method replaces imagined
observations with measured observations and updates the cache before forecasting
again. DreamZero applies this idea through key-value-cache observation replacement
and reaches $7$~Hz closed-loop control with a $14$B
autoregressive video-diffusion model~\citep{DreamZero}. LingBot-VA follows the same
principle with a closed-loop rollout that re-grounds on feedback before predicting
the next chunk~\citep{LingBotVA}. Re-grounding is cheap because it uses the robot's
new observation rather than generating a longer imagined future.

\noindent\textbf{Bounded memory.} Re-grounding limits drift, but memory still needs a
fixed budget. DexWM predicts future latent states from finger-keypoint actions and
maintains bounded test-time memory instead of an unbounded cache, which lets it train
and operate over $900$ hours of human and robot interaction~\citep{DexWM}. Act2Goal uses
multi-scale temporal hashing to keep near-future frames dense while sparsifying
distal frames, preserving long-horizon goal consistency at bounded cost~\citep{Act2Goal}.
LingBot-VA and MotuBrain add per-step cost control through Mixture-of-Transformers
streams, which cap the cost of processing history even when a longer context is
available~\citep{LingBotVA,MotuBrain}.

\subsection{Physical Plausibility}
\label{sec:prop-physical}

The video-world-model instinct equates a good future with a photorealistic one. A
WAM needs something stricter. Its future exists only to constrain action, so what
matters is whether the predicted dynamics are realizable by the body that will
execute them, not whether they look convincing. Contact, force, and kinematics are
what decide that, and three kinds of physical constraint push prediction toward it.

\noindent\textbf{Visual-geometric abstraction ladder.} Prediction targets form a
ladder, and a WAM can choose how far down it to predict. At the top sits full RGB. One
rung down, RGB-D, surface normals, and 4D point maps add explicit geometry to generated
video, as in WAMs that reconstruct structure from RGB-D-Normal
rollouts~\citep{TesserAct1,guo2026xwam,wang2026mvista4d}. Below that,
optical flow and 3D object flow keep only motion, the variable that transfers most
cleanly across humans, robots, simulation, and reality. Im2Flow2Act and 3DFlowAction use
flow to bridge domains and embodiments~\citep{xu2024im2flow2act,zhi2025threedflowaction},
and FlowDreamer carries the same abstraction into flow-conditioned generation~\citep{FlowDreamer}.
Lower still, semantic masks discard photometric detail and keep only what the policy
acts on. MWM makes the case for this rung directly, arguing that RGB prediction is
misaligned with control because masks preserve the geometry while discarding the
photometric nuisance~\citep{lou2026mwm}.

At the bottom, the future stops being rendered at all. VPP and S-VAM decode action from
predicted or distilled foresight features, S-VAM foreseeing geometric and semantic
representations in a single forward pass rather than a video~\citep{hu2024vpp,yan2026svam}.
GigaWorld-Policy, Fast-WAM, WAV, and DexWM keep the future latent or action-centered at
inference~\citep{gigaworld-policy,FastWAM,WAV,DexWM}, and FLARE and LDA-1B push
the target into frozen or structured embedding spaces~\citep{FLARE,LDA1B}. The
descent obeys a single logic. Each rung down trades away appearance the controller never
needed and forces capacity toward geometry, contact, and dynamics, at the price of a
future that is harder to inspect.

\noindent\textbf{Richer physical modalities.} Geometry narrows the target, but vision
alone does not expose the variables that decide contact-rich manipulation, such as
contact state, friction, slip, and normal force. A WAM that predicts only pixels is
therefore fragile in the regime where manipulation is hardest. OmniVTA predicts
future tactile contact states
and pairs a slow policy with a high-rate reflexive controller that corrects
discrepancies between predicted and observed tactile signals~\citep{zheng2026omnivta}.
AdaWorldPolicy adds force prediction as a dedicated branch alongside the world model
and the action expert, then uses force-torque mismatch for online
adaptation~\citep{AdaWorldPolicy}. DexWM predicts finger-level interaction rather than
coarse end-effector motion~\citep{DexWM}, the same fine-grained bet that buys it bounded
memory under persistence (Section~\ref{sec:prop-persist}). These modalities are low-dimensional,
but they carry dense information exactly where contact determines success. They often
offer a cheaper and more action-relevant target than the pixels of the same event.

\noindent\textbf{Proprioceptive and kinematic-chain coherence.} Contact signals still
need to be grounded in the body that executes the action. A predicted future can look
plausible and still be impossible for a particular robot. The arm may pass through
itself, the end-effector may be unreachable, or the hand configuration may violate
joint limits. Physical plausibility therefore also requires
embodiment coherence. PAD, GR-1, Cosmos Policy, and X-WAM condition prediction on
proprioception so that the forecast is tied to joint state~\citep{PAD,GR-1,CosmosPolicy,guo2026xwam}.
Cosmos Policy goes further by treating future proprioception and image observations
as jointly modeled latent frames~\citep{CosmosPolicy}. DexWM adds a
hand-consistency loss that penalizes kinematically inconsistent predictions~\citep{DexWM}.
MVISTA-4D uses trajectory-level inference with a residual inverse model to project an
imagined future onto an action the embodiment can execute~\citep{wang2026mvista4d}.
Each design shrinks the effective search space to the robot's feasible manifold.

These constraints all draw on physical priors learned from internet-scale video
pretraining~\citep{Wan,CogVideoX,cosmos-predict2,Sora,Sora_2,Latte}. The main
lesson is that how useful the imagined future is for the action predicts task
success better than how good it looks~\citep{VideoVLA,FastWAM,gigaworld-policy,WAV}.
A WAM should therefore predict the coarsest embodiment-grounded representation that
still constrains the action, and flow, masks, tactile and force signals, and
proprioceptively consistent latents often fill that role better than photorealism.

\subsection{Generalization}
\label{sec:prop-generalization}

Generalization in robotics is a bundle of shifts rather than a single axis: new
tasks, objects, appearances, rooms, cameras, embodiments, and action spaces. Under
the WAM contract, generalization requires both sides of the model to survive the
shift. The predictive substrate must remain useful, and the action pathway must still
execute the prediction. This is stricter than VLA-style semantic generalization and
stricter than video generalization, because an action-relevant future is required in
addition to a plausible instruction-to-action mapping or a plausible video.

\noindent\textbf{Video-prior transfer.} The broadest transfer source is the video
prior itself. GR-1 and GR-2 show that large-scale language-conditioned video
pretraining improves manipulation in unseen scenes and novel scenarios once action
prediction is fine-tuned~\citep{GR-1,GR2}. VideoVLA ties the prior directly to
action by jointly denoising future frame latents and action tokens, improving
novel-object and cross-embodiment performance while retaining the cost of multi-step diffusion
sampling~\citep{VideoVLA}. DreamZero makes a similar argument with an autoregressive
video-action model and reports unseen-motion gains over VLA baselines, together with
robot-to-robot and human-to-robot transfer from short video-only adaptation
sets~\citep{DreamZero}. LVP keeps the connection cascaded: it trains a large video
planner on human and robot clips, then extracts executable motion through pose
estimation and retargeting for zero-shot real-robot tasks~\citep{chen2025lvp}.
RoboEnvision targets task transfer by turning instructions into future keyframes that
condition the downstream policy, so unseen tasks can still be routed through a visual
subgoal interface~\citep{yang2025roboenvision}. $\pi_{0.7}$ uses generated futures as
subgoal context rather than as the online action generator~\citep{pi0.7}. The common thread is that scale helps only once the
future is connected to the action, and making that connection usually costs an
inverse model, a subgoal generator, or a large video backbone.

\noindent\textbf{Substrate transfer.} When pixels carry too much appearance-specific
detail, transfer shifts to the substrate. Flow-based methods treat object motion as
the transferable variable. Im2Flow2Act uses object flow to bridge human-to-robot and
simulation-to-real gaps before a flow-conditioned policy acts~\citep{xu2024im2flow2act}.
3DFlowAction lifts the same idea into 3D object flow, separating object dynamics
from the robot that later executes the trajectory~\citep{zhi2025threedflowaction}.
What flow buys is suppression of texture, camera style, and part of the embodiment
gap, but it leaves the hard part to the executor, which still has to solve grasping,
collision avoidance, and contact.

Mask, latent, and feature substrates make the same trade-off at a deeper level. MWM
replaces RGB prediction with semantic-mask dynamics, improving robustness to
background, lighting, and object-color shifts~\citep{lou2026mwm}. S-VAM distills a
diffusion backbone into geometric and semantic foresight features, so inference no
longer depends on full video denoising~\citep{yan2026svam}. Fast-WAM and
GigaWorld-Policy keep video prediction as a training signal but remove future-video
generation from part or all of inference~\citep{FastWAM,gigaworld-policy}. The deepest
version of the move forecasts in feature space rather than pixels. FRAPPE and
LDA-1B predict in self-supervised vision-foundation features, where appearance
variation washes out and what survives the shift is the action-relevant
structure~\citep{FRAPPE,LDA1B}.
These substrates make generalization less dependent on photorealistic rendering, but
they also make the predicted future harder to inspect and harder to evaluate with
video metrics.

\noindent\textbf{Action-abstraction transfer.} Substrate transfer still needs an
executor, so another line of work transfers the action abstraction and the data
recipe. LDA-1B scales this separation with universal embodied data ingestion, a
hand-centric action space, and DINO latent forecasting across high-quality
demonstrations, low-quality trajectories, and actionless videos~\citep{LDA1B}. DUST
shows the diffusion-side version of the point: action-free BridgeV2 video pretraining
improves a dual-stream policy even when future observations are predicted only in an
embedding space~\citep{DUST}. ALAM adds an algebraic consistency constraint over
latent action transitions, so the abstract action space keeps a predictable
composition law while the action expert remains tied to the embodiment~\citep{alamalgebraicallyconsistent2026}. The
shared move is to separate what should transfer from what should stay local. The
substrate carries task and environment dynamics. The action
decoder absorbs embodiment-specific kinematics. The data recipe supplies enough
variation that neither side memorizes the training platform.

What falls out in practice is a single principle. Predict at the invariant level that
still constrains control. Pixels transfer semantic and physical priors but carry appearance cost.
Flow and masks transfer object motion and geometry but require a reliable executor.
Latent actions and feature forecasts scale to action-free video and heterogeneous
data, but they need careful grounding before they become precise control. No design
dominates across all shifts. A WAM that claims generalization should therefore state
which shift it targets, which design axis is meant to transfer, and which adapter
remains specific to the robot that executes the action. This framing turns
generalization from a broad claim into a checkable property.

Across the five properties, one lesson outweighs the rest. They are not a checklist but
a set of competing pressures. The stronger action interface that makes a model interactable also
narrows the channel that causality must keep leak-free. The bounded memory that buys
persistence is the same budget that starves long-horizon plausibility. The abstraction
that lets a substrate transfer is what makes its predicted future hard to evaluate. A
WAM is therefore never tuned for one property alone. Every design sits at a point where
a gain in one property is paid for in another, and turning that balance into something
measurable, rather than asserted, is the task we take up in Section~\ref{sec:data-eval}.

%% file: sections/06-data-and-evaluation/data-and-evaluation.tex
\section{Data and Evaluation}
\label{sec:data-eval}


A WAM can only preserve the properties of Section~\ref{sec:core-properties} when its
training data and evaluation protocol expose those properties. Data determines which
future variables the model can learn, which action labels it can trust, and which
embodiments it can transfer across. Evaluation determines whether those learned
variables are useful in a control loop rather than only plausible as video. The two
questions are therefore coupled. This section first organizes WAM data by the
trade-off between scale, action-label quality, and embodiment match. It then
organizes evaluation by the trade-off between visual fidelity, closed-loop success,
physical plausibility, and evaluation cost.

\subsection{Data sources}
\label{sec:data}

WAM training data falls into five groups. Each group provides a different compromise
between scale, action-label fidelity, physical grounding, and access.

\paragraph{Robot teleoperation.} Robot teleoperation gives the cleanest
action-conditioned trajectories because a human operator drives the embodiment and
the command stream is recorded with the visual observation. Open X-Embodiment~\citep{OXE}
aggregates this recipe across laboratories and robot platforms, making it the
canonical cross-embodiment resource for robot policy training. RoboMIND~\citep{RoboMIND},
RoboSet~\citep{RoboSet}, RoboTurk~\citep{RoboTurk}, and Robo360~\citep{Robo360} extend the same supervision pattern with different
platform mixes, task suites, and camera configurations. Human2Robot~\citep{Human2Robot}
pairs each teleoperated trajectory with a synchronized video of a bare human hand
performing the same task, adding a human-to-robot correspondence on top of the exact
action labels.
RoboNet~\citep{RoboNet} departs from human teleoperation by collecting real-robot trajectories
through scripted and random policies, trading demonstration precision for
throughput while still logging exact action commands. The benefit of
this group is label fidelity. The cost is that every additional hour consumes robot
time, operator time, or both.

\paragraph{Portable human demonstrations.} Portable demonstrations increase
throughput by moving data collection from a robot platform to a human-worn or
handheld capture setup. EgoMimic~\citep{EgoMimic} records egocentric video and three-dimensional hand
tracks with a lightweight wearable rig. EgoVerse~\citep{EgoVerse} scales the same
idea with wearable and phone-based capture, then retargets the demonstrations across
robot embodiments. EgoDex~\citep{EgoDex} carries the head-worn route to large scale,
recording dexterous bimanual hand-pose tracks on Apple Vision Pro with no robot in
the loop. This group buys
scale because human demonstrations are faster to collect than robot teleoperation.
It pays for that scale with an embodiment gap that must be closed before the data can
train a robot policy.

\paragraph{Internet-scale egocentric and instructional video.} Internet video gives
WAMs their scale, but it usually lacks robot actions. Ego4D~\citep{Ego4D}, Ego-Exo4D~\citep{Ego-Exo4D}, and
EgoPAT3D~\citep{EgoPAT3D} emphasize first-person activity and hand motion.
EPIC-KITCHENS~\citep{EPIC-KITCHENS}, HD-EPIC~\citep{HD-EPIC}, EgoVid-5M~\citep{EgoVid-5M}, OpenEgo~\citep{OpenEgo}, IndEgo~\citep{IndEgo}, EgoScale~\citep{egoscale}, and Egocentric-10k~\citep{Egocentric-10k}
broaden the pool with instructional, household, and large-scale egocentric
video.
This group is valuable because it exposes broad visual variation, object dynamics,
and human manipulation patterns. Its limitation is the missing action channel. One
route uses such video, together with broader internet video, to pretrain a video
backbone, as in V-JEPA~2~\citep{V-JEPA-2} and Wan~\citep{Wan}. Another route attaches an inverse-dynamics or
latent-action model that recovers a control proxy from video. AVDC~\citep{ko2024avdc} recovers
flow-conditioned action from unlabeled video, while LDA-1B~\citep{LDA1B} and DUST~\citep{DUST} use action-free video to strengthen latent future prediction before action decoding. Internet video is
therefore a scale source only after the action-label problem has been made explicit.

\paragraph{Simulation.} Simulation provides exact action labels, controlled
curricula, and low marginal cost after the environment has been authored. Classical
physics platforms for manipulation include Robosuite~\citep{robosuite}, ManiSkill~2~\citep{ManiSkill2}, ManiSkill~3~\citep{tao2024maniskill3},
MetaWorld~\citep{yu2019metaworld}, and LIBERO~\citep{LIBERO}.
General simulation backends such as IsaacSim~\citep{IssacSim}, CoppeliaSim~\citep{CoppeliaSim}, and SimplerEnv~\citep{li2024simplerenv} provide the
execution substrate for broader embodied tasks.
Household and digital-twin datasets build on these engines to generate demonstrations
programmatically, including RoboCasa~\citep{RoboCasa}, RoboTwin~\citep{RoboTwin}, and RoboTwin~2~\citep{RoboTWIN_2}.
RoboCerebra~\citep{RoboCerebra} instead collects human-teleoperated trajectories inside the
simulator for long-horizon evaluation. Aggregators sit above
both lines: RoboVerse~\citep{geng2025roboverse} unifies several simulator backends, and RoboData~\citep{RoboData} merges
simulation benchmarks with a real-robot set. The
benefit is scalable label quality. The limitation is the sim-to-real gap, which a
WAM often tries to close with an internet-pretrained visual prior or a more abstract
substrate.

\paragraph{Synthetic data from WAMs themselves.} A WAM can also become a data engine.
IRASim~\citep{IRASim} follows this pattern from the simulator side by
replacing a physics engine with a learned action-conditioned diffusion model whose
rollouts stand in for simulated trajectories. Cosmos-Transfer1~\citep{Cosmos-Transfer1} and
InteractiveWorldSimulator~\citep{InteractiveWorldSimulator} support similar data-generation use, and Sora-class video
models are sometimes used to seed downstream policies~\citep{Sora,Sora_2}.
DreamGen~\citep{DreamGen} makes the cascaded WAM version concrete by generating robot
videos, recovering pseudo-actions or latent actions, and training the final policy
from those neural trajectories. RIGVid~\citep{patel2025rigvid} and GraspDreamer~\citep{graspyoudream2026}
use generated demonstrations in a more direct action-recovery path, where tracking
and retargeting replace manual action labels.
Synthetic neural trajectories sit between simulation and real teleoperation. They
inherit the realism of internet pretraining, but they also inherit the generator's
failure modes.

In practice, WAM training mixes these sources. Internet video supplies visual and
physical priors, teleoperation supplies trusted action labels, portable human data
adds scale, simulation supplies controllable coverage, and synthetic trajectories
fill gaps that are expensive to collect physically. The unresolved question is how to
choose the mixture for a target deployment regime. Access is a second constraint.
Sora-class generators~\citep{Sora,Sora_2} and the EgoScale release~\citep{egoscale} are
closed or not yet public, which limits
independent verification of their contribution.

\subsection{Evaluation}
\label{sec:eval}

Evaluation must judge both prediction and action. Video-generation metrics ask
whether the future looks plausible. Robot-learning benchmarks ask whether the policy
succeeds in the loop. A WAM needs both perspectives, but neither is sufficient on its
own. A useful protocol must also ask whether the prediction is physically realizable,
whether coherence survives long horizons, and what compute, memory, and latency are
required to obtain the result.

\paragraph{Visual fidelity metrics.} The video-generation lineage contributes FVD~\citep{unterthiner2018fvd},
FID, LPIPS~\citep{lpips}, PSNR, SSIM~\citep{ssim}, and DreamSim~\citep{fu2023dreamsim}.
These metrics are cheap and familiar, but they reward visual realism rather than
action utility. A crisp rollout can still be useless when contact, geometry, or
proprioception is wrong. A visually sparse representation can still drive a
successful policy when it captures the variables needed for control. Fast-WAM~\citep{FastWAM}
and GigaWorld-Policy~\citep{gigaworld-policy} bear this out by dropping the rendered
forecast at inference without losing control, while WorldScore~\citep{duan2025worldscoreunifiedevaluationbenchmark}
and WorldSimBench~\citep{qin2024worldsimbenchvideogenerationmodels} fold visual fidelity
into broader assessments that pair perceptual scores against closed-loop action success.

\paragraph{Closed-loop benchmarks.} Closed-loop evaluation measures whether the
policy actually completes the task. These benchmarks differ mainly in where the
rollout comes from, which sets a throughput-versus-validity trade-off. Simulation sits
at the high-throughput end, where benchmarks probe different competences. The LIBERO family~\citep{LIBERO,zhou2025liberopro,wang2026liberox,fei2025liberoplus} and RoboMME~\citep{dai2026robomme} focus on language-conditioned manipulation and
multimodal evaluation.
robomimic~\citep{mandlekar2021robomimic}, MetaWorld~\citep{yu2019metaworld}, ManiSkill~2~\citep{ManiSkill2}, and ManiSkill~3~\citep{tao2024maniskill3} provide manipulation suites with
different levels of control diversity and physics coverage.
HomeRobot~\citep{yenamandra2023homerobot}, VLABench~\citep{zhang2024vlabench}, RoboEval~\citep{wang2025roboeval}, and RoboVerse~\citep{geng2025roboverse} broaden the benchmark space toward
navigation, embodied reasoning, and multi-backend evaluation.
Real-robot arenas such as RoboArena~\citep{atreya2025roboarena} and RoboChallenge~\citep{yakefu2025robochallenge} trade throughput for physical
validity by running the policy on hardware.
DWorldEval~\citep{li2026dworldevalscalableroboticpolicy} closes the loop inside a learned world model instead, scoring a policy
from imagined rollouts whose success correlates with real execution.
Cascaded WAMs add another evaluation layer because the generated future and the
action translator can each fail separately. Video-to-flow and video-to-trajectory
methods therefore need both substrate checks and execution checks, as illustrated by
NovaFlow, Dream2Flow, TraceGen, and RoboFlow4D~\citep{li2025novaflow,dharmarajan2025dream2flow,tracegenworldmodeling2025,roboflow4dlightweightflow2026}.
Closed-loop evaluation measures the right outcome, but every trial consumes a robot,
a simulator, or a learned generator for the duration of the task.

\paragraph{Physical plausibility and long-horizon coherence.} Some WAM properties are
visible in a rollout but poorly captured by a single success number. Physical
plausibility can be evaluated through tactile and force prediction errors, as in
OmniVTA~\citep{zheng2026omnivta} and AdaWorldPolicy~\citep{AdaWorldPolicy}, or through
kinematic and geometry-consistency checks, as in DexWM~\citep{DexWM} and
MVISTA-4D~\citep{wang2026mvista4d}. A standard metric has not yet emerged.
Long-horizon coherence is similarly under-specified. Reported numbers usually reduce
to task success after a fixed number of replans, and current benchmarks do not test
hour-scale rollout at control frequency. RoboScape~\citep{RoboScape} brings physics signals into evaluation,
while PhysWorld~\citep{Robot_Learning_from_a_Physical_World_Model} points to the broader problem of judging physical world
models.

\paragraph{What current evaluation misses.} Current evaluation has a predictable
ordering. Visual-fidelity metrics are cheap but weakly tied to downstream success.
Closed-loop simulation is more predictive but consumes task-time rollouts. Real-robot
arenas are physically valid but expensive to repeat. Physical-plausibility and
long-horizon metrics remain underdeveloped. The same ordering appears in data
collection. A teleoperation hour costs a robot and an operator, a simulator hour
costs compute, an internet-video hour costs curation, and a learned-model rollout
costs a forward pass. This is why proxy evaluators are attractive despite depending
on the generator's own fidelity. The WAMs that matter for real use are also the
hardest to evaluate, because chunked and interactive closed-loop control require
longer horizons, stricter latency constraints, and real-environment variation. The field is
therefore moving toward a two-stage protocol: inexpensive visual and representation
screens, followed by selective closed-loop tests under explicit compute, memory, and
latency budgets. The conversion factor between the cheap screen and the expensive
test remains the central open question in WAM evaluation, and it motivates the
frontiers in Section~\ref{sec:open}.

%% file: sections/07-open-challenges/open-challenges.tex
\section{Open Challenges}
\label{sec:open}


Sections~\ref{sec:what-makes-wam} through~\ref{sec:data-eval} define the design axes, core properties, data sources, and evaluation practice of current World Action Models. The open challenges below use those axes to separate problems that are often discussed together. The boundaries are analytical rather than absolute: data affects generalization, action abstraction affects physics, and inference cost affects almost every design choice. Still, each subsection asks a different question. We first ask how much future computation a WAM should build into its design or spend at runtime, then ask which data should train each stage. We then move through memory, generalization, action grounding, physical plausibility, and evaluation.

\subsection{Dream More or Act More?}
\label{sec:open-fidelity}
\label{sec:open-budget}

Prediction helps action only when its latency fits the decision loop. Richer futures, larger backbones, and deeper denoising can improve the substrate available to the action side, but they also lengthen the path from observation to control. Current WAMs therefore split between computing a detailed future and extracting just enough future-trained structure for action. S-VAM distills video diffusion into geometric and semantic foresight features, Fast-WAM removes the future-video branch at inference, GigaWorld-Policy masks attention so future video can be skipped, and X-WAM lets action denoising finish before video denoising completes~\citep{yan2026svam,FastWAM,gigaworld-policy,guo2026xwam}. These methods suggest that the video objective can be more important than a fully rendered test-time forecast.

The other side keeps stronger imagination in the control path and pays for it in latency. DreamZero brings a large video backbone into closed-loop control through a custom optimization stack, CosmosPolicy uses repeated model and value queries for planning, and NovaPlan places video generation before flow extraction and geometric grounding~\citep{DreamZero,CosmosPolicy,novaplanzeroshot2026}. The open problem is not simply to make every model smaller. It is to expose a controllable fidelity-latency curve, so a WAM can choose how much future information the action actually needs.

That choice should also become a runtime decision. A video policy may run the same denoising schedule at every step, and a latent planner may fix its horizon and search depth before it knows whether the next state is routine transit or contact-sensitive manipulation~\citep{VideoVLA,PlaNet}. Coarse switches help but remain limited. A forecast-confidence method can choose when to replan, yet a binary execute-or-replan decision cannot decide how much of the remaining horizon should be regenerated~\citep{whentrustimagination2026}. DreamAvoid, SANTS, NoiseGate, and HarmoWAM show that the trigger can be learned from task phase, denoising state, or process cues~\citep{dreamavoidcriticalphase2026,santsstateadaptive2026,noisegatelearningper2026,harmowamharmonizinggeneralizable2026}. The open problem is to spend generative compute by expected action value: more when uncertainty, contact, or irreversible error is near, and less when the next action is routine.

\label{sec:open-unify}
This budget question also clarifies the relation between video world models and WAMs. Video generation has pushed toward longer and more detailed futures, while WAMs increasingly try to compute only the future information needed for action. The two lines now meet at the substrate level of Section~\ref{sec:what-substrate}: video generators add action conditioning, and WAMs borrow latent or joint-embedding substrates from video backbones~\citep{Cosmos-Transfer1,cosmos-predict2,InteractiveWorldSimulator,pelicanunify12026,UnifiedVLA}. The practical obstacle is that a video-scale model reaches the control rate only after substantial optimization, while skipping video tokens recovers speed by giving up the rendered forecast~\citep{DreamZero,gigaworld-policy,FastWAM}. A stronger design would train one backbone with video objectives, then let the action path choose how much of that backbone to run at inference.

\subsection{What Data Should Each Stage Learn From?}
\label{sec:open-curriculum}

Section~\ref{sec:data-eval} organizes WAM data by source, but the harder question is where each source belongs in training. A WAM usually needs visual dynamics, embodiment alignment, executable action, and sometimes a post-training controller or scheduler. These variables are not learned from the same evidence. Internet video can teach broad visual regularities, egocentric or third-person human demonstrations can align motion with intent, robot teleoperation can ground the final action decoder, and rollout data can train gates or budget controllers. Pooling these sources hides the variable each source supervises. The open challenge is to assign each data source to a stage, an objective, and a model component.

The pretraining stage is the clearest starting point and also the easiest stage to overgeneralize. GR-2 and DreamDojo show the appeal of broad video pretraining before robot-specific adaptation, while VPP shows that a video prior can be decisive for downstream action when the robot data alone is too narrow~\citep{GR2,DreamDojo,hu2024vpp}. Yet robot-domain video and general human video are not interchangeable. VidMan reports that robot-domain pretraining helps while a mismatched egocentric source can hurt the same downstream setting, and VLA-JEPA finds that human-video gains depend on the evaluation split rather than appearing uniformly~\citep{VidMan,VLA-JEPA}. The question is therefore not whether video pretraining helps. The question is which kind of video teaches the dynamics that the later action path will need.

The alignment stage is less settled. Egocentric human data, paired human-robot demonstrations, and wearable capture can bridge the gap between passive video and robot control, but each format carries a different bias. EgoScale separates broad human pretraining from an aligned mid-training stage, EgoMimic uses first-person human demonstrations to improve robot imitation, and EgoVerse finds that scene diversity can matter more than raw volume~\citep{egoscale,EgoMimic,EgoVerse}. FRAPPE makes the same point from a different angle: web-scale egocentric data, task-specific egocentric data, and robot data play distinct roles in transfer~\citep{FRAPPE}. These results argue for a staged curriculum, but they do not yet tell us the minimum capture hardware, viewpoint match, or data quality needed for each embodiment.

The final action stage is where the data bottleneck becomes sharpest. Cross-platform corpora and simulation benches provide a practical base for action supervision, but their action spaces remain uneven across arms, hands, mobile bases, and deformable-object tasks~\citep{OXE,RoboMIND,RoboCasa,RoboCerebra,RoboTWIN_2,geng2025roboverse}. Action-free recipes try to reduce this bottleneck by discovering latent actions, flow actions, or structured transition codes from video, as in LAPA, DUST, ALAM, and LDA-1B~\citep{LAPA,DUST,alamalgebraicallyconsistent2026,LDA1B}. These recipes reduce the amount of labeled robot control needed, but they do not remove it. Every abstract action still needs a decoder, an inverse-dynamics model, or a robot-specific calibration step before it can become executable control.

Synthetic trajectories add a further stage rather than a complete replacement for action labels. DreamGen uses generated futures to produce neural trajectories, and physical-world-model pipelines can turn video generation into policy training data, but both still inherit generator errors and still need grounding in robot dynamics~\citep{DreamGen,Robot_Learning_from_a_Physical_World_Model}. Human2Robot shows the other side of the same problem: paired human-robot data can be valuable, but contact-rich skills remain hard to capture even with specialized teleoperation hardware~\citep{Human2Robot}. The missing result is a joint scaling law over source, stage, quality, model size, and action grounding. Without such a law, adding more data remains an engineering choice rather than a predictable design decision.

\subsection{Can Memory Keep Up?}
\label{sec:open-persist}

Section~\ref{sec:core-properties} catalogs mechanisms for persistence, including observation replacement, bounded memory, and spatially indexed state. What remains open is not simply a longer prediction horizon. A useful WAM must maintain task state while actions are being produced, and its memory cost should grow with scene complexity rather than with episode length. Autoregressive prediction exposes the failure mode directly: IRASim accumulates temporal drift despite overlapping context, PhysGen is bounded by a fixed context of physical packages, and sparse-memory chunking still degrades once the scene becomes dynamic~\citep{IRASim,PhysGen,EnerVerse}. These are memory failures, not only prediction-quality failures.

Promising ingredients exist in separate places. A test-time memory can keep per-step recall bounded, but its update remains sensitive enough to risk forgetting earlier state~\citep{dexworldmodelcausallatent2026}. Dreamer~4 shows that compact latent state can support continual imagination under closed-loop control, but the abstraction leaves open how to preserve object-level spatial detail across long manipulation episodes~\citep{Dreamer4}. The challenge is to combine bounded memory, spatial indexing, and observation replacement into one method that remains reactive over long tasks.

\subsection{How Can WAMs Generalize?}
\label{sec:open-generalize}

Section~\ref{sec:core-properties} treats generalization as a family of shifts, not as one scalar property. A WAM therefore cannot be made generalizable by adding scale alone. The target shift has to be named first, because visual appearance, object dynamics, morphology, action space, camera placement, and contact regime stress different parts of the model. Domain randomization remains a sharp test of this separation. Act2Goal drops steeply under harder randomized conditions, and FRAPPE remains far below a practical rate even when it leads the compared baselines~\citep{Act2Goal,FRAPPE}. Transfer to new motions and objects is also bounded: a video-conditioned policy improves over baselines, but the hardest motion and object splits still leave a large gap~\citep{bharadhwaj2024gen2act}. The lesson is that future prediction helps only when the predicted substrate matches the shift that the action path must absorb.

The path toward generalizable WAMs is to match data, substrate, and adaptation to the shift being targeted. Cross-embodiment transfer is the clearest case, because the action space itself changes. DreamZero reports transfer from short video-only adaptation sets, but the need for adaptation shows that morphology shift is not solved by scale alone~\citep{DreamZero}. Removing video pretraining from a strong policy sharply reduces task completion, which shows that the borrowed video prior matters, but it does not explain which embodiment changes the prior can absorb~\citep{hu2024vpp}. EgoScale offers one positive sign through a clean scaling law for dexterous pretraining volume~\citep{egoscale}. The open challenge is to design for a declared shift before training, then test whether the chosen substrate and action decoder generalize across that shift rather than reporting transfer after the fact.

\subsection{What Grounds Abstract Action?}
\label{sec:open-grounding}

The action-coupling axis of Section~\ref{sec:what-makes-wam} often replaces raw control with an abstract target such as a latent code, a flow field, a point track, or a structured transition. This substitution is useful because it lets a method learn from action-free or weakly labeled video. It is also risky because the action can lose a physical handle. One failure mode is estimator dependence. Two-dimensional object flow cannot represent depth translation, out-of-plane rotation, or contact force, and a rigid-transform decoder breaks when the object is articulated or deformable~\citep{xu2024im2flow2act,ko2024avdc}. Point-track and flow-based adapters make the action interface inspectable, but they inherit errors from trackers, depth estimators, and local linearity assumptions~\citep{pointtrackingimproves2026,3pointr3dpoint2026,turningvideomodels2026}.

The second failure mode is opacity. LAPA learns discrete latent actions from video and improves action-free pretraining, but fine grasping remains difficult when the codebook is too coarse~\citep{LAPA}. ALAM adds algebraic consistency to latent transitions, yet the resulting structure is still a composition law rather than a force, torque, or contact explanation~\citep{alamalgebraicallyconsistent2026}. LDA-1B learns a broad latent action space, but execution still depends on the decoder that maps that space back to an embodiment-specific command~\citep{LDA1B}. The open challenge is to keep the data advantage of abstract actions while adding a physical grounding signal and an inspection handle for failure analysis.

\subsection{When Is a Future Physical?}
\label{sec:open-physics}

Section~\ref{sec:core-properties} frames physical plausibility through visual-geometric abstraction, tactile and force prediction, and proprioceptive or kinematic coherence. The difficulty is that a future can look plausible while failing the embodiment. Momentum is a simple example. A flow-based world model conditioned on one frame cannot encode object velocity, so push tasks fail when the object keeps moving after contact ends~\citep{FlowDreamer}. Contact and force are harder still. A tactile world model derives a force proxy from tactile-flow divergence, but the proxy is uncalibrated, and a point-cloud substrate omits force magnitude, friction, and surface compliance~\citep{vtamvideotactile2026,pointworldscaling3d2026}. Even an executability reward can miss contact if it only scores smoothness and joint limits~\citep{evaaligningvideo2026}.

Some methods point toward a better standard by making physics part of the substrate or loss. RoboScape and RoboScape-R add geometric and consistency constraints, OmniVTA and DexWM bring tactile or dexterous state into the prediction problem, and AdaWorldPolicy includes action-conditioned future structure in its training objective~\citep{RoboScape,RoboScape-R,zheng2026omnivta,DexWM,AdaWorldPolicy}. These examples remain partial. The open challenge is to train and evaluate futures for being executable by the embodiment, not only visually convincing.

\subsection{What Should Evaluation Report?}
\label{sec:open-eval}

Section~\ref{sec:data-eval} describes visual screens, simulator tests, and closed-loop robot evaluation. The open challenge is to report the variables that make these tiers comparable. World-model quality is not yet a reliable proxy for policy utility, as MotuBrain finds only a weak correlation between its world-model score and downstream success~\citep{MotuBrain}. A single success number can also hide opposite regimes. HarmoWAM shows that a schedule useful for transit can fail at precise interaction, which means that averaging the phases removes the failure mode the evaluation should expose~\citep{harmowamharmonizinggeneralizable2026}.

The missing protocol is an accuracy-at-budget report. A WAM that succeeds slowly is not equivalent to one that acts at the control rate, and a WAM that succeeds over a short horizon is not equivalent to one that keeps state over a long task. Some papers report latency, but often apart from accuracy and on different hardware or task settings~\citep{CoVAR,DiT4DiT,multiviewvideo2026,navdreamervideomodels2026,du2023vlp}. Better examples are emerging: Fast-WAM isolates the value of test-time video generation once video co-training is preserved, while SANTS and NoiseGate report success together with learned reductions in denoising cost~\citep{FastWAM,santsstateadaptive2026,noisegatelearningper2026}. Evaluation should place success, latency, sustained horizon, peak memory, and contact-sensitive failure tags on the same axis.

\subsection*{The challenges are coupled}

These challenges are coupled even when the boundaries above separate their main questions. A memory fix that deepens search raises the compute budget. A richer substrate can improve contact realism while making evaluation and inference more expensive. Latent actions reduce the need for labeled robot control, but the same abstraction weakens physical grounding unless the decoder is calibrated. Contact-aware prediction requires force, tactile, or failure data that cheap visual screens do not provide. Upstream estimator errors can appear as both grounding failures and transfer failures. The shared lesson is that a WAM should be advanced as one coupled design space, where progress on one challenge changes the practical trade-offs of the others.

%% file: sections/08-conclusion/conclusion.tex
\section{Conclusion}
\label{sec:conclusion}


This survey has treated World Action Models as a contract between prediction and
control: given past observations, past actions, and a task context, a WAM must
anticipate the actionable future while producing the action that should follow. This
framing separates WAMs from broad world models, video generation models, and plain VLAs, and it
explains why works with different names can still belong to the same family. We
organized existing works through two complementary views. The philosophy-level view
asks what a method is required to generate before action is decoded, while the
component-level anatomy locates the same method by predictive substrate, backbone,
action coupling, and deployment regime. These views then support the core properties
that matter after a model enters a control loop: interactability, causality,
persistence, physical plausibility, and generalization. They also show why data and
evaluation are not secondary details, since a benchmark can only validate the
properties it exposes and a dataset can only teach the action variables it records or
infers. Across the works considered here, the common trajectory is not simply toward larger
video prediction, but toward more selective imagination that keeps the parts of the
future needed for action and discards the rest. The remaining challenges therefore lie
in choosing the right substrate, coupling action without leakage, preserving
long-horizon state, grounding predictions in physical constraints, and reporting
capability together with compute, memory, and latency. By giving these choices a
shared vocabulary, this survey aims to make future WAMs easier to compare and to help
the field generate less of the future while preserving more of what embodied action
requires.

%% file: sections/_refs_nocite.tex
\nocite{unipi,du2023vlp,ko2024avdc,GR1,liang2024dreamitate,huang2024ardup,wang2024thisthat,xu2024im2flow2act,grmg,bharadhwaj2024gen2act,GR2,PAD,hu2024vpp,xu2025vilp,UVA,CoTVLA,TesserAct1,UWM,yang2025roboenvision,zhi2025threedflowaction,WorldVLA,vidar,liu2025fourdgen,patel2025rigvid,liang2025videopolicy,f1vla,li2025novaflow,rynnvla2,udvla1,dharmarajan2025dream2flow,chen2025lvp,pai2025mimicvideo,VideoVLA,CoVAR,DexWM,Motus,Act2Goal,CosmosPolicy,LingBotVA,gu2026saydreamact,wang2026mvista4d,DreamZero,PhysGen,AdaWorldPolicy,yan2026svam,zheng2026omnivta,gigaworld-policy,DiT4DiT,FastWAM,zhang2026veoact,lang2026vag,pi0.7,lou2026mwm,guo2026xwam,AIM,MotuBrain,WAV,iVideoGPT,FlowDreamer,EnerVerse,PlaNet,TransDreamer,V-JEPA,MoCoGAN,UNet,Latte,Wan,Sora_2,SWIM,DreamDojo,RoboDreamer,RoboScape,VideoPoet,bagel,CogVideoX,cosmos-predict2,guo2024animatediff,NOVA,InteractiveWorldSimulator,geminiroboticsteam,oord2017vqvae,labbe2022megapose,wen2024foundationpose,karaev2023cotracker,kirillov2023sam,Dreamer3,DreamerV3,Dreamer4}